\pdfoutput=1

\documentclass[11pt]{article}
\usepackage[dvipsnames,table,xcdraw]{xcolor}

\usepackage[final]{acl}

\usepackage{times}
\usepackage{latexsym}

\usepackage[T1]{fontenc}

\usepackage[utf8]{inputenc}

\usepackage{microtype}

\usepackage{inconsolata}

\usepackage{graphicx}               
\usepackage{tabularx}               
\usepackage{booktabs}
\usepackage{amsmath}
\usepackage{stmaryrd}

\usepackage{soul}

\newcolumntype{C}{>{\centering\arraybackslash}X}
\usepackage{multirow}               
\usepackage{diagbox}                
\usepackage{hhline}                 
\usepackage{color,colortbl}         
\usepackage{amsmath}                
\usepackage{amssymb}                
\usepackage{mathtools}              
\usepackage{enumitem}               

\usepackage{subcaption}
\usepackage{caption}
\usepackage{booktabs}
\usepackage{extarrows}
\usepackage{makecell}
\usepackage{cleveref}
\usepackage{hyperref}
\usepackage{fdsymbol}

\usepackage{amssymb}
\usepackage{pifont}

\newcommand{\cmark}{\ding{51}}%
\newcommand{\xmark}{\ding{55}}%

\definecolor{RoseQuartzBg}{HTML}{F7CAC9}
\definecolor{ForestGreen}{HTML}{228b22}
\definecolor{RoseQuartz}{HTML}{F5A798}
\definecolor{Serenity}{HTML}{92A8D1}
\definecolor{OrangeRed}{rgb}{1.0, 0.27, 0.0}
\definecolor{Red}{rgb}{1.0, 0.0, 0.0}
\definecolor{forestgreen}{rgb}{0.13, 0.55, 0.13}
\definecolor{Turquoise}{HTML}{0F4C81}
\definecolor{columbiablue}{rgb}{0.61, 0.87, 1.0}
\definecolor{Gray}{gray}{0.9}

\usepackage{xparse}
\usepackage{footmisc}
\NewDocumentCommand{\ying}{ mO{} }{\textcolor{teal}{\textsuperscript{\textit{chao}}\textsf{\textbf{\small[#1]}}}}

\NewDocumentCommand{\lifu}{ mO{} }{\textcolor{blue}{\textsuperscript{\textit{Lifu}}\textsf{\textbf{\small[#1]}}}}

\NewDocumentCommand{\zhiyang}{ mO{} }{\textcolor{Turquoise}{\textsuperscript{\textit{zhiyang}}\textsf{\textbf{\small[#1]}}}}

\newcommand*\modelname{\textsc{Vision-Flan Base}}

\newcommand*\chatmodelname{\textsc{Vision-Flan Chat}}
\newcommand*\llavamodel{LLaVA-Architecture} 
\newcommand*\dataname{\textsc{Vision-Flan}}
\newcommand*\datanum{1,664,261}

\newcommand*\numtasks{187}

\title{Vision-Flan: Scaling Human-Labeled Tasks in Visual Instruction Tuning}

\author{Zhiyang Xu$^{\spadesuit}$ \quad Chao Feng$^{\clubsuit}$ \quad Rulin Shao$^{\heartsuit}$ \quad \textbf{Trevor Ashby}$^{\spadesuit}$ \quad \textbf{Ying Shen}$^{\spadesuit}$ \\ \textbf{Di Jin}$^{\diamondsuit}$ \quad \textbf{Yu Cheng}$^{\vardiamondsuit}$ \quad \textbf{Qifan Wang}$^{\diamondsuit}$ \quad \textbf{Lifu Huang}$^{\spadesuit}$\\
  $^{\spadesuit}$Virginia Tech \quad $^\heartsuit$University of Washington \quad $^{\clubsuit}$University of Michigan \\ $^\diamondsuit$Amazon Inc. \quad $^{\vardiamondsuit}$Microsoft \quad $^{\diamondsuit}$Meta AI \\
  \texttt{\{zhiyangx,lifuh\}@vt.edu}}

\begin{document}
\maketitle
\begin{abstract}
Despite vision-language models' (VLMs) remarkable capabilities as versatile visual assistants, two substantial challenges persist within the existing VLM frameworks: (1) \textit{lacking task diversity} in pretraining and visual instruction tuning, and (2) \textit{annotation error} and \textit{bias} in GPT-4 synthesized instruction tuning data. Both challenges lead to issues such as poor generalizability, hallucination, and catastrophic forgetting.
To address these challenges, we construct \dataname{}, the most diverse publicly available visual instruction tuning dataset to date, comprising \numtasks{} diverse tasks and \datanum{} instances sourced from academic datasets, and each task is accompanied by an expert-written instruction.
In addition, we propose a two-stage instruction tuning framework, in which VLMs are firstly finetuned on \dataname{} and further tuned on GPT-4 synthesized data. We find this two-stage tuning framework significantly outperforms the traditional single-stage visual instruction tuning framework and achieves the state-of-the-art performance across a wide range of multi-modal evaluation benchmarks.
Finally, we conduct in-depth analyses to understand visual instruction tuning and our findings reveal that: 
(1) GPT-4 synthesized data does not substantially enhance VLMs' capabilities but rather modulates the model's responses to human-preferred formats;
(2) A minimal quantity (e.g., 1,000) of GPT-4 synthesized data can effectively align VLM responses with human-preference; 
(3) Visual instruction tuning mainly helps large-language models (LLMs) to understand visual features. 
\end{abstract}    
\section{Introduction}

Recent vision-language models (VLMs)~\cite{liu2023llava,li2023blip,Dai2023InstructBLIPTG}, built upon pre-trained large-language models (LLMs)~\cite{vicuna2023, gao2023llama} and pretrained image encoders~\cite{sun2023eva}, have shown impressive capabilities as general visual assistants. Besides the unimodal encoders, the main ingredients of these VLM frameworks encompass: (1) a bridging module, such as the MLP layers in the LLaVA model~\cite{liu2023llava,li2023blip}, that establishes connections between the pretrained image encoders and LLMs, (2) large-scale text-image pairs~\cite{laion} used for pre-training the bridging module, and (3) GPT-4 synthesized visual instruction tuning datasets~\cite{liu2023llava,li2023otter} to align the responses of VLMs with human preferences (i.e., following users' instruction to generate detailed and helpful responses). Despite their notable successes, we identify two remaining challenges that merit further investigation.

Firstly, the data used in the pre-training stage is dominated by the image captioning task, which lacks diversity, resulting in limited generalizability of VLMs~\cite{infoseek, zhang2023llavar}. For instance, 
the LLaVA model~\cite{liu2023llava} performs poorly on the optical character recognition (OCR) task due to the absence of instances related to text detection during pre-training~\cite{zhang2023llavar}. Several recent studies address this problem by further fine-tuning VLMs on instruction tuning datasets covering more tasks~\cite{zhang2023llavar, hu2023bliva, liu2023improved} such as visual question answering and OCR but the coverage of the tasks is still limited.

\begin{figure*}[ht!]
  \centering
   \includegraphics[width=\textwidth]{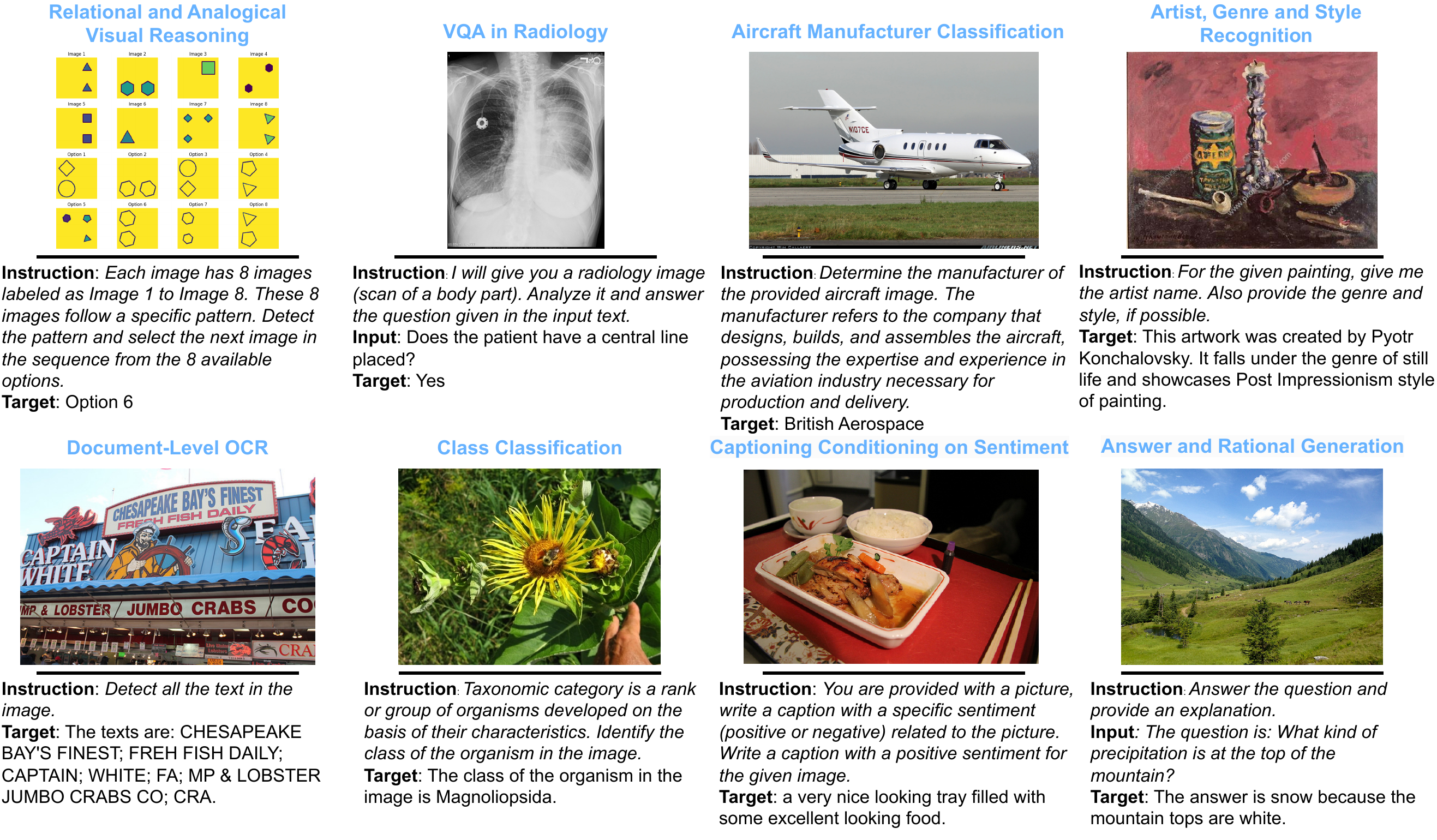}
   \caption{Sample tasks in \dataname{}. \textbf{Instruction} denotes a task instruction crafted by annotators. \textbf{Input} means text input in the given task, and \textbf{Target} is the target response based on the instruction.}
   \label{fig:teaser}
   \vspace{-4mm}
\end{figure*}

Secondly, most of the existing visual instruction tuning datasets~\cite{liu2023llava,li2023otter,yin2023lamm} are synthetically generated via GPT-4 by repurposing text annotations such as captions or dense captions from existing computer-vision datasets to generate new tasks, such as visual dialogue, Complex VQA and detail captions.
While they enable VLMs to generate fluent and detailed responses aligned with human preferences, the lack of task diversity, spurious co-occurring patterns between objects, and long-form outputs may cause severe hallucination~\cite{liu2023robustVIT, li2023evaluating, liu2023hallusionbench, zhou2023hallucination}, and catastrophic forgetting -- VLMs fail to maintain a similar classification performance on basic detection tasks, such as MNIST~\cite{lecun1998mnist} and CIFAR-10~\cite{krizhevsky2009learning}, compared to the zero-shot performance of their vision encoders~\cite{zhai2023investigating}.


To address both challenges, we introduce \dataname{}, the most diverse public-available visual instruction tuning dataset consisting of \numtasks{} tasks drawn from academic datasets, covering \textit{perception} tasks such as object detection and OCR, \textit{domain-specific} tasks such as image-quality classification and image-style classification, \textit{complex reasoning} tasks such as graph understanding and geometric question answering, and many more.
Each task in \dataname{} is accompanied by an expert-written instruction. We show some sample tasks from \dataname{} in Figure \ref{fig:teaser} and provide the full list of tasks in Appendix~\ref{appx:tasks_show}. 

In addition, we introduce a two-stage instruction tuning framework. In the first stage, we utilize the pre-trained LLaVA model~\cite{liu2023llava}
as our initial model, and finetune it on \dataname{} to gain diverse capabilities, resulting in the \modelname{} model. However, due to the concise nature of target outputs in academic datasets, the responses generated by \modelname{} tend to be brief and not aligned with human preferences. Therefore, in the second stage, we further finetune \modelname{} using a minimal amount of GPT-4 synthesized data. This step aims to adjust the model's outputs to be more in line with human preferences, resulting in the \chatmodelname{} model.

Our experimental results demonstrate that high-quality human annotations from \dataname{} significantly enhance the capabilities of both \modelname{} and \chatmodelname{} while reducing the risk of hallucination and catastrophic forgetting. The two-stage instruction tuning framework enables \chatmodelname{} to achieve better human preference alignment using much less GPT-4 synthesized data compared to the state-of-the-art VLMs.
Finally, we perform extensive analysis to understand visual instruction tuning including the roles of human-labeled and GPT-4 synthesized data, and the impacts of various training strategies.
Our investigation yields several key insights:
\begin{itemize}
  \setlength\itemsep{-0.5em}
  \item Increasing the number of human-labeled tasks in visual instruction tuning can substantially enhance VLMs' capabilities across extensive evaluation benchmarks. 
  \item GPT-4 synthesized data does not substantially enhance VLMs capabilities and 
  yields marginal improvements in the VLMs' performance on comprehensive evaluation benchmarks, such as MME~\cite{fu2023mme} and MM-Bench~\cite{liu2023mmbench}.
  \item A minimal quantity (1,000) of GPT-4 synthesized data is sufficient to align VLMs' responses with human preferences. Notably, increasing GPT-4 synthesized data does not correspond to a proportional enhancement in alignment and introduces hallucination and bias into the VLMs.
  \item Visual instruction tuning mainly enhances the ability of large-language models (LLMs) to process and understand visual features. The training of the bridging module, which maps visual features to the embedding space of LLMs, is predominantly achieved during the pre-training phase.
\end{itemize}

\section{\dataname{}}
\begin{figure*}[ht!]
  \centering
   \includegraphics[width=\textwidth]{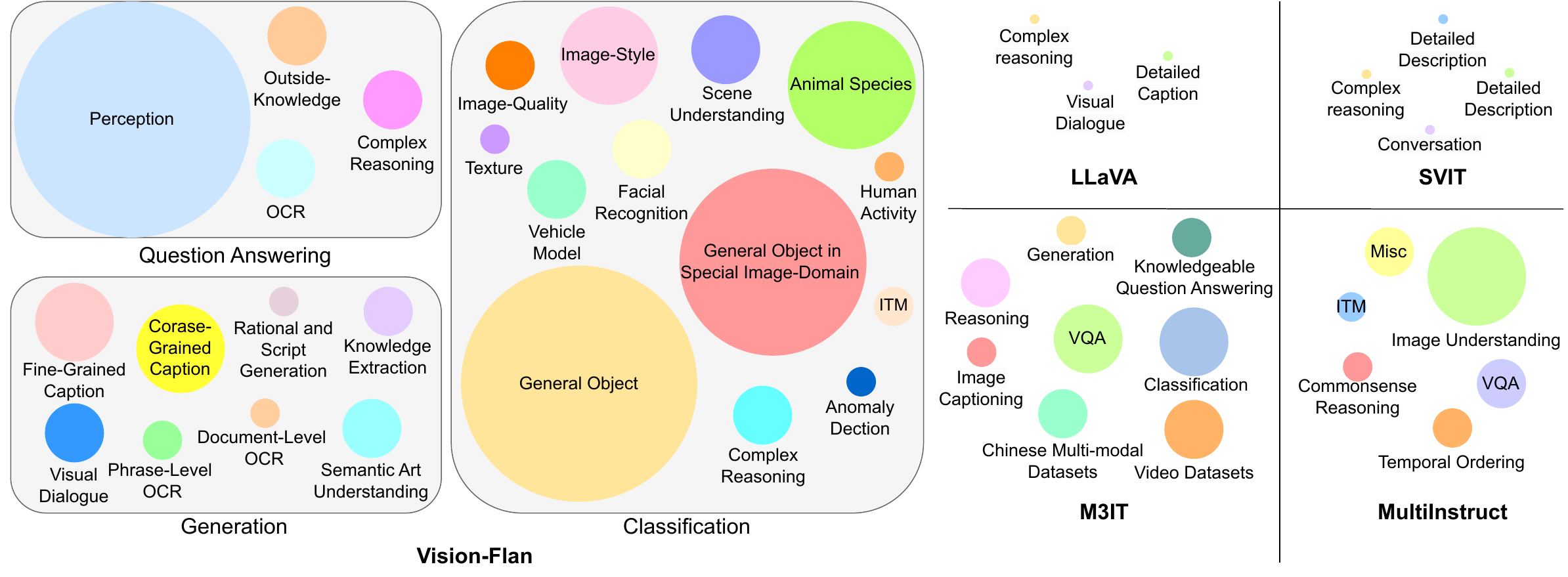}
   \vspace{-6mm}
   \caption{Comparison of task diversity between \dataname{} and previous visual instruction tuning datasets. LLaVA and SVIT report very coarse-grained categories of tasks. Each circle represents a task category and the radius is proportional to the number of tasks in that category. The radius of circles for different datasets are comparable.}
   \label{fig:task_split}
   \vspace{-4mm}
\end{figure*}
\subsection{Collection Pipeline}
We carefully design an annotator selection process to identify qualified annotators, which involves 2 iterations of training and testing. More details of the selection process and compensation can be found in Appendix~\ref{appx:annotator_selection}. In the end, we hire 7 out of 21 candidates as our annotators and all of them are graduate students in computer science. 
To ensure the diversity and quality of the tasks in \dataname, we design a rigorous annotation pipeline with four major steps: 

\paragraph{Existing dataset collection and pre-processing:} Two expert researchers (i.e., senior Ph.D. students in the fields of natural language processing and computer vision) search online and identify high-quality vision-language datasets. The datasets are then equally distributed to 7 annotators to download and preprocess the datasets. Each processed instance consists of an image, an instruction (the task definition from the original dataset with minor modifications), a text input if applicable, and a target output.

\paragraph{Creating new tasks:} The two expert researchers and annotators also discuss potential new tasks that could be derived from the existing annotations. We derive new tasks by combining the annotations of two or more existing tasks on a dataset.
For example, in the Concadia dataset~\cite{kreiss2022concadia}, each instance consists of an image caption and a knowledge snippet related to the image. We propose a new task to predict both the caption and the background knowledge given an image, which is a free-form generation task. The new target output is formed by concatenating the caption with the knowledge snippet. We also develop new tasks by creating more basic versions of the original tasks. For example, given the object detection annotations in MSCOCO~\cite{lin2014microsoft}, we propose an object selection task in which we provide a list of objects and ask the model to select the object that appears in the image (the negative options are created by sampling objects that appear in other images but not in the given image).
The expert researchers and annotators manually solve 20 instances for each newly developed task.  
If the human predictions match the target outputs, this new task is considered valid.


\paragraph{Iteratively refining the task instructions and output templates:} For existing tasks, we ask annotators to write instructions based on the original task definitions with minor modifications. For newly developed tasks, the annotators write instructions by discussing with the expert researchers.
Once an annotator finishes writing a new instruction, one of the two expert researchers is randomly assigned to examine the instances and provide feedback for revising the instruction. This step iterates repeatedly until the instruction meets our requirements. We require the instruction to be \textit{clear}, \textit{easy to understand}, and can \textit{be correctly executed by a human}.
Each task together with its associated dataset and instruction is then added to the pool of candidate tasks for \dataname{}. 

\paragraph{Verifying the quality of each task:} From the candidate task pool, two expert researchers, including a native English speaker, work together to select the high-quality tasks where the instruction is fluent and effectively conveys the intended task and the task does not overlap with other tasks.

Based on these four steps, we finally collect \numtasks{} high-quality tasks, and for each task, we randomly sample 10,000 instances from its corresponding dataset. If a dataset contains less than 10,000 instances, we include all of them. We name the dataset as \dataname{}, consisting of \datanum{} instances for \numtasks{} tasks in total. We include references to all the datasets used in \dataname{} in Appendix~\ref{appx:dataset_used_in_vision_flan} and show an instance for each task in Appendix~\ref{appx:tasks_show}.
\subsection{Comparison with Existing Datasets}
\begin{table}[ht]
  \centering
  \resizebox{\linewidth}{!}{%
  \begin{tabular}{l c c c c c c c c c c }
  \toprule
   \textbf{Dataset} & Instances \# & Tasks \# & Source \\
   %
  \midrule
  LLaVA~\citep{liu2023llava} & 150K & 3  & Synthetic \\
  LAMM~\citep{yin2023lamm} & 196K & 8   & Synthetic \\
  VL-Qwen~\citep{qwen} &350K & Unknown   & Private\\
  M$^{3}$IT~\citep{li2023m} & 2.4M & 40  & Synthetic\\
  mPlug-Owl~\cite{ye2023mplug} & 150K & 3 & Synthetic \\
  Shikra~\citep{chen2023shikra} & 156K & 4 & Synthetic \\
  SVIT~\citep{zhao2023svit} & 4.2M & 4  & Synthetic \\
  MultiInstruct~\cite{xu-etal-2023-multiinstruct} & 510K & 62  & Public \\
  \dataname{} (Ours) & 1.6M & 196 & Public \\
  \bottomrule
  \end{tabular}}
  \caption{Comparison between \dataname{} and existing visual instruction tuning datasets.}
  \label{table:vision-flan-compa}
  \vspace{-4mm}
  \end{table}

Table~\ref{table:vision-flan-compa} presents a comparison between existing visual instruction tuning datasets and \dataname{}. For existing visual instruction tuning datasets, we directly adopt the numbers of tasks and instances reported in their original papers. The majority of these datasets are generated using proprietary language models, such as ChatGPT\footnote{\url{https://openai.com/blog/chatgpt}} and GPT-4\footnote{\url{https://openai.com/research/gpt-4}}, and exhibit a narrow range of task diversity. VL-Qwen~\cite{qwen} is a recently introduced large-scale dataset annotated by humans but remains inaccessible to the public. Although MultiInstruct~\cite{xu-etal-2023-multiinstruct} is based on publicly available datasets, it mainly focuses on visual grounding tasks and only contains 29 tasks that do not involve region-specific information. In contrast, \dataname{} encompasses a significantly more diverse array of tasks, offering a three-times increase compared to the number of tasks in MultiInstruct. 

In Figure~\ref{fig:task_split}, we compare the task categories covered by \dataname{} and other datasets.
Tasks within \dataname{} are first categorized into three primary groups: \textit{Question Answering}, \textit{Classification}, and \textit{Generation}, and each of these primary groups is further divided into specific, fine-grained categories. For instance, within the \textit{Classification} group, the \textit{General Object} category involves classifying objects in images into various concepts, such as ``fish'', ``car'', and ``dog''. Contrastingly, the \textit{Vehicle Model} category demands the models to accurately identify specific car brands or models, like ``Toyota'' and ``Camry''. 
The visualization in Figure~\ref{fig:task_split} clearly demonstrates the superior diversity and volume of tasks in \dataname{} compared to existing datasets. We list tasks in each category in Appendix \ref{appx:task_group_in_vision_flan}.

\section{\dataname{} Finetuning}
\begin{figure}[ht!]
  \centering
   \includegraphics[width=\linewidth]{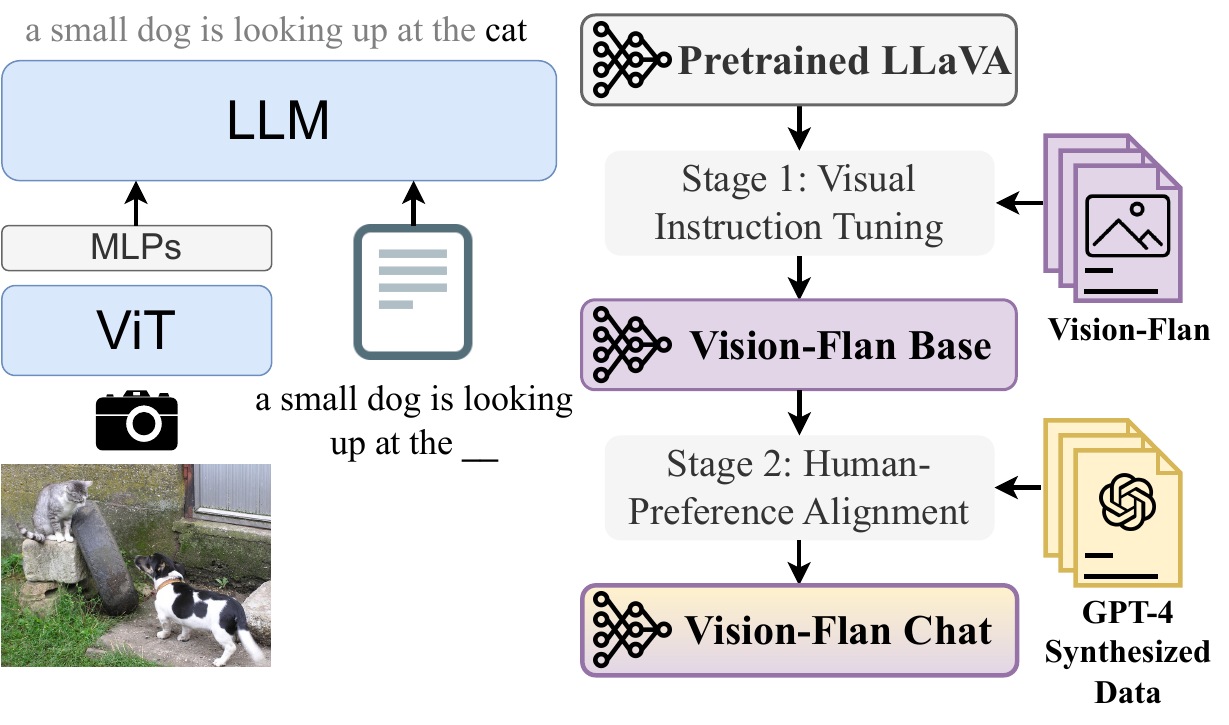}
   \vspace{-6mm}
   \caption{The left of the figure shows the LLaVA-Architecture and the right of the figure shows the two-stage visual instruction tuning pipeline.}
   \label{fig:method}
   \vspace{-6mm}
\end{figure}

\paragraph{Model Architecture} We adopt the same VLM architecture as LLaVA~\cite{liu2023improved} and denote it as \llavamodel{}. As shown in Figure~\ref{fig:method}, it consists of a pre-trained vision encoder, a pre-trained large language model, and two layers of MLPs to connect them. In the vision-language pre-training phase of the \llavamodel{}, both the pre-trained vision encoder and large language model remain frozen, and only the MLP layers are trained on a large-scale image captioning dataset~\cite{laion}. We leverage this pre-trained LLaVA model
, without any visual instruction tuning, as our initial model and finetune it on \dataname{}. During visual instruction tuning, we finetune both the MLP layers and the language model while keeping the vision encoder frozen.

\paragraph{Two-stage Visual Instruction Tuning}
Contrary to prior approaches~\cite{liu2023improved,Dai2023InstructBLIPTG} that mix human-labeled data with GPT-4 synthesized data for visual instruction tuning, our study introduces a two-stage instruction tuning pipeline. As shown in Figure~\ref{fig:method}, in the first stage, we finetune the VLM on all \numtasks{} tasks of \dataname{} to acquire diverse capabilities and name the resulting model as \modelname{}. However, due to the brevity of target outputs presented in academic datasets, the responses from \modelname{} are not in human-preferred formats. Hence, we further finetune \modelname{} on GPT-4 synthesized data to align the model's outputs with human preference. We denote the yielded model as \chatmodelname{}. This training framework requires minimal GPT-4 synthesized data while providing deep insights into the distinct contributions of human-labeled and GPT-4 synthesized data in visual instruction tuning.
\paragraph{Implementation Details} We leverage \llavamodel{} with Vicuna-13B v1.5~\cite{vicuna2023}, CLIP-ViT-L-336px~\cite{radford2021learning} and two layers of MLP as our VLM. For the first-stage instruction tuning, we finetune the MLP layers and the language model on \dataname{} for 1 epoch with a learning rate 2e-5 and per device batch size 16 on 8 A100 GPUs. For the second-stage instruction tuning, we further finetune the MLP layers and the language model on 1,000 instances randomly sampled from the LLaVA dataset~\cite{liu2023llava} with learning rate 1e-5 and per device batch size 8 on 8 GPUs for 128 steps. In the following sections, we use LLaVA dataset and GPT-4 synthesized data interchangeably.
\section{Experiment Setup}

\paragraph{Evaluation Datasets} We evaluate the models on several widely adopted multimodal evaluation benchmark datasets including \textit{multiple-choice} benchmarks: \textbf{MMbench}~\cite{liu2023mmbench}, \textbf{MME}~\cite{fu2023mme}, and \textbf{MMMU}; \textit{free-form generation} benchmarks: \textbf{MM-Vet}~\cite{yu2023mm} and \textbf{LLaVA-Bench}; the \textit{hallucination} benchmark: \textbf{POPE}~\cite{li2023evaluating}, and \textit{catastrophic forgetting} benchmarks: \textbf{CIFAR-10 and CIFAR-100}~\cite{krizhevsky2009learning}, \textbf{MNIST}~\cite{lecun1998mnist}, and \textbf{miniImageNet}~\cite{vinyals2016matching}. More details of the evaluation datasets can be found in Appendix \ref{sec:appendix_eval_data}.
\paragraph{Evaluation Protocols} For MMbench, MME, MM-Vet,  LLaVA-Bench, POPE and MMMU, we strictly follow their official implementations of evaluation code to evaluate the performance of each model. 
For datasets that do not have official evaluation codes including CIFAR-10, CIFAR-100, MNIST, and miniImageNet, we leverage the state-of-the-art open-source LLM, Vicuna 1.5 13B, to perform the evaluation and report the averaged performance on these four datasets in the CF column in Table \ref{table:main_results}. More details of evaluation protocols can be found in Appendix \ref{sec:appendix_eval_protocol}.
\paragraph{Baselines} We compare our models with several recent state-of-the-art vision-language models, including \textbf{BLIP-2}~\cite{li2023blip}, \textbf{InstructBLIP}~\cite{Dai2023InstructBLIPTG},
\textbf{Shikra}~\cite{chen2023shikra}, \textbf{LLaVA}~\cite{liu2023llava}, \textbf{Qwen-VL}, \textbf{Qwen-VL-Chat}~\cite{bai2023qwenvl}, and  \textbf{LLaVA-1.5}~\cite{liu2023improved}. The LLMs and image encoders used in all baselines are shown in Table \ref{table:main_results}. Details of baselines can be found in Appendix \ref{sec:appendix_baseline}.

\section{Results and Analysis}

\begin{table*}[t]
  \centering
  \resizebox{\textwidth}{!}{%
  \begin{tabular}{l c c c c c c c c c c c }
  \toprule
   \textbf{Model} & LLM &Image Encoder & MM-Bench & MME & MMMU & LLaVA-Bench & MM-Vet& Pope & CF\\
   %
  \midrule
  BLIP-2 &FlanT5-XXL &ViT-g/14 & - &1293.8 & 34.0& - & 22.4& 85.3& -\\
  InstructBlip &Vicuna-13B &ViT-g/14  & 36.0 & 1212.8 & 33.8 & 58.2 &25.6 & 78.9& - \\
  Mini-GPT4 &Vicuna-13B &ViT-g/14 & 24.3 & 581.67 & 27.6  & - & - & - & -\\
  Shikra &Vicuna-13B  &ViT-L/14   & 58.8 & - & - & -&  - & -& -\\
  LLaVA &Vicuna-13B v1.5  &CLIP-ViT-L-336px  & 38.7 & 1151.6 & -  &70.8 & 33.4 & 75.3 & - \\
  Qwen-VL &Qwen-7B& ViT-bigG & 38.2 & - & - & -  & - &  -  & -\\
  Qwen-VL-Chat &Qwen-7B& ViT-bigG  & 60.6 & 1487.5 &32.9  & \underline{73.6} & - & - & 72.1 \\
  LLaVA 1.5 &Vicuna-13B v1.5  &CLIP-ViT-L-336px   & 66.7 & \underline{1531.3} & 33.6  & 70.7 & \underline{35.4} & 83.6 & 73.3 \\
  \midrule
  \midrule
  \modelname{} &Vicuna-13B v1.5  &CLIP-ViT-L-336px & \bf{69.8} & \bf{1537.8} & \bf{34.4}  & 38.5 & 33.4& \underline{85.9} & \bf{87.2} \\
    \midrule
    
    \rowcolor{Gray}\multicolumn{10}{l}{\textbf{Second-Stage Tuning with 1,000 GPT-4 Synthesized Instances} \hfill}   \\
    \chatmodelname{} &Vicuna-13B v1.5  &CLIP-ViT-L-336px  & \underline{67.6} & 1490.6 & \underline{34.3}  & \bf{78.3} & \bf{38.0} & \bf{86.1} & \underline{84.0} \\
  \bottomrule
  \end{tabular}}
  \caption{Comprehensive evaluation of VLMs on widely adopted benchmark datasets. 
 CF denotes the averaged performance of VLMs on four catastrophic forgetting benchmarks.}
  \label{table:main_results}
  \vspace{-5mm}
  \end{table*}

\subsection{Main Results}

As demonstrated in Table \ref{table:main_results}, \modelname{} achieves state-of-the-art performance on comprehensive evaluation benchmarks including MME, MM-Bench and MMMU, while reducing hallucination and catastrophic forgetting. However, we observe that  \modelname{} scores significantly lower on the LLaVA-Bench dataset in comparison to VLMs trained using GPT-4 synthesized data. We attribute this discrepancy to the conciseness and brevity of target outputs within academic datasets. As shown in Figure~\ref{fig:teaser}, VQA tasks frequently yield outputs comprising a single or a few words. Even outputs of many generation tasks are typically confined to one or two succinct sentences. Training on these tasks leads \modelname{} to generate brief responses, which are not aligned with human preferences.
Conversely, through the second-stage tuning on a mere 1,000 GPT-4 synthesized data instances, \chatmodelname{} achieves
significant performance improvement on LLaVA-Bench, a benchmark measuring human-preference alignment, 
while maintaining a relatively lower rate of hallucination and catastrophic forgetting. To better understand why \dataname{} models are better than current VLMs, we conduct two case studies focusing on OCR and Entity Recognition and analyze both quantitative and qualitative results in Appendix~\ref{appx:why_vision_flan_better}.

Another finding in Table~\ref{table:main_results} is that compared to \modelname{}, \chatmodelname{} achieves slightly inferior performance on comprehensive evaluation benchmarks demonstrating the bias and hallucination inevitably introduced by the GPT-4 synthesized data, which is discussed in detail in Section~\ref{human_vs_gpt4}.

\subsection{Effect of Human-Labeled and GPT-4 Synthesized Datasets}\label{human_vs_gpt4}
\begin{figure}[h!]
  \centering
   \includegraphics[width=0.8\linewidth]{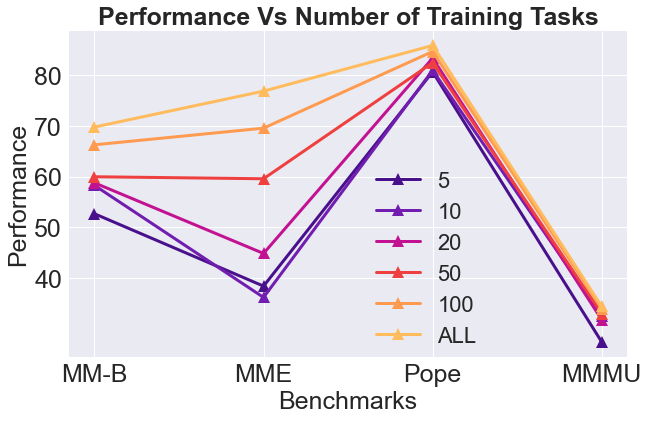}
   \caption{Performance on four comprehensive benchmarks versus the number of training tasks.}
   \label{fig:num_training_instance}
   \vspace{-5mm}
\end{figure} 
\paragraph{Effect of Task Diversity in \dataname{}}
Figure \ref{fig:num_training_instance} illustrates the relationship between the number of tasks from \dataname{} employed during visual instruction tuning and the performance of \modelname{} across four comprehensive evaluation benchmarks. It's apparent that as the number of tasks increases, the performance of \modelname{} on all datasets is improved. To evaluate the impact of varying numbers of instances from different tasks, we fix the total amount of instances used for visual instruction tuning and experiment with different numbers of tasks. As demonstrated in Table~\ref{table:task_diversity}, when the number of training instances is constant, augmenting the number of tasks significantly enhances model performance.
These findings substantiate our hypothesis that \textit{the diverse array of human-labeled tasks within \dataname{} is essential for improving the capabilities of VLMs}.
\begin{table}[h!]
  \centering
  \resizebox{\linewidth}{!}{%
  \begin{tabular}{l c c c c c c c c c c }
  \toprule
     \# of Tasks & \# of Instances per Task & MMB & MME & Pope & MMMU \\
     \rowcolor{Gray}\multicolumn{6}{l}{\textbf{Training with 100,000 Instances} \hfill}   \\
     10 & 10,000 & 58.3 & 723.9 & 81.0 & 32.6 \\
     \numtasks{} & 500 & 58.8  & 1314.3 & 83.3 & 33.3 \\
     \rowcolor{Gray}\multicolumn{6}{l}{\textbf{Training with 200,000 Instances} \hfill}   \\
     20 & 10,000 & 58.8 & 897.3 & 83.4 & 31.8 \\
     \numtasks{} & 1,000 & 63.5  & 1373.5 & 83.6 & 33.7\\
   %
  \bottomrule
  \end{tabular}}
  \caption{Comparison of \modelname{} trained with a fixed total amount of data instances.
  }
  \label{table:task_diversity}
  \vspace{-5mm}
  \end{table}



\begin{figure}[ht!]
  \centering
   \includegraphics[width=0.85\linewidth]{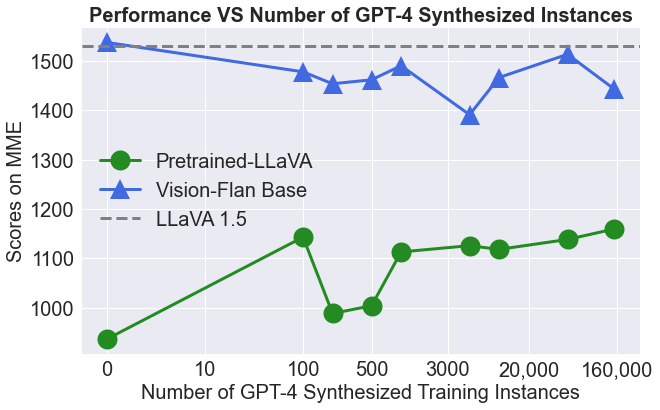}
   \caption{Effect of the number of GPT-4 synthesized training instances on MME. The \textcolor{gray}{dashed gray} line indicates the performance of LLaVA 1.5.}
   \vspace{-4mm}
   \label{fig:num_llava_vs_mme}
\end{figure}

\paragraph{Effect of GPT-4 Synthesized Data on Comprehensive Evaluation Benchmarks} Furthermore, we analyze if GPT-4 synthesized data can improve the model's performance on comprehensive evaluation benchmarks and show the results in Figure \ref{fig:num_llava_vs_mme}. Further tuning \modelname{} on GPT-4 synthesized data instances does not lead to performance improvement. Tuning pretrained LLaVA model on a small amount of GPT-4 synthesized data (100) can improve its performance on MME but further increasing the number of training instances does not lead to any improvement. We also observe a similar trend on the MM-Bench dataset and report the result in Appendix~\ref{appx:num_llava_vs_mmbench}. These observations are in line with recent findings in LLMs: 
\textit{GPT-4 synthesized data does not improve model's capability but rather modulates the responses towards human-preferred formats} \cite{jain2023analyzeAlignment,gudibande2023false}.

\begin{figure}[ht!]
  \centering
   \includegraphics[width=0.85\linewidth]{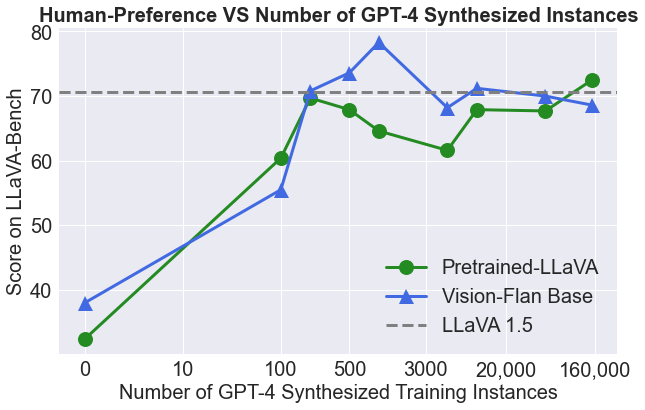}
   \caption{Effect of the number of GPT-4 synthesized instances on human preference alignment. The \textcolor{gray}{dashed gray} line indicates the performance of LLaVA 1.5.}
   \vspace{-4mm}
   \label{fig:num_llava_vs_llava}
\end{figure}

\paragraph{Effect of GPT-4 Synthesized Data on Human-Preference Alignment}
When utilizing our proposed two-stage tuning framework, 
we find that by performing a second-stage finetuning on a mere 1,000 GPT-4 synthesized instances from the LLaVA dataset, \chatmodelname{} achieves significantly better performance (78.5 v.s. 38.5) on the LLaVA-Bench dataset. This observation leads us to raise the question: \textit{Is extensive finetuning on large-scale GPT-4 synthesized datasets necessary for aligning VLMs with human preferences?} To answer it, we finetune both \modelname{} and pretrained LLaVA model on different numbers of GPT-4 synthesized instances ranging from 100 to 158,000, and show the results in Figure~\ref{fig:num_llava_vs_llava}. As we can see, with 1,000 instances, \modelname{} achieves a score of 78.3 and further increasing the number of training instances leads to a performance drop. A similar trend can also be seen for the pretrained LLaVA model.

\begin{figure}[ht!]
  \centering
   \includegraphics[width=0.8\linewidth]{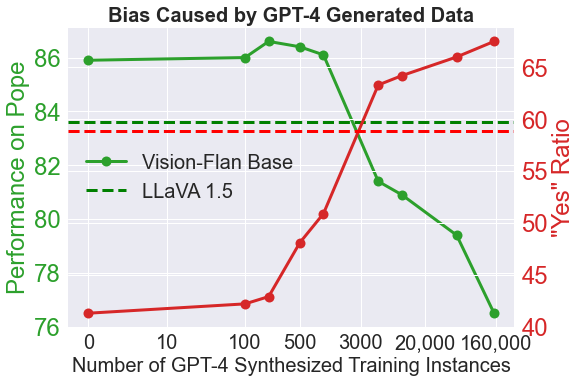}
   \caption{Effect of the number of GPT-4 synthesized training instances on the \textcolor{ForestGreen}{hallucination benchmark} and the \textcolor{red}{ratio of ``Yes''}. The dashed lines indicate the performance of the state-of-the-art LLaVA 1.5 model.}
   \vspace{-6mm}
   \label{fig:pope_bias}
\end{figure}

\paragraph{GPT-4 Synthesized Data Causes Hallucination and Bias}
Concurrent work~\cite{liu2023robustVIT} identifies that hallucination in current VLMs can be caused by their bias toward positive answers (i.e., ``Yes''). In Figure~\ref{fig:pope_bias}, we explicitly show the relationship between hallucination, the ratio of ``Yes'', and the number of training instances from GPT-4 synthesized dataset. As the number of GPT-4 synthesized instances increases, the model's responses are biased towards the answer ``Yes'' even if the objects are not in the images, causing the model to hallucinate. This observation suggests that a small amount of GPT-4 synthesized training instances is preferred to avoid hallucination and bias in VLMs.

\subsection{Single-stage Tuning on Mixed Data Vs. Two-stage Tuning}
In this section, we compare the performance of two training strategies based on the same pretrained LLaVA model: (1) finetuning it on the mix of \dataname{} and the LLaVA dataset; (2) finetuning it utilizing \dataname{} and 1,000 instances from the LLaVA dataset with our two-stage tuning method. As illustrated in Table~\ref{table:mixed_vs_two_stage}, the performance of VLMs finetuned on the mix of \dataname{} and GPT-4 synthesized data is notably inferior compared to \chatmodelname{} trained through our two-stage tuning framework.
\begin{table}[ht!]
  \centering
  \resizebox{\linewidth}{!}{%
  \begin{tabular}{l c c c c c c c c c c }
  \toprule
    Method & \# of LLaVA & MME & LLaVA-Bench & MM-Vet  \\ \midrule
     Mixed Data & 1,000 &1364.0 & 52.7 & 36.6  \\
     Mixed Data & 158,000 & 1317.9 & 63.9 & 36.8 \\
     Two-stage & 1,000 & 1490.6 & 78.3 & 38.0 \\
   %
  \bottomrule
  \end{tabular}}
  \caption{
  Comparison between single-stage finetuning on mixed data and two-stage finetuning. 
  }
  \label{table:mixed_vs_two_stage}
  \vspace{-6mm}
  \end{table}
  
\subsection{What is Essentially Improved in VLMs during Visual Instruction Tuning}
\label{what_improved_visual_instruction_tuning}
\begin{table}[ht]
  \centering
  \resizebox{\linewidth}{!}{%
  \begin{tabular}{l c c c c c c c c c c }
  \toprule
     LLM & MLPs & MM-Bench & MME & LLaVA-Bench & Pope \\
     \midrule
     \textcolor{red}{\xmark} & \textcolor{red}{\xmark} & 45.0 & 936.3 & 32.4 & 51.9 \\
    \textcolor{red}{\xmark} & \textcolor{ForestGreen}{\cmark} & 52.4 &1107.3 & 39.1 & 83.3\\
    \textcolor{ForestGreen}{\cmark} & \textcolor{red}{\xmark} & 69.2 & 1495.5 & 39.3 & 85.6\\
    \textcolor{ForestGreen} \cmark & \textcolor{ForestGreen}{\cmark} & 69.8 & 1537.8 & 38.5 & 85.9\\
   %
  \bottomrule
  \end{tabular}}
  \caption{Effect of tuning different modules in \modelname{}. \textcolor{ForestGreen}{\cmark} denotes the module is tuned and \textcolor{red}{\xmark} denotes the module is frozen during visual instruction tuning.}
  \vspace{-5mm}
  \label{table:module_comp}
  \end{table}
In \llavamodel{}, the MLP layers map the visual features from a vision encoder into the embedding space of LLMs. The LLMs then interpret the visual features and follow text instructions to generate responses. In Table~\ref{table:module_comp}, we show the results of training different modules during visual instruction tuning and observe that solely tuning MLPs causes a significant performance drop compared to tuning both MLPs and LLMs during visual instruction tuning. However, tuning LLMs with frozen MLPs results in similar performance as tuning both modules, demonstrating that visual instruction tuning mainly enables LLMs to better understand visual features while MLPs have been sufficiently learned during pretraning. To further support this claim, we replace the instruction-tuned MLPs in \modelname{} and \chatmodelname{} with the pretrained MLPs from the pre-trained LLaVA model, and show that with the pretrained MLPs, both models can retain more than 90\% of performance on most tasks as shown in Table~\ref{table:mlp_results}. We also compute the Pearson Correlation Coefficient between the parameters of pretrained MLPs and instruction-tuned MLPs, and find that their correlation coefficient is higher than 0.99.

\begin{table}[ht]
  \centering
  \resizebox{\linewidth}{!}{%
  \begin{tabular}{l c c c c c c c c c}
  \toprule
   \textbf{Model}  & MMB & MME & LLaVA-Bench & Pope\\
  \midrule
  \rowcolor{Gray}\multicolumn{1}{l}{\bf{\modelname{}}} & 69.8 & 1537.8 & 38.5 & 85.9 \\
  \rowcolor{yellow}\multicolumn{1}{l}{+ Pretrained MLP} & 68.0 & 1403.1 & 36.4  & 84.0 \\
  \rowcolor{Gray}\multicolumn{1}{l}{\bf{\chatmodelname{}}} & 67.6 & 1490.6 &78.3 & 86.1\\
  \rowcolor{yellow}\multicolumn{1}{l}{+ Pretrained MLP} & 65.7 &1332.2  & 73.8  & 85.4 \\
  \bottomrule
  \end{tabular}}
  \caption{Results of replacing visual instruction tuned MLPs with pretrained MLPs. \textcolor{gray}{Gray rows} show the performance of the original models and \textcolor{yellow}{yellow rows} show the performance after replacing instruction-tuned MLPs with pretrained MLPs.}
  \label{table:mlp_results}
  \vspace{-5mm}
  \end{table}

\vspace{-2mm}
\section{Related Work}
\vspace{-2mm}
Instruction tuning~\cite{wei2021finetuned} is first introduced in NLP and has been adapted to the visual-language domain.
MultiInstruct~\cite{xu-etal-2023-multiinstruct} propose the first human-label multi-modal instruction tuning dataset for improving the zero-shot performance of pre-trained VLMs. LLaVA~\cite{liu2023llava} leverage GPT-4 to repurpose text annotations such as captions or dense captions from existing computer-vision datasets to generate visual dialogues, Complex VQA and detail captions for visual instruction tuning. Following LLaVA, mPLUG-Owl~\cite{ye2023mplug}, LAMM~\cite{yin2023lamm}, MIMIC-IT~\cite{mimic} and Macaw-LLM~\cite{lyu2023macaw} leverage proprietary LLMs such as GPT-4 and ChatGPT to further extend the instruction tuning tasks into 3D-domain, multiple-images and videos, and increase the amount of training instances. MiniGPT-4~\cite{zhu2023minigpt} utilizes ChatGPT to refine output from the pre-trained VLM itself. InstructBLIP~\cite{Dai2023InstructBLIPTG} and LLaVA-1.5~\cite{liu2023improved} mix the human-annotated and GPT4 synthesized datasets to enhance visual instruction tuning. 

Several recent work explores different strategies to improve visual instruction tuning. StableLLaVA~\cite{li2023stablellava} and VPG-C~\cite{Li2023FinetuningML} generate both images and texts using Stable Diffusion~\cite{rombach2022high} or Blended Diffusion~\cite{avrahami2022blended} to alleviate domain bias and encourage VLMs attend to visual details. \cite{liu2023aligning} demonstrate the bias introduced by positive instructions and introduce negative instruction examples for improving robustness. Shikra~\cite{chen2023shikra} incorporate visual grounding tasks in visual instruction tuning to improve the VLM's referential capability. LLaVAR~\cite{zhang2023llavar} and BLIVA~\cite{hu2023bliva} leverage OCR tools and GPT-4 to generate tasks helping VLMs to understand text in images. \cite{lu2023empirical} and SVIT~\cite{zhao2023svit} empirically study the effect of scaling the size of VLMs and the size of GPT-4 synthesized dataset. Two concurrent works~\cite{prompt_GPT4V, ShareGPT4V} directly prompt GPT-4V with images as input to generate visual instruction tuning data and achieve superior performance. Additional related work can be found in Appendix~\ref{appx:additional_related_work}.

Unlike all prior work, our work mainly focuses on scaling human-labeled tasks in visual instruction tuning to improve VLMs' capabilities. Additionally, we perform extensive analysis to understand the characteristics of human-labeled and GPT-4 synthesized data and draw meaningful conclusions.

\vspace{-2.5mm}
\section{Conclusion}
\vspace{-3mm}
We construct \dataname{}, the most diverse public-available visual instruction tuning dataset, consisting of \numtasks{} diverse tasks and \datanum{} instances collected from academic datasets, and each task is accompanied by an expert-written instruction. We demonstrate that VLMs trained on \dataname{} with proposed two-stage tuning framework achieve state-of-the-art performance on comprehensive evaluation benchmarks. Additionally, we perform extensive analysis and reveal the distinct contributions of human-labeled and GPT-4 synthesized data in visual instruction tuning.

\section{Limitations}
All the tasks included in \dataname{} are in English, which confines the usage of our dataset and models to English speaking populations. Future work should extend \dataname{} with multi-lingual tasks. In addition, all the tasks in \dataname{} only consists of a single image. Many real-world vision-language tasks require the model to take multiple images as inputs. Thus, future work should explore vision-language tasks that involve multiple images or videos.

Our analysis mainly focuses on the GPT-4 synthesized visual instruction tuning dataset. Recently, as the GPT-4V~\footnote{\url{https://openai.com/research/gpt-4v-system-card}} becomes publicly available, there are some concurrent works~\cite{prompt_GPT4V, ShareGPT4V} prompting GPT-4V with images as inputs to generate visual instruction tuning data. Future work can analyze the effect of tuning VLMs on such datasets and identify the advantages and disadvantages.

In our experiments, we mainly focus on the \llavamodel{}~\cite{liu2023llava}  due to its strong performance and high efficiency. However, there are other foundation architectures such as Q-former in BLIP2~\cite{li2023blip} and Perceiver Resampler in Flamingo~\cite{alayrac2022flamingo}. More diverse VLM architectures can be explored in the future to draw more general conclusions.

\bibliography{anthology,custom}

\appendix
\section{More Details on the Annotation Process of \dataname{}}
\subsection{Annotator Selection}
\label{appx:annotator_selection}
Due to the complexity of the annotation task, we carefully design a selection process to select qualified annotators. Specifically, at beginning, the authors send out emails looking for graduate students in computer science who are interested in NLP and multi-modal learning. A group of 21 graduate computer science students signed up for a tutorial section. In the tutorial section, two PhD students in NLP explain the requirements for writing instructions, downloading the dataset and processing raw datasets into a unified format. After the tutorial, each candidate is assigned with three datasets and they have totally three days to process the raw datasets and write instructions. In the end, each candidate submits their annotations and two PhD students provide feedback to each candidate. The candidates then have two days to modify their instructions or formats based on the feedback. After two days, the candidates submit their final version of annotations and two PhD students discuss the quality of the annotations case by case. In the end, 7 out of 21 students were selected as qualified annotators. The compensation is 15\$ per hour.

\section{Evaluation Datasets}
\label{sec:appendix_eval_data}
We evaluate our models on several widely used multimodal evaluation benchmark datasets: (1) \textbf{MMbench}~\cite{liu2023mmbench} is a comprehensive evaluation benchmark measuring VLM's capabilities from 20 dimensions. (2)\textbf{MME}~\cite{fu2023mme} measures VLM's perception and cognition capabilities based on 14 diverse tasks. (3) \textbf{MM-Vet}~\cite{yu2023mm} focuses on measuring the integration of various capabilities of VLMs, including OCR, recognition, knowledge, spatial awareness, math, and language generation. (4) \textbf{LLaVA-Bench}~\cite{liu2023llava} evaluates the instruction following and chat ability of VLMs in diverse daily-life visual tasks. (5) \textbf{POPE}~\cite{li2023evaluating} is an evaluation benchmark that probes object hallucination in VLMs. (6) \textbf{MMMU}~\cite{yue2023mmmu} evaluates VLMs on multi-discipline tasks that require college-level subject knowledge and deliberate reasoning.

We also evaluate the newly proposed catastrophic forgetting problem~\cite{zhai2023investigating} of VLMs on 4 datasets: \textbf{CIFAR-10 and CIFAR-100}~\cite{krizhevsky2009learning}, \textbf{MNIST}~\cite{lecun1998mnist}, and \textbf{miniImageNet}~\cite{vinyals2016matching}. We report the averaged performance of VLMs on the four benchmarks in the CF column of Table~\ref{table:main_results}.

\section{Evaluation Protocols}
\label{sec:appendix_eval_protocol}
For MM-Bench, MME, MM-Vet,  LLaVA-Bench, POPE and MMMU, we use their official implementations of evaluation code\footnote{\url{https://github.com/BradyFU/Awesome-Multimodal-Large-Language-Models/tree/Evaluation}\\ \url{https://mmbench.opencompass.org.cn/leaderboard}\\ \url{https://github.com/yuweihao/MM-Vet}\\ \url{https://github.com/haotian-liu/LLaVA/blob/main/docs/LLaVA_Bench.md}\\ \url{https://github.com/RUCAIBox/POPE}\\
\url{https://github.com/MMMU-Benchmark/MMMU}} to evaluate the performance. Specifically,  the evaluation scripts of MM-bench and MM-Vet call GPT-4 API to evaluate the correctness of a prediction given the target output and produce a binary score (0 or 1). Similarly, the evaluation of LLaVA-Bench also leverages GPT-4, and in addition to the target outputs, the evaluation method considers detail descriptions of images. The evaluation results are scores indicating not only the correctness but the human-preference of the predictions. MME and POPE are binary classification tasks and their evaluation is based on string matching between the predictions and target labels.

\section{Baselines}
\label{sec:appendix_baseline}
We compare our method with recent vision-language models. All the baselines listed below have similar architectures which consist of a pretrained LLM, a pretrained image encoder, and a bridging module that connects them. 
\textbf{BLIP-2}~\cite{li2023blip} utilizes the Q-Former to bridge a pretrained image encoder with a pretrained LLM, and achieves strong zero-shot capabilities. 
\textbf{InstructBLIP}~\cite{Dai2023InstructBLIPTG} is a visual-instruction-tuned BLIP-2~\cite{li2023blip} model. The instruction tuning dataset is a mixture of 13 academic datasets and the LLaVA~\cite{liu2023llava} dataset.
\textbf{Shikra}~\cite{chen2023shikra} focuses more on the object grounding capability and is instruction tuned on referential dialogue dataset and LLaVA dataset~\cite{liu2023llava}, both of which are synthetically generated via GPT-4. 
\textbf{LLaVA}~\cite{liu2023llava} is the first VLM finetuned on GPT-4 synthesized visual instruction tuning dataset and achieves remarkable performance as a general-purpose visual chatbot. 
\textbf{Qwen-VL} and \textbf{Qwen-VL-Chat}~\cite{bai2023qwenvl} are recently proposed VLMs based on Qwen~\cite{qwen} language model and are trained on a large-scale (50 million instances) private visual instruction tuning dataset.
\textbf{LLaVA-1.5}~\cite{liu2023improved} is a LLaVA model trained on a mixture of shareGPT~\footnote{\url{https://sharegpt.com/}}, LLaVA~\cite{liu2023llava} and 8 academic image-text datasets.

\section{Additional Results}

    

\subsection{Effect of GPT-4 synthesized data on comprehensive evaluation benchmarks}
\label{appx:num_llava_vs_mmbench}

\begin{figure}[ht!]
  \centering
   \includegraphics[width=\linewidth]{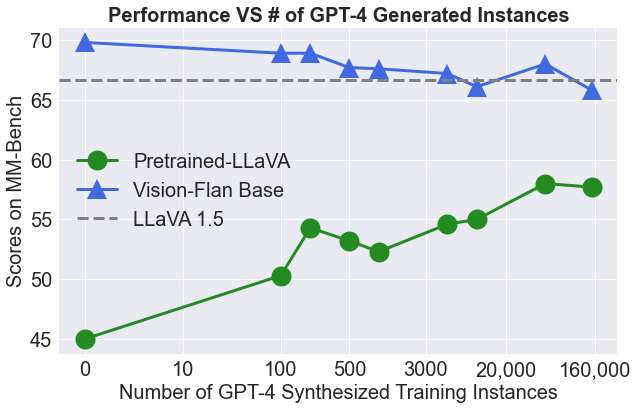}

   \caption{Effect of increasing the number of GPT-4 synthesized training instances on the comprehensive evaluation benchmark, namely MM-Bench. The \textcolor{gray}{dashed gray} line indicates the performance of the-state-of-the-art LLaVA 1.5 model.}
   \label{fig:num_llava_vs_mmbench}
\end{figure}

\subsection{Why VLMs Trained on \dataname{} are Better than State-of-the-Art VLMs?}
\label{appx:why_vision_flan_better}
In this section, we perform two case studies to explain why models trained on \dataname{} can perform better compared to state-of-the-art VLMs.

\subsubsection{Case Study on OCR}
\begin{figure}[h!]
  \centering
   \includegraphics[width=0.9\linewidth]{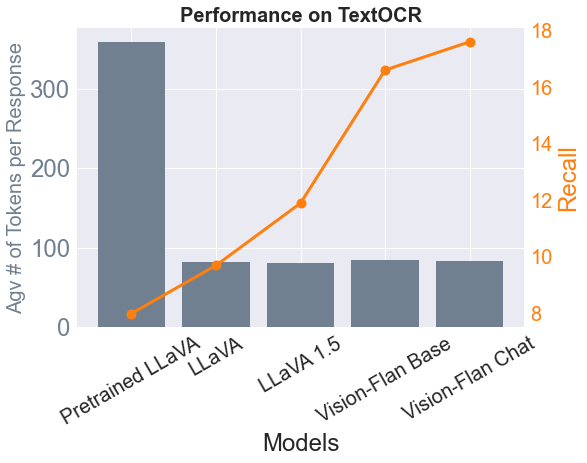}

   \caption{Performance of various VLMs on TextOCR. The \textcolor{gray}{gray bars} shows the averaged number of tokens per prediction and the \textcolor{orange}{orange line} show the recall of predictions.}
   \label{fig:textocr}
\end{figure}
\begin{figure}[ht!]
  \centering
   \includegraphics[width=\linewidth]{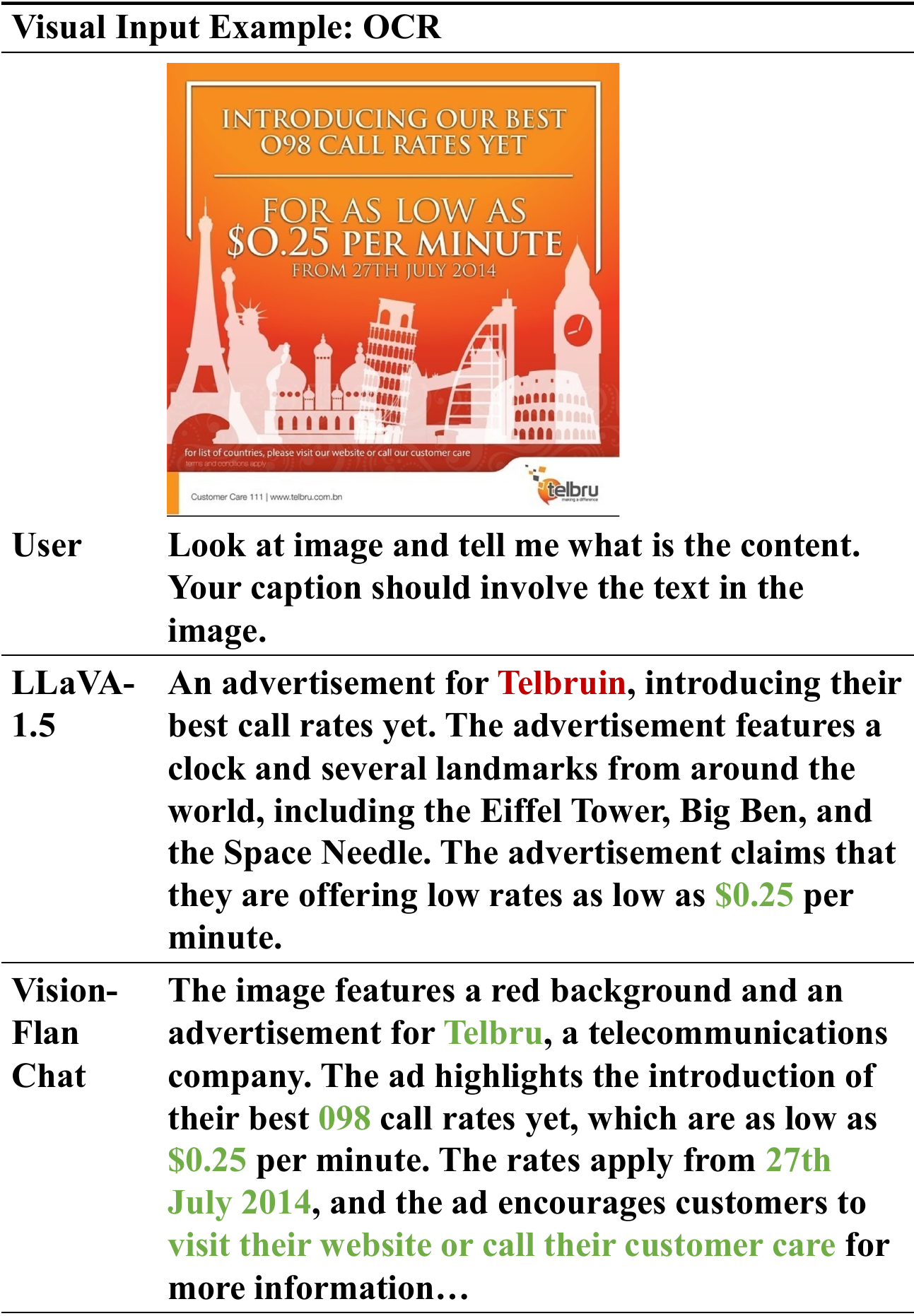}

   \caption{An example from TextCap to show that Vision-Flan allows VLMs to better recognize text.}
   \label{fig:ocr_example}
\end{figure}

When we manually check the predictions of \chatmodelname{} and compare them to other VLMs, the first trend we observe is that \chatmodelname{} can better perform OCR as shown in Figure~\ref{fig:ocr_example}. To quantify this observation, we evaluate LLaVA, LLaVA 1.5 and our models on the challenging TextOCR dataset~\cite{textocr}. 
We ask the VLMs to predict all the text on each image and check the overlap between the target list of text pieces and the predictions. As shown in Figure~\ref{fig:textocr}, the recall of \modelname{} and \chatmodelname{} is much higher compared to LLaVA 1.5 while the averaged numbers of predicted tokens per response are similar.

\subsubsection{Case Study on Entity Recognition}
\begin{figure}[ht!]
  \centering
   \includegraphics[width=\linewidth]{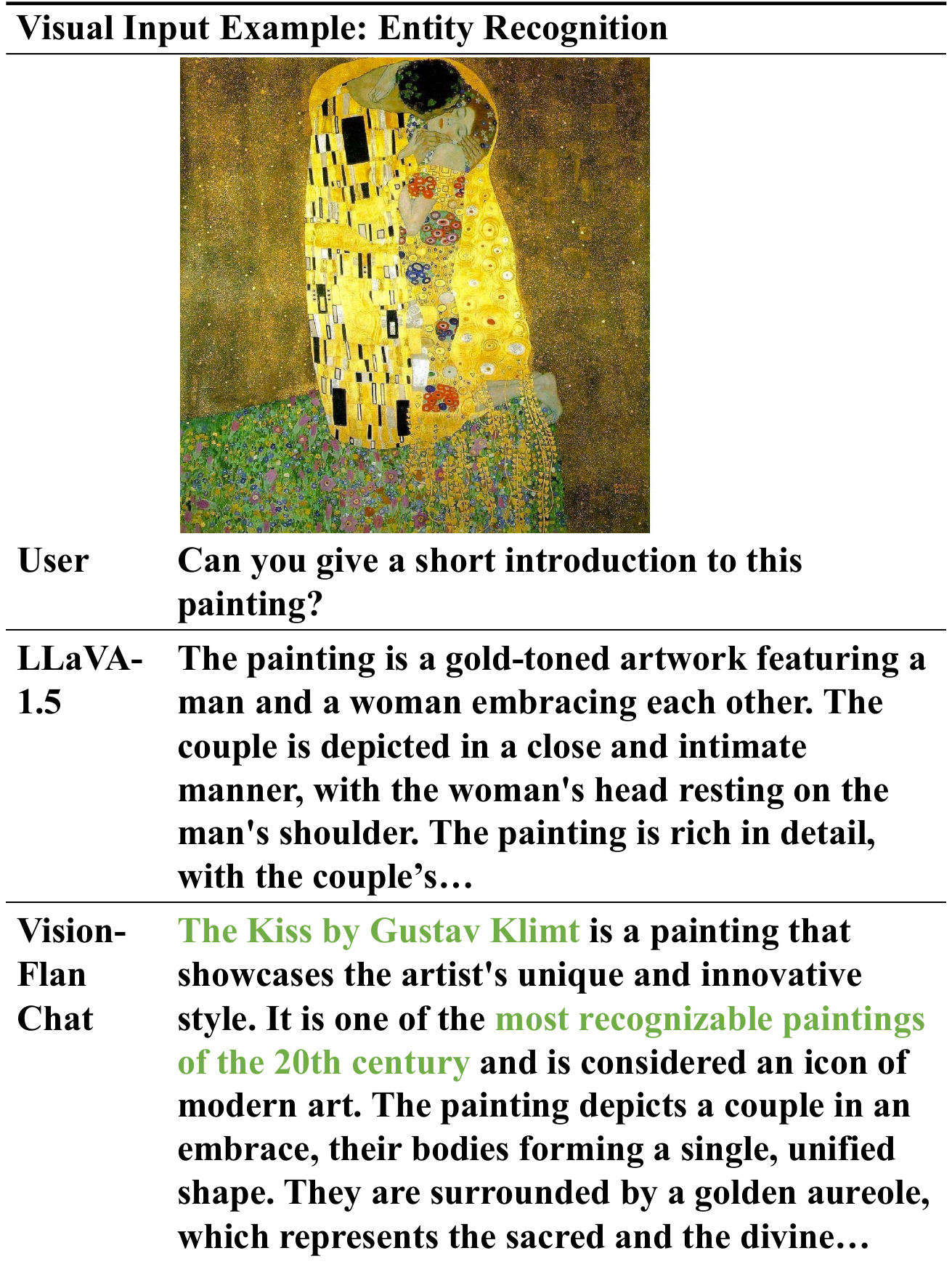}
   \caption{An example from MM-Vet to show that Vision-Flan allows VLMs to better recognize entities.}
   \label{fig:entity_qualitative}
\end{figure}

We also spot that models trained on \dataname{} can better identify entities in an image while LLaVA 1.5 simply captions the appearance of entities in an image. A qualitative example is shown in Figure \ref{fig:entity_qualitative}.

\begin{figure}[ht!]
  \centering
   \includegraphics[width=0.9\linewidth]{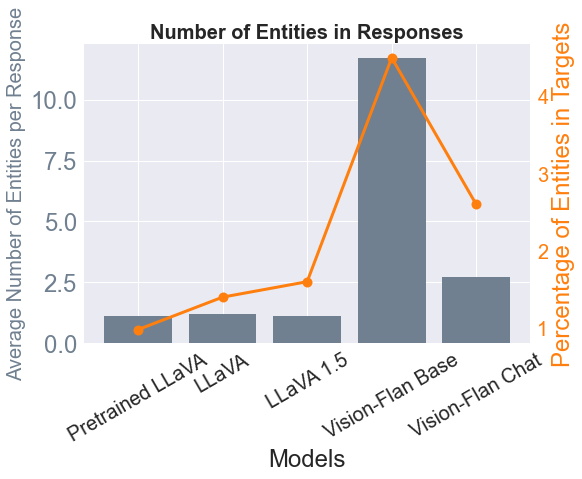}

   \caption{Performance of various VLMs on Entity Recognition. The \textcolor{gray}{gray bars} show the average number of entities per response and the \textcolor{orange}{orange line} shows the percentage of entities in the target response that appears in the prediction.}
   \label{fig:entity_wit}
\end{figure}

To compute quantitative results, we randomly sample 1,000 images with their captions from the WIT dataset~\cite{10.1145/3404835.3463257}, in which the images are from Wikipedia pages and the captions commonly contain entities appearing in the images. We prompt VLMs to introduce the entities in the image with some background knowledge. We leverage the EntityRecognizer from spaCy~\footnote{\url{https://spacy.io/api/entityrecognizer}} to recognize the entities in both predictions and ground truth captions and compute the percentage of target entities appearing in the predictions. As shown in Figure~\ref{fig:entity_wit}, it is clear that \modelname{} and \chatmodelname{} predict more entities in their responses (\textcolor{gray}{gray bars}) and have higher coverage of entities (\textcolor{orange}{orange line}) compared to LLaVA 1.5.

\section{Additional Analysis}

\subsection{The Bridging Module Can Be Shared Across LLMs with Same Architecture}

\begin{table*}[h!]
  \centering
  \resizebox{0.8\textwidth}{!}{%
  \begin{tabular}{l c c c c c c c c c}
  \toprule
   \textbf{Model}  & MM-Bench & MME & LLaVA-Bench & Pope\\
  \midrule
  \rowcolor{Gray}\multicolumn{1}{l}{\bf{Pretrained \llavamodel{}}} & 45.0 & 936.3  & 32.4 & 51.9\\
  \rowcolor{columbiablue}\multicolumn{1}{l}{+ LLaMA 2 Chat }& 45.3 \textcolor{forestgreen}{(100.6)}&557.0 \textcolor{Red}{(59.5)} & 59.2 \textcolor{forestgreen}{(182.7)} &66.9 \textcolor{forestgreen}{(128.9)}\\
  \rowcolor{Gray}\multicolumn{1}{l}{\bf{\modelname{} w/ frozen LLM}} & 52.4 & 1107.3 & 41.6 & 83.3  \\
  \rowcolor{columbiablue}\multicolumn{1}{l}{+ LLaMA 2 Chat} & 46.6 \textcolor{forestgreen}{(88.9)} & 1095.8 \textcolor{forestgreen}{(99.0)} & 56.4 \textcolor{forestgreen}{(135.6)} & 80.9 \textcolor{forestgreen}{(97.1)}\\
  \rowcolor{Gray}\multicolumn{1}{l}{\bf{\modelname{}}} & 69.8 & 1537.8 & 38.5 & 85.9 \\
  \rowcolor{columbiablue}\multicolumn{1}{l}{+ LLaMA 2 Chat}& 47.2 \textcolor{Red}{(67.6)} & 852.6 \textcolor{Red}{(55.4)} & 69.9 \textcolor{forestgreen}{(181.6)} &66.1 \textcolor{Red}{(76.9)}\\
  \rowcolor{Gray}\multicolumn{1}{l}{\bf{\chatmodelname{}}} & 67.6 & 1490.6 &78.3 & 86.1\\
  \rowcolor{columbiablue}\multicolumn{1}{l}{+ LLaMA 2 Chat}&47.0 \textcolor{Red}{(69.5)}& 869.6 \textcolor{Red}{(59.3)} &74.6 \textcolor{forestgreen}{(95.3)} & 65.8 \textcolor{Red}{(76.4)}\\
    
  \bottomrule
  \end{tabular}}
  \caption{Results of replacing Vicuna 1.5 with LLaMA 2 Chat in four VLMs. The \textcolor{gray}{gray} rows denote the performance of original models and \textcolor{blue}{blue} rows denote the performance of the VLMs after replacing the LLMs. The number in each bracket denotes the percentage of VLMs' performance after integration of LLaMA 2 Chat, compared to their original performance.}
  \label{table:share_mlp_results}
  \end{table*}
Recent studies~\cite{jain2023analyzeAlignment} in aligning and finetuning LLMs suggest that alignment happens very localized with pruning of a few weights or neurons to alter the style and format of outputs from LLMs, and does not substantially change the parameter space of LLMs. Following this finding, we hypothesize that \textit{the MLP layers that map visual features into LLMs' embedding space can be shared across LLMs with identical architecture but are tuned on different text alignment datasets}. As shown in Table~\ref{table:share_mlp_results}, we take four different models including \modelname{} w/ frozen LLM which is finetuned on \dataname{} but with LLMs kept frozen as a case study, and directly replace their LLMs (Vicuna v1.5) with off-the-shelf LLaMA 2 Chat model.
During inference, we use the official prompting template of LLaMA 2 chat instead of Vicuna v1.5. The results demonstrate that MLPs can be shared between LLMs with the same architecture but trained on different alignment datasets. An interesting observation is that there is a significant performance boost on LLaVA-Bench after we swap in LLaMA 2 Chat. If we finetune both the MLPs and the LLMs in \modelname{} and \chatmodelname{}, we observe a remarkable performance drop when we swap in LLaMA 2 chat. This is understandable because the LLaMA 2 chat can not effectively interpret visual features compared to the visual-instruction-tuned Vicuna v1.5.

\subsection{Discrepancy Between Evaluation Benchmarks}
In Table \ref{table:main_results} and \ref{table:share_mlp_results}, we identify large performance discrepancy between multiple-choice benchmarks (e.g., MME and MM-Bench) and LLaVA-Bench on several models. 
Specifically, in Table \ref{table:main_results}, LLaVA achieves a score of 70.8 on LLaVA-Bench, comparable to the performance level of LLaVA 1.5. In contrast, LLaVA's performance on MME and MM-Bench is markedly lower, with scores of 1151.6 and 38.7, respectively, compared to LLaVA 1.5, which scores 1531.3 and 66.7. Furthermore, this trend is also evident in Table \ref{table:share_mlp_results}. Upon substituting the LLMs in \modelname{} and \chatmodelname{} with off-the-shelf LLaMA 2 Chat, both models exhibit a notable decline in performance on MME and MM-Bench, while maintaining comparable performance on LLaVA-Bench. Our hypothesis posits that LLaVA-Bench does not require LLM's strong understanding of the visual features, but rather relies on the language-prior of LLMs~\cite{lin2023languagePriorVLM}. Furthermore, the data synthesized by GPT-4 facilitates the model's ability to generate long-form responses, aligning with the preferences of the evaluation metric, namely, GPT-4 itself.

\section{Additional Related Work}
\label{appx:additional_related_work}
\paragraph{Vision-Language Models.} 
Previous works~\cite{li2019visualbert,chen2020uniter,tan2019lxmert,su2019vl, wang2022image}  mainly pretrain vision-language models (VLMs) from scratch with a unified masked-language modeling (MLM) objective~\cite{devlin2018bert}, which can impose significant training cost and inferior performance.
Recently, a line of works proposes to build VLMs from the off-the-shelf visual encoders and LLMs by introducing a small amount of bridging parameters that maps visual features into the embedding space of LLMs. Flamingo~\cite{alayrac2022flamingo} presents a VLM that is capable of processing interleaved image-text inputs and generating responses based on the visual content. It proposes Perceiver Resampler as the bridging module to connect the frozen LLM and visual encoder.
OFA~\cite{wang2022ofa} proposes a sequence-to-sequence learning framework that maps images to discrete visual tokens and feeds the discrete visual tokens into LLMs. 
BLIP-2~\cite{li2023blip} introduces Q-Former to bridge pre-trained and frozen vision and language models, based on which, MiniGPT-4~\cite{zhu2023minigpt} further adds a linear projector to bridge the gap between the visual encoder and language model encoder.
LLaVA~\cite{liu2023llava} introduces a projector to fuse visual information into a large language model and unfreezes language model during visual instruction tuning. 

\clearpage
\section{Datasets Used in \dataname{}}
\label{appx:dataset_used_in_vision_flan}
\begin{table*}[ht]
\begin{tabular}{|p{75mm}|p{75mm}|}
 \hline
 Dataset \& Reference & Tasks\\ \hline\hline

CINIC-10 \cite{darlow2018cinic} & 1. animal recognition in low resolution image \newline 2. shipping method recognition in low resolution image \newline 3. transportation option recognition in low resolution image \newline 4. animal presence classification in low resolution image \newline 5. object shipping object presence in low resolution image\\ \hline
MSCOCO \cite{lin2014microsoft} & 1. multiple choice VQA \newline
2. short image captioning \newline 
3. appliance recognition \newline
4. furniture recognition \newline
5. kitchen object recognition \newline
6. vehicle recognition \newline
7. animal recognition  \newline
8. sports object recognition \newline
9. image text matching \newline
10. image text selection
\\ \hline
FairFace \cite{karkkainenfairface} & 1. human age classification \newline
2. human gender classification \newline 
3. human race classification  \\ \hline
IconQA \cite{lu2021iconqa} & 1. abstract diagram understanding \newline
2. fill in blank in abstract diagram understanding 
\\ \hline
ImageNet-A \cite{hendrycks2021nae} & 1. object recognition of natural adversarial examples \\ \hline
ImageNet-C \cite{hendrycks2019benchmarking} & 1. blur type classification \newline
2. coarse-grained image corruption classification \newline 
3. weather type classification \newline
4. fine-grained image corruption classification \\ \hline
InfographicVQA \cite{mathew2022infographicvqa} & 1. VQA \newline
2. document level VQA \\ \hline
SemArt \cite{garcia2018read} & 1. painting time frame recognition \newline
2. painting type recognition \newline 
3. painting school recognition \newline 
4. painting technique recognition \newline 
5. detailed image description \\ \hline
Set5
 \cite{bevilacqua2012low} & 1. object recognition in low resolution image \\ \hline
TextCaps \cite{sidorov2019textcaps} & 1. image captioning with reading comprehension \\ \hline
VisDial \cite{das2017visual} & 1. visual dialogue with short context \newline 
2. visual dialogue with medium context \newline 
3. visual dialogue with long context \newline
4. visual dialogue with very long context 
\\ \hline
STL-10 \cite{DBLP:journals/jmlr/CoatesNL11} & 1. object recognition \\ \hline
Places365 \cite{zhou2017places} & 1. scene classification  \\ \hline
Office-31 \cite{DBLP:conf/eccv/SaenkoKFD10} & 1. image domain and office object classification 
\newline
2. office object recognition
\\ \hline
\end{tabular}
\end{table*}

\begin{table*}[ht]
\begin{tabular}{|p{75mm}|p{75mm}|}
        \hline
        Dataset \& Reference  & Tasks\\ \hline\hline
LSUN \cite{yu15lsun} & 1. scene classification   \\ \hline
FGVC-Aircraft \cite{maji13fine-grained} & 1. aircraft family classification \newline
2. aircraft manufacturer classification \newline
3. aircraft variant classification \newline
4. aircraft model classification \\ \hline
DeepFashion \cite{liu2016deepfashion} & 1. cloth texture classification \\ \hline
CUB-200-2011 \cite{WahCUB_200_2011} & 1. bird species recognition  \\ \hline
CLEVR \cite{DBLP:conf/cvpr/JohnsonHMFZG17} & 1. VQA in 3D rendered images \newline
2. question answer matching \newline
3. visual dialogue in 3D rendered images \newline
4. VQA in 3D rendered images with multiple questions
\\ \hline
CLEVR-CoGenT~\cite{DBLP:conf/cvpr/JohnsonHMFZG17}& 1. VQA in 3D rendered images \newline
2. question answer matching \newline
3. VQA in 3D rendered images with multiple questions
\\ \hline
A-OKVQA \cite{AOKVQA} & 1. rationales generation  
\newline
2. answer rationale generation \newline
3. outside knowledge VQA
\\ \hline
AI2D \cite{kembhavi2016diagram} & 1. diagram VQA  \\ \hline
AID \cite{xia2017aid} & 1. aerial scene classification \\ \hline
Caltech-256 \cite{griffin2007caltech} & 1. object recognition \\ \hline
CoVA \cite{kumar-etal-2022-cova} & 1. webpage recognition \\ \hline
DeepWeeds \cite{DeepWeeds2019} & 1. weed species recognition \\ \hline
ExDark \cite{Exdark} & 1. object recognition in low light environments \\ \hline
FFHQ-Text \cite{zhou2021generative} & 1. facial attribute textual descriptions generation \\ \hline
FlickrLogos-27 \cite{kalantidis2011scalable} & 1. logo recognition \\ \hline
FoodLogoDet-1500~\cite{hou2021foodlogodet}  & 1. food logo recognition
 \\ \hline
ImageNet-R \cite{hendrycks2021many} & 1. object recognition in diverse image domain \newline
2. image style classification\\ \hline
ImageNet-Sketch \cite{wang2019learning} & 1. object recognition in sketch \\ \hline
JHU-CROWD++ \cite{sindagi2019pushing} & 1. scene classification \\ \hline
MNIST-M \cite{ganin2016domain} & 1. number recognition \\ \hline
MVTecAD \cite{bergmann2019mvtec} & 1. object anomaly detection \newline
2. industrial item recognition
\\ \hline

\end{tabular}
\end{table*}

\begin{table*}
    \begin{tabular}{|p{75mm}|p{75mm}|}
        \hline
        Dataset \& Reference  & Tasks\\ \hline\hline

NABirds \cite{van2015building} & 1. bird species recognition in north America \newline
2. bird body parts detection \\ \hline
Road-Anomaly \cite{lis2019detecting} & 1. road anomaly detection \\ \hline
SCUT-CTW1500 \cite{yuliang2017detecting} & 1. curve text detection in the wild \\ \hline
Total-Text \cite{CK2019} & 1. scene text detection and recognition \\ \hline
VisDA-2017 \cite{peng2017visda} & 1. object recognition in 3D rendered image \newline
2. multiple choice object recognition in 3D rendered image\\ \hline
Yoga-82 \cite{verma2020yoga} & 1. yoga pose recognition \\ \hline
Caltech101~\cite{FeiFei2004LearningGV} & 1. object recognition \newline 
2. living organism classification \\ \hline
Cars~\cite{krause20133d} & 1. car brand maker and year classification \newline
2. car brand classification \\ \hline
Core50 \cite{lomonaco2017core50} & 1. object recognition \\ \hline
NUS-WIDE \cite{nus-wide-civr09} & 1. animal presence classification \\ \hline
ObjectNet \cite{NIPS2019_9142} & 1. object recognition \\ \hline
Places205 \cite{zhou2014learning} & 1. indoor outdoor classification \\ \hline
300w  \cite{sagonas2016300} & 1. indoor outdoor classification \\ \hline
Yahoo \cite{farhadi2009describing} & 1. object recognition \\ \hline
LFW \cite{LFWTech} & 1. celebrity recognition \\ \hline
model-vs-human \cite{geirhos2018} & 1. image-style classification \\ \hline
Office-Home \cite{venkateswara2017deep} & 1. object recognition \\ \hline
Winoground \cite{thrush_and_ross2022winoground} & 1. image caption matching \\ \hline
ConceptualCaptions \cite{sharma2018conceptual} & 1. conceptual image captioning  \\ \hline
KVQA+image question answer \cite{shahMYP19} & 1. knowledge-aware VQA\newline 2. visual entity recognition \\ \hline
MemeCap \cite{hwang2023memecap} & 1. meme understanding \\ \hline
PlotQA \cite{Methani_2020_WACV} & 1. VQA over scientific plots \\ \hline
SentiCap \cite{mathews2016senticap} & 1. image captioning conditioned on sentiment  \\ \hline
VQA-E \cite{li2018vqae} & 1. VQA \newline
2. short image captioning \\ \hline
VQG \cite{mostafazadeh2016generating} & 1. visual question generation\newline2. short image captioning \\ \hline

    \end{tabular}
\end{table*}

\begin{table*}
    \begin{tabular}{|p{75mm}|p{75mm}|}
        \hline
        Dataset \& Reference  & Tasks\\ \hline\hline

WIT \cite{10.1145/3404835.3463257} & 1. background knowledge extraction \\ \hline
WikiArt \cite{artgan2018} & 1. artist genre style recognition \\ \hline
VQA-RAD \cite{Lau_Gayen_Demner_Ben_Abacha_2019} & 1. VQA in radiology \\ \hline
VOC2007 \cite{pascal-voc-2007} & 1. multiple object recognition \\ \hline
VizWiz \cite{DBLP:conf/eccv/GurariZZB20} & 1. answering visual questions from blind people\newline 2. captioning image taken by blind people\newline 3. quality issue classification of image taken by blind people \\ \hline
ViQuAE \cite{lerner2022viquae} & 1. knowledge based VQA about entities \\ \hline
ST-VQA \cite{biten2019scene} & 1. scene text VQA \\ \hline
Stanford Dogs \cite{KhoslaYaoJayadevaprakashFeiFei_FGVC2011} & 1. dog species classification \\ \hline
Sketch \cite{eitz2012hdhso} & 1. living organism classification in sketch \newline
2. object recongnition in sketch
 \\ \hline
RAVEN \cite{zhang2019raven} & 1. relational and analogical visual reasoning \\ \hline
PICKAPIC \cite{Kirstain2023PickaPicAO} & 1. image prompt generation \\ \hline
PACS \cite{DBLP:conf/iccv/LiYSH17} & 1. object recognition in art painting\newline 
2. object recognition in cartoon\newline
3. object recognition in photograph\newline
4. dog image style classification\newline
5. elephant image style classification\newline
6. giraffe image style classification\newline
7. guitar image style classification\newline
8. horse image style classification\newline
9. house image style classification\newline
10. person image style classification\\ \hline
NOCAPS \cite{agrawal2019nocaps} & 1. multiple short captions generation \\ \hline
Localized Narratives \cite{PontTuset_eccv2020} & 1. COCO detailed image captioning\newline 2. flickr30k detailed image captioning\newline 3. open images detailed image captioning\newline 4. ade20k detailed image captioning \\ \hline
INATURALIST \cite{vanhorn2018inaturalist} & 1. class classification\newline
2. family classification
\newline
3. genus classification
\newline
4. Latin English name classification
\newline
5. order classification
\newline
6. phylum classification
\newline
7. supercategory classification\\ \hline
HICO \cite{chao:iccv2015} & 1. human activity detection \\ \hline
GEOMETRY3K \cite{lu2021inter} & 1. geometry question answering \\ \hline
FUNSD \cite{jaume2019} & 1. text detection in noisy scanned documents \\ \hline
FLICKR30K \cite{flickrentitiesijcv} & 1. multiple captions generation\\ \hline
DVQA \cite{kafle2018dvqa} & 1. chart question answering \\ \hline
DTD \cite{cimpoi14describing} & 1. coarse grained texture classification
\newline
2. multiple texture detection\\ \hline

    \end{tabular}
\end{table*}

\begin{table*}
    \begin{tabular}{|p{75mm}|p{75mm}|}
        \hline
        Dataset \& Reference  & Tasks \\ \hline\hline
DOMAIN NET \cite{peng2019moment} & 1. object recognition in clip art
\newline
2. object recognition in infograph
\newline
3. object recognition in painting
\newline
4. object recognition in quickdraw
\newline
5. object recognition in real image
\newline
6. image style classification\\ \hline

DOCVQA \cite{DBLP:journals/corr/abs-2007-00398} & 1. document level VQA \\ \hline
DAQUAR \cite{malinowski2014nips} & 1. VQA \\ \hline
CONCADIA \cite{kreiss2022concadia} & 1. caption with background knowledge 
\newline
2. short image captioning \\ \hline

Visual7W \cite{zhu2016visual7w} & 1. VQA object attribute
 \\ \hline

VQAv2 \cite{goyal2017making} & 1. general VQA
\newline
2. question image matching
 \\ \hline

Visual Genome\cite{krishna2017visual} & 1. spatial relationship question answering 
 \\ \hline

OK-VQA\cite{marino2019ok} & 1. outside knowledge VQA
 \\ \hline

ScienceQA~\cite{lu2022learn}& 1. VQA
\newline
2. explanation generation 
 \\ \hline 

OCR-VQA~\cite{mishraICDAR19}& 1. VQA by reading text in image
 \\ \hline 

wikiHow-image~\cite{yang2021visual}& 1. next step generation
\newline
2. image text step ordering
\newline
3. immediate next step selection
\newline
4. text image step ordering
 \\ \hline 

SciCap~\cite{hsu2021scicap}& 1. figure captioning 
 \\ \hline 

LAD~\cite{zhao2019large}& 1. detailed object description generation
 \\ \hline 
 
Dark Zurich~\cite{sakaridis2019guided}& 1. time of the day classification
 \\ \hline 

RAF-DB~\cite{li2019reliable}& 1. human emotion detection
 \\ \hline 

GQA~\cite{hudson2019gqa}& 1. spatial relationship question answering
 \\ \hline 

VQA~\cite{antol2015vqa}& 1. color 
\newline
2. activity recognition
\newline
3. counting
\newline
4. object presence
\newline
5. object recognition
\newline
6. positional reasoning
\newline
7. scene recognition
\newline
8. sentiment understanding
\newline
9. sport recognition
\newline
10. utility affordance
 \\ \hline 
Multimodal Factual Checking~\cite{DBLP:conf/sigir/YaoS0CH23}& 1. multimodal factual checking
\\ \hline 


    \end{tabular}
\end{table*}
\clearpage
\section{Task Categories in \dataname{}}
\label{appx:task_group_in_vision_flan}
\begin{table*}[ht]
\begin{tabular}{|p{75mm}|p{75mm}|}
 \hline
 Category & Tasks\\ \hline\hline

Perception & 1. CLEVR-CoGenT VQA  in 3D rendered images \newline
2. CLEVR-CoGenT question answer matching \newline
3. CLEVR-CoGenT VQA in 3D rendered images with multiple
questions \newline
4. CLEVR VQA in 3D rendered images with multiple
questions \newline
5. GQA spatial relationship question answering \newline
6. VQA color 
\newline
7. VQA activity recognition
\newline
8. VQA counting
\newline
9. VQA object presence
\newline
10. VQA object recognition
\newline
11. VQA positional reasoning
\newline
12. VQA scene recognition
\newline
13. VQA sentiment understanding
\newline
14. VQA sport recognition
\newline
15. VQA utility affordance \newline
16. VQA-E VQA \newline
17. VQAv2 general VQA \newline
18. Visual Genome spatial relationship question answering \newline
19. CLEVR question answer matching \newline
20. VizWiz answering visual questions from blind people \newline
21. DAQUAR VQA \newline
22. MSCOCO multiple choice VQA \newline
23. Visual7W VQA object attribute \newline
24. CLEVR VQA in 3D rendered images 
\\ \hline
Outside Knowledge & 1. KVQA knowledge aware VQA \newline
2. VIQUAE knowledge based VQA about entities \newline
3. VQARAD VQA in radiology \newline
4. OK-VQA outside knowledge VQA \newline
5. A-OKVQA outside knowledge VQA 
\\ \hline
Reasoning & 1. GEOMETRY3K geometry question answering \newline
2. IconQA abstract diagram understanding \newline
3. IconQA fill in blank in abstract diagram understanding \newline
4. InfographicVQA VQA \newline
5. InfographicVQA document level VQA
\newline
6. ScienceQA VQA \newline
7. AI2D diagram VQA 
\\ \hline
OCR & 1. DOCVQA document level VQA\newline
2. DVQA chart question answering \newline
3. PlotQA VQA over scientific plots \newline
4. OCR-VQA VQA by reading text in image \newline
5. ST-VQA scene text VQA
\\ \hline

\end{tabular}
\end{table*}

\begin{table*}[ht]
\begin{tabular}{|p{75mm}|p{75mm}|}
 \hline
 Category & Tasks\\ \hline\hline

Document-Level OCR
 & 1. FUNSD text detection in noisy scanned documents \newline
 2. SCUT-CTW1500 curve text detection in the wild \newline
 3. Total-Text scene text detection and recognition
\\ \hline
Phrase-Level OCR & 1. CoVA webpage recognition \newline
2. FlickrLogos-27 logo recognition \newline
3. FoodLogoDet-1500 food logo recognition 
\\ \hline
Knowledge Extraction & 1. CONCADIA caption with background knowledge \newline
2. KVQA visual entity recognition \newline
3. WIT background knowledge extraction
\\ \hline
Semantic Art Understanding & 1. Semart painting time frame recognition \newline
2. Semart painting type recognition \newline
3. Semart painting school recognition \newline
4. Semart painting technique recognition \newline
5. Semart detailed image description \newline
6. WikiArt artist genre style recognition
\\ \hline
Visual Dialogue & 1. CLEVR visual dialogue in 3D rendered images \newline
2. Visdial visual dialogue with short context \newline
3. Visdial visual dialogue with medium context \newline
4. Visdial visual dialogue with long context  \newline
5. Visdial visual dialogue with very long context  
\\ \hline
Rational and Script Generation & 1. ScienceQA explanation generation\newline
2. A-OKVQA rationales generation\newline
3. A-OKVQA answer rationale generation \newline
4. MemeCap meme understanding \newline
5. wikiHow-image  next step generation \newline
6. VQG visual question generation
\\ \hline
Coarse-grained Captioning & 1. ConceptualCaptions conceptual image captioning\newline
2. FLICKR30K multiple captions generation\newline
3. NOCAPS multiple short captions generation \newline
4. PICKAPIC image prompt generation \newline
5. VizWiz captioning image taken by blind people \newline
6. VQA-E short image captioning \newline
7. 
VQG short image captioning \newline
8. MSCOCO short image captioning \newline
9. CONCADIA short image captioning
\\ \hline

\end{tabular}
\end{table*}

\begin{table*}[ht]
\begin{tabular}{|p{75mm}|p{75mm}|}
 \hline
 Category & Tasks\\ \hline\hline
 Fine-grained Captioning & 1. LAD detailed object description generation \newline
2. FFHQ-Text facial attribute textual descriptions generation \newline
3. Localized Narratives COCO detailed image captioning \newline
4. Localized Narratives flickr30k detailed image captioning \newline
5. Localized Narratives open images detailed image captioning \newline
6. Localized Narratives ade20k detailed image captioning \newline
7. SciCap figure captioning  \newline
8. SentiCap image captioning conditioned on sentiment \newline
9. TextCaps image captioning with reading comprehension
\\ \hline
Scene Classification & 1. 300w indoor outdoor classification \newline
2. AID aerial scene classification \newline
3. Dark-Zurich time of the day classification \newline
4. JHU-CROWD scene classification \newline
5. LSUN scene classification \newline
6. Places205 indoor outdoor classification \newline
7. places365 scene classification \\ \hline
Animal Classification & 1. CUB-200-2011 bird species recognition\newline
2. DeepWeeds weed species recognition \newline
3. INATURALIST class classification\newline
4. INATURALIST family classification \newline
5. INATURALIST genus classification \newline
6. INATURALIST Latin English name classification
\newline
7. INATURALIST order classification
\newline
8. INATURALIST phylum classification
\newline
9. INATURALIST supercategory classification \newline
10. NABirds bird species recognition in north America \newline
11. NUS-WIDE animal presence classification \newline
12. STANFORD DOGS dog species classification \newline
13. NABirds bird body parts detection
\\ \hline

\end{tabular}
\end{table*}

\begin{table*}[ht]
\begin{tabular}{|p{75mm}|p{75mm}|}
 \hline
 Category & Tasks\\ \hline\hline
Vehicle Classification & 1. Cars car brand maker and year classification\newline
2. Cars car brand classification\newline
3. FGVC-Aircraft aircraft family classification\newline
4. FGVC-Aircraft aircraft manufacturer classification\newline
5. FGVC-Aircraft aircraft variant classification \newline
6. FGVC-Aircraft aircraft model classification
\\  \hline
Human Activity & 1. HICO human activity detection\newline
2. RAF-DB human emotion detection \newline
3. Yoga-82 yoga pose recognition
\\ \hline
Facial Recognition & 1. LFW celebrity recognition \newline
2. Fairface human age classification \newline
3. Fairface human gender classification \newline
4. Fairface human race classification
\\ \hline
Anomaly Detection & 1. Road-Anomaly road anomaly detection \newline
2. MVTecAD object anomaly detection
\\ \hline
General Object & 1. Caltech-256 object recognition\newline
2. Caltech101 object recognition \newline
3. Caltech101 living organism classification \newline
4. Core50 object recognition \newline
5. ImageNet-A object recognition of natural adversarial examples \newline
6. MNIST-M number recognition \newline
7. MVTecAD industrial item recognition \newline
8. ObjectNet object recognition \newline
9. Office-Home object recognition \newline
10. Office-31 image domain and office object classification \newline
11. Office-31 office object recognition \newline
12. STL-10 object recognition \newline
13. Set5 object recognition in low resolution image \newline
14. VOC2007 multiple object recognition \newline
15. MSCOCO appliance recognition \newline
16. MSCOCO furniture recognition \newline
17. MSCOCO kitchen object recognition \newline
18. MSCOCO vehicle recognition \newline
19. MSCOCO animal recognition \newline
20. MSCOCO sports object recognition \newline
21. Yahoo object recognition
\\ \hline

 \end{tabular}
\end{table*}

\begin{table*}[ht]
\begin{tabular}{|p{75mm}|p{75mm}|}
 \hline
 Category & Tasks\\ \hline\hline
Complex Reasoning & 1. RAVEN relational and analogical visual reasoning \newline
2. Multimodal Factual Checking multimodal factual checking \newline
3. wikiHow-image image text step ordering \newline
4. wikiHow-image immediate next step selection \newline
5. wikiHow-image text image step ordering 
\\ \hline
Image Text Matching & 1. MSCOCO image text matching \newline
2. Winoground image caption matching \newline
3. MSCOCO image text selection \newline
4. MSCOCO question image matching
\\ \hline
General Object Classification in Special Image Domain & 1. DOMAIN NET object recognition in clip art \newline
2. DOMAIN NET object recognition in infograph \newline
3. DOMAIN NET object recognition in painting \newline
4. DOMAIN NET object recognition in quickdraw \newline
5. DOMAIN NET object recognition in real image \newline
6. ExDark object recognition in low light environments \newline
7. ImageNet-R object recognition in diverse image domain  \newline
8. ImageNet-Sketch object recognition in sketch \newline
9. PACS object recognition in art painting \newline
10. PACS object recognition in cartoon \newline
11. PACS object recognition in photograph \newline
12. SKETCH living organism classification in sketch \newline
13. SKETCH object recognition in sketch \newline
14. Cinic-10 animal recognition in low resolution image \newline
15. Cinic-10 shipping method 
recognition in low resolution image \newline
16. Cinic-10 transportation option recognition in low resolution image \newline
17. Cinic-10 animal presence classification in low resolution image \newline
18. Cinic-10 object shipping object presence in low resolution image \newline
19. VisDA-2017 object recognition in 3D rendered image \newline
20. VisDA-2017 multiple choice object recognition in 3D rendered image
\\ \hline

  \end{tabular}
\end{table*}

\begin{table*}[ht]
\begin{tabular}{|p{75mm}|p{75mm}|}
 \hline
 Category & Tasks\\ \hline\hline
Image-Style Classification & 1. 
DOMAIN-NET image style classification \newline
2. ImageNet-R image style classification \newline
3. PACS dog image style classification \newline
4. PACS elephant image style classification \newline
5. PACS giraffe image style classification \newline
6. PACS guitar image style classification \newline
7. PACS horse image style classification \newline
8. PACS house image style classification \newline
9. PACS person image style classification \newline
10. Model-vs-human image style classification
\\ \hline

Image Quality Classification & 1. ImageNet-C blur type classification \newline
2. ImageNet-C coarse-grained image corruption classification \newline
3. ImageNet-C weather type classification\newline
4. ImageNet-C fine-grained image corruption classification\newline
5. VizWiz quality issue classification of image taken by blind people
\\ \hline
Texture Classification & 1. DTD coarse grained texture classification \newline
2. DTD multiple texture detection \newline
3. DeepFashion cloth texture classification
\\ \hline

   \end{tabular}
\end{table*}

\clearpage
\section{\dataname{} Tasks}
\label{appx:tasks_show}
\subsection{Generation Tasks}

\begin{figure*}[h!]
  \centering
   \includegraphics[width=\linewidth]{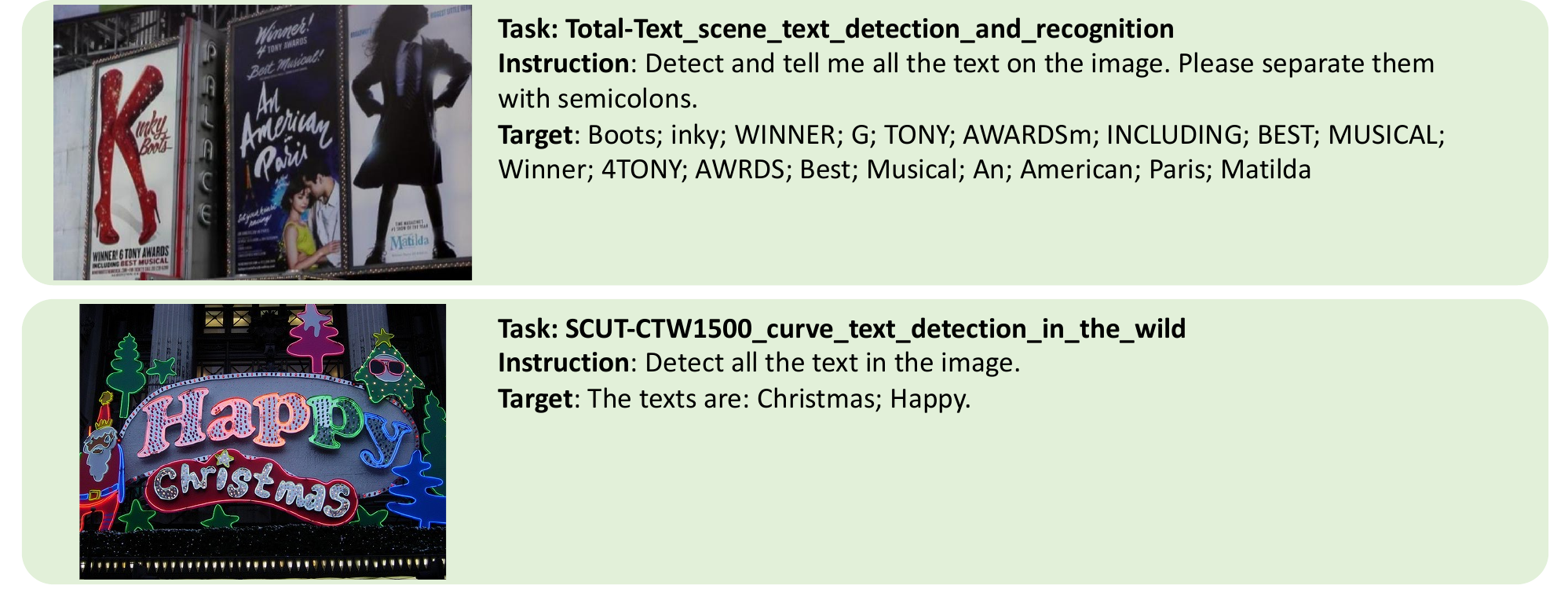}
   \caption{}
   \label{fig:}
\end{figure*}

\begin{figure*}[h!]
  \centering
   \includegraphics[width=\linewidth]{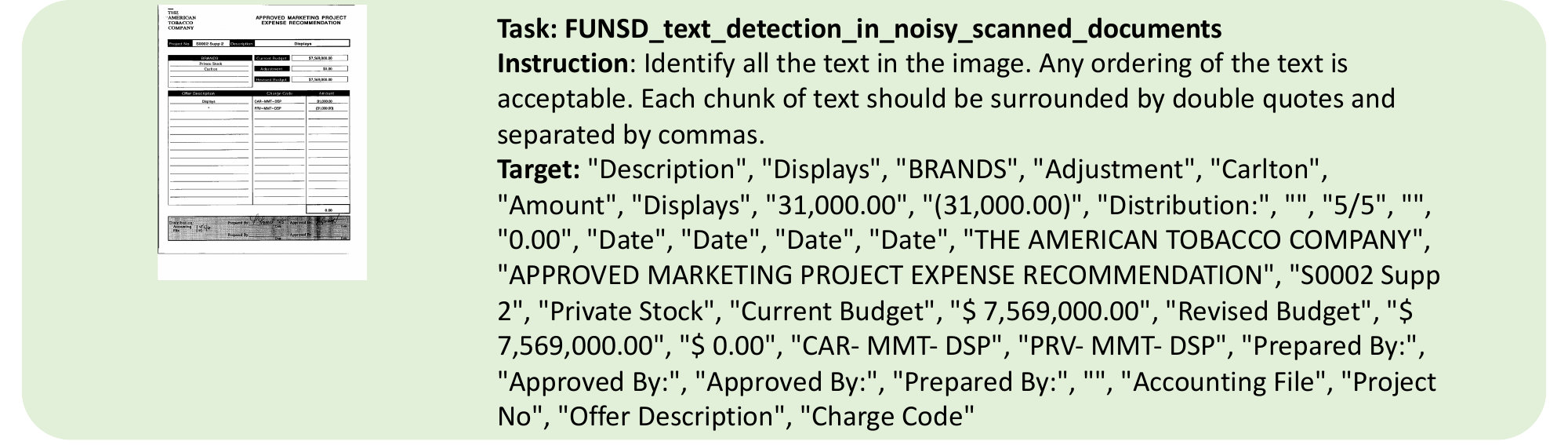}
   \caption{}
   \label{fig:}
\end{figure*}

\begin{figure*}[h!]
  \centering
   \includegraphics[width=\linewidth]{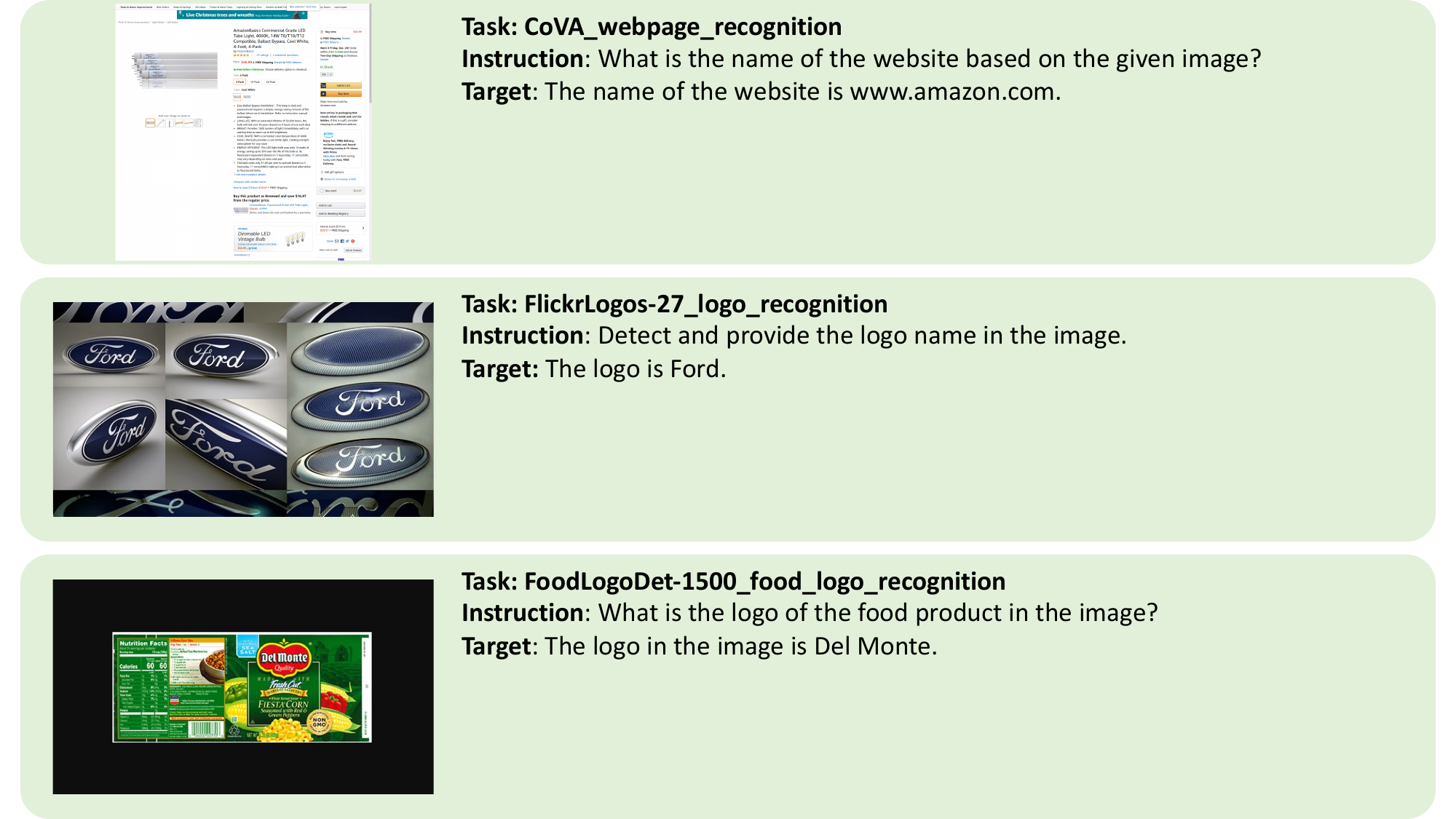}
   \caption{}
   \label{fig:}
\end{figure*}

\begin{figure*}[h!]
  \centering
   \includegraphics[width=\linewidth]{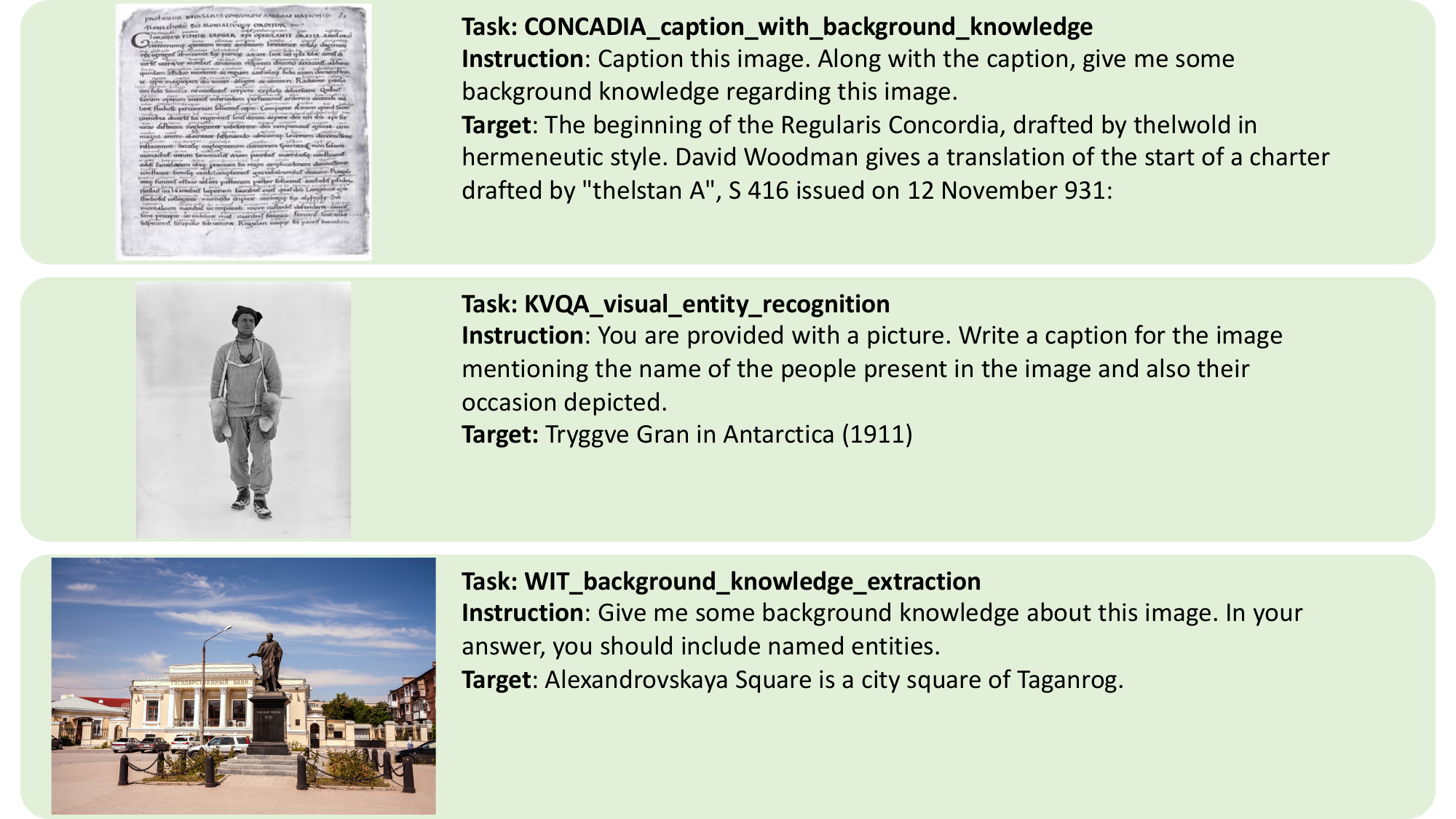}
   \caption{}
   \label{fig:}
\end{figure*}

\begin{figure*}[h!]
  \centering
   \includegraphics[width=\linewidth]{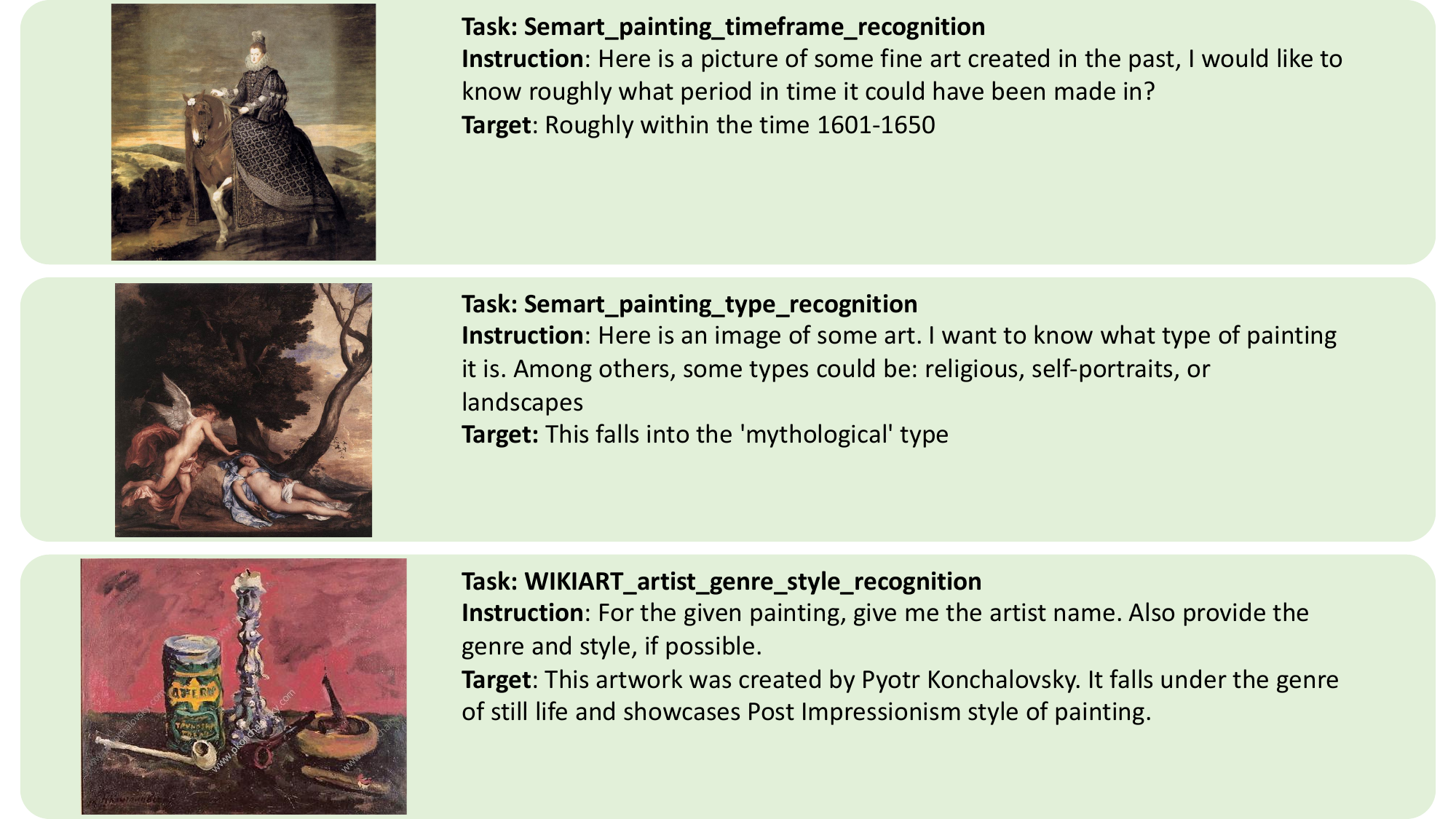}
   \caption{}
   \label{fig:}
\end{figure*}

\begin{figure*}[h!]
  \centering
   \includegraphics[width=\linewidth]{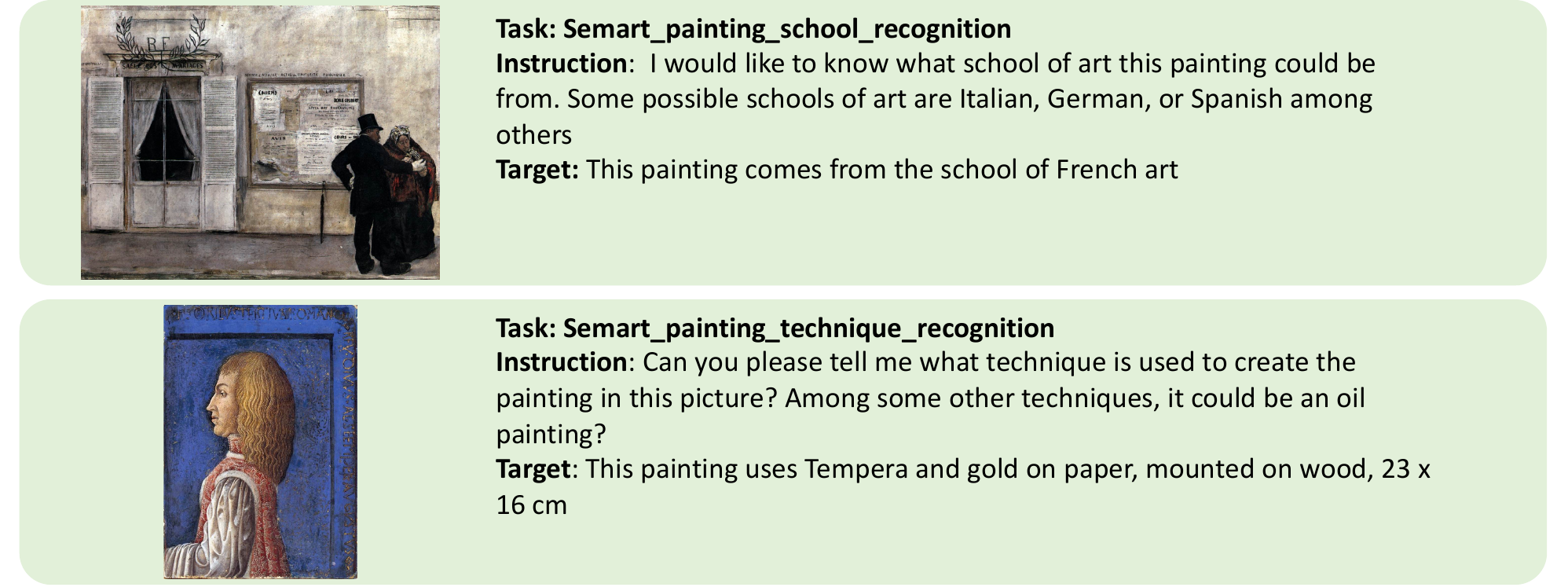}
   \caption{}
   \label{fig:}
\end{figure*}

\begin{figure*}[h!]
  \centering
   \includegraphics[width=\linewidth]{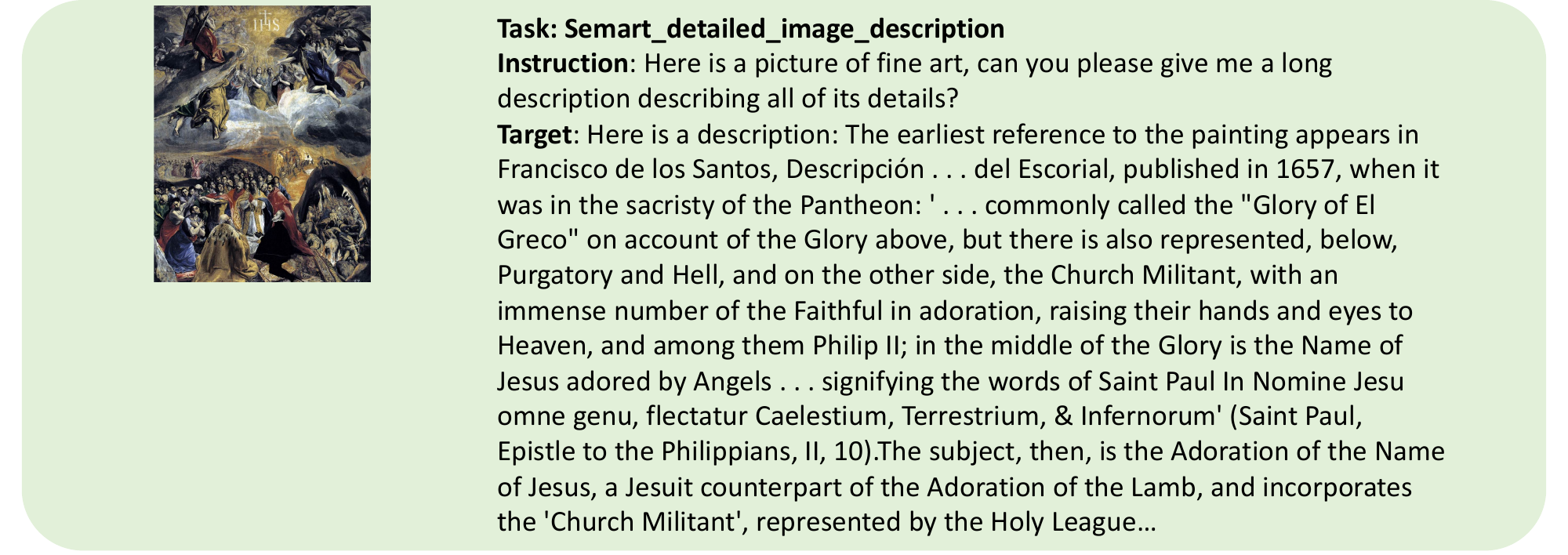}
   \caption{}
   \label{fig:}
\end{figure*}

\begin{figure*}[h!]
  \centering
   \includegraphics[width=\linewidth]{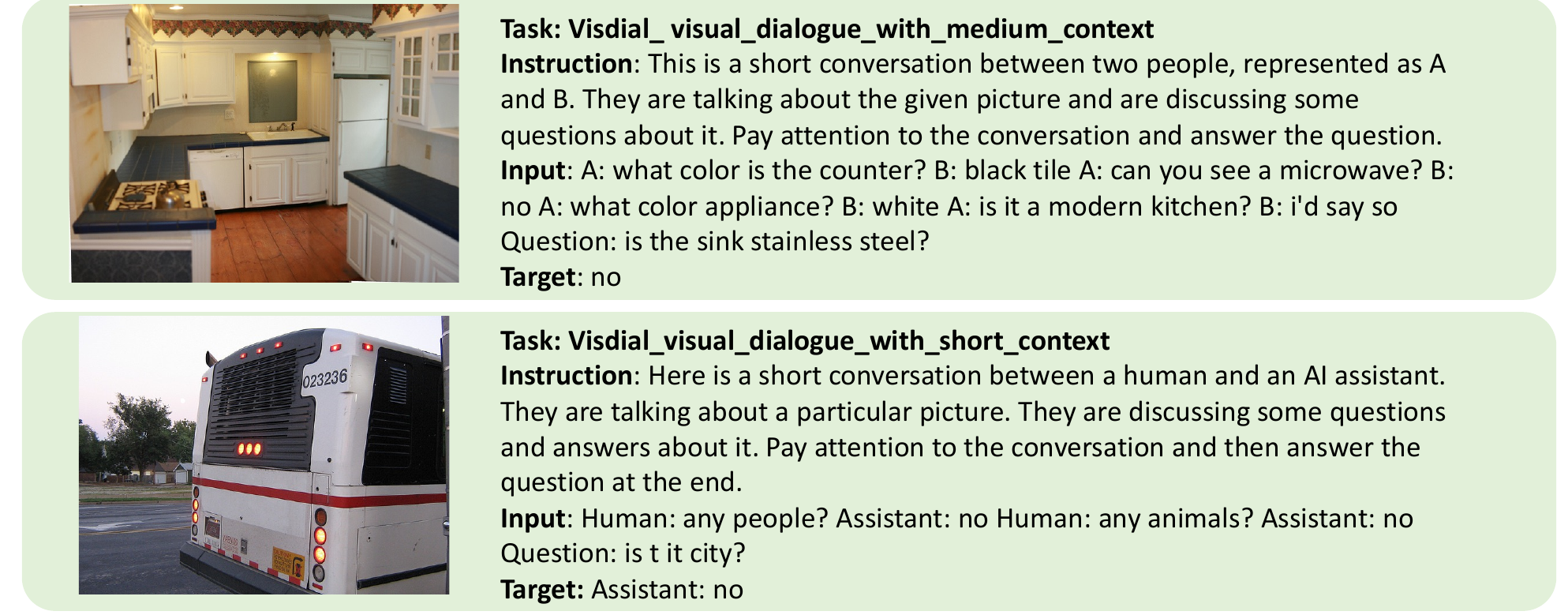}
   \caption{}
   \label{fig:}
\end{figure*}

\begin{figure*}[h!]
  \centering
   \includegraphics[width=\linewidth]{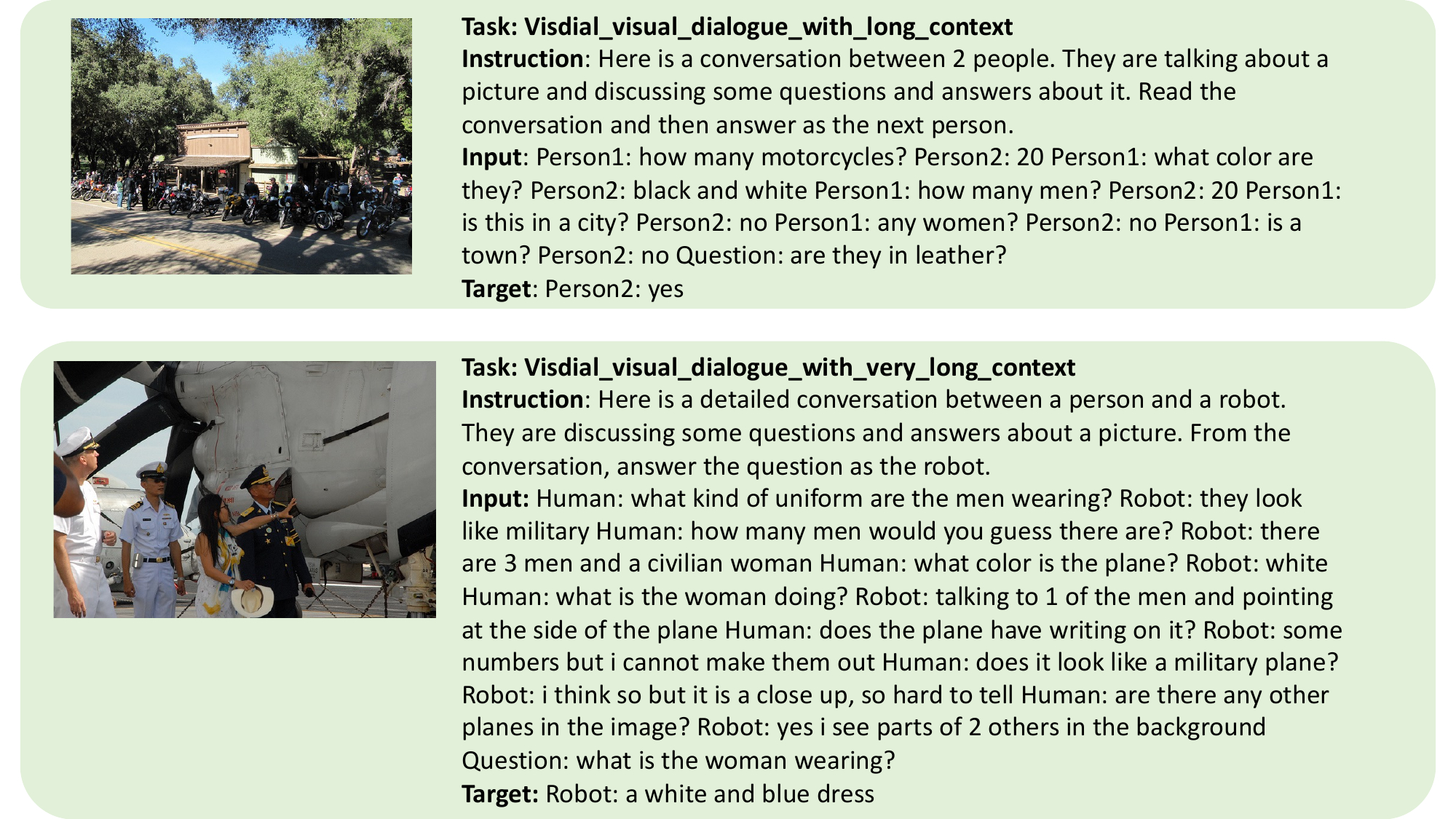}
   \caption{}
   \label{fig:}
\end{figure*}

\begin{figure*}[h!]
  \centering
   \includegraphics[width=\linewidth]{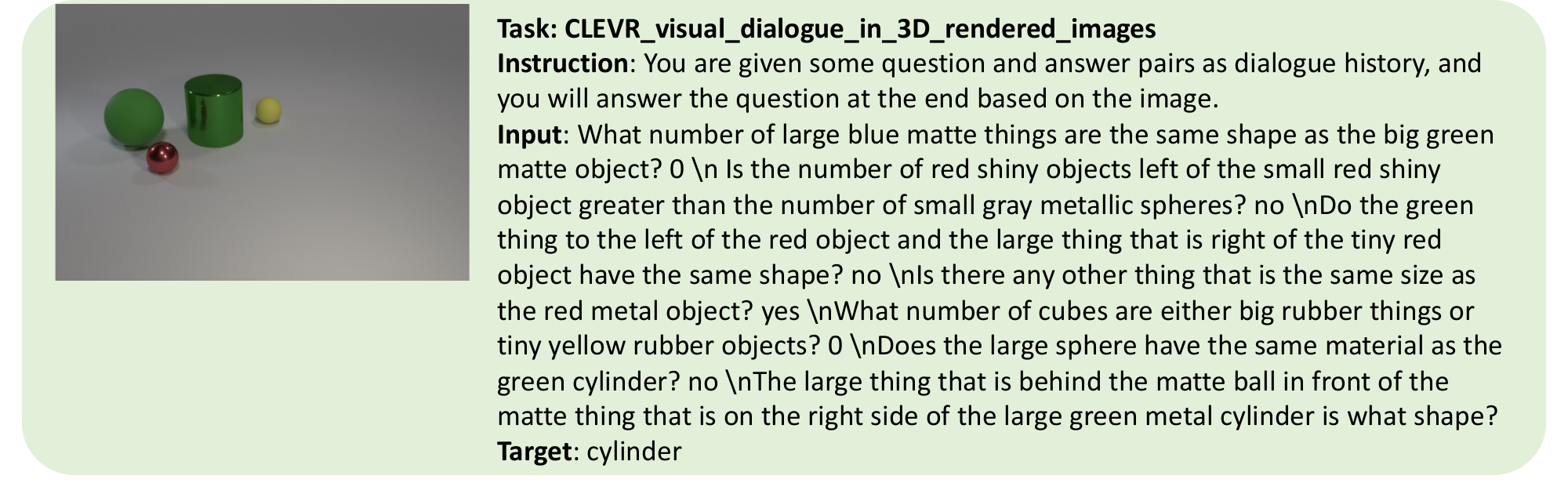}
   \caption{}
   \label{fig:}
\end{figure*}

\begin{figure*}[h!]
  \centering
   \includegraphics[width=\linewidth]{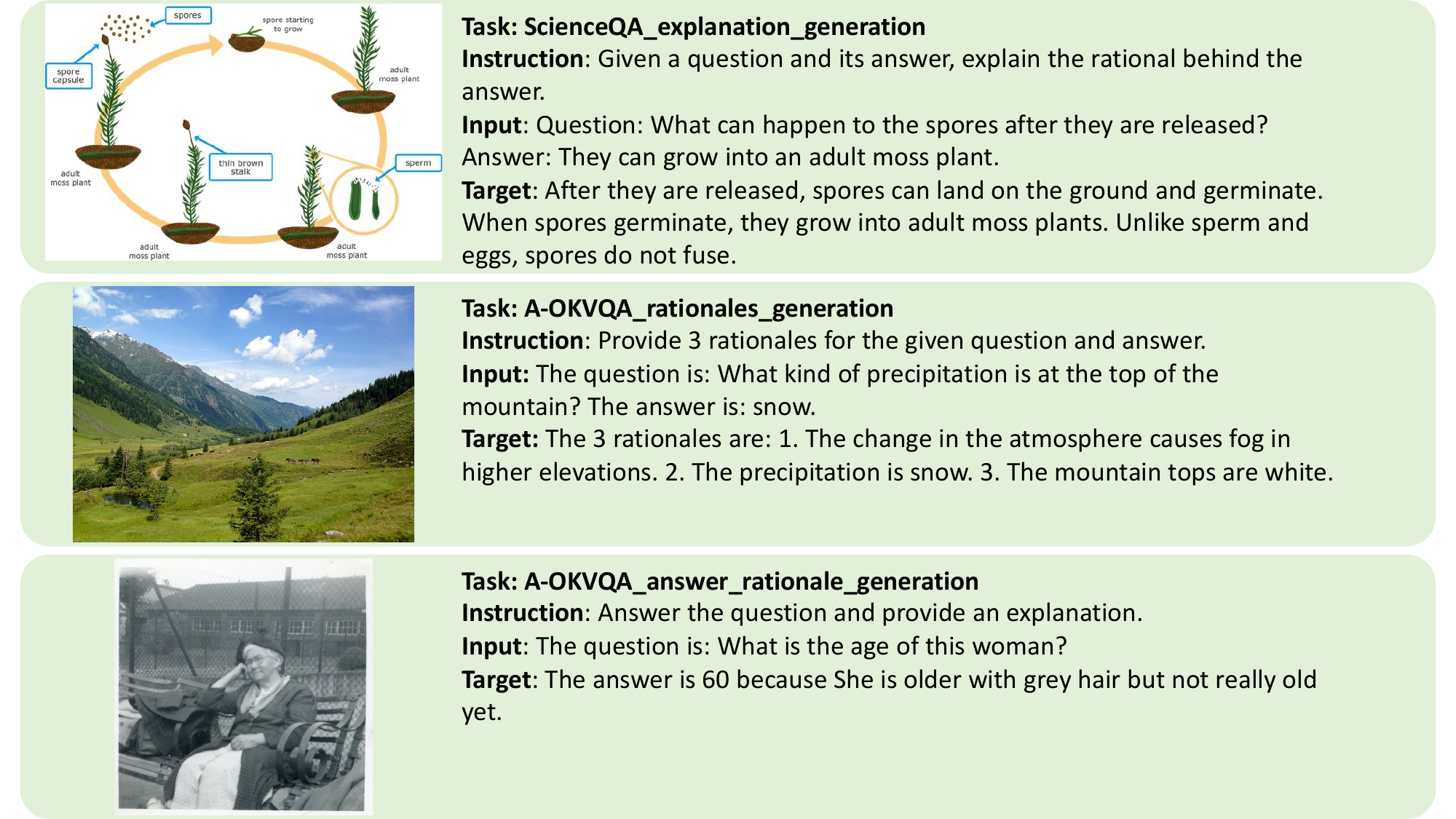}
   \caption{}
   \label{fig:}
\end{figure*}

\begin{figure*}[h!]
  \centering
   \includegraphics[width=\linewidth]{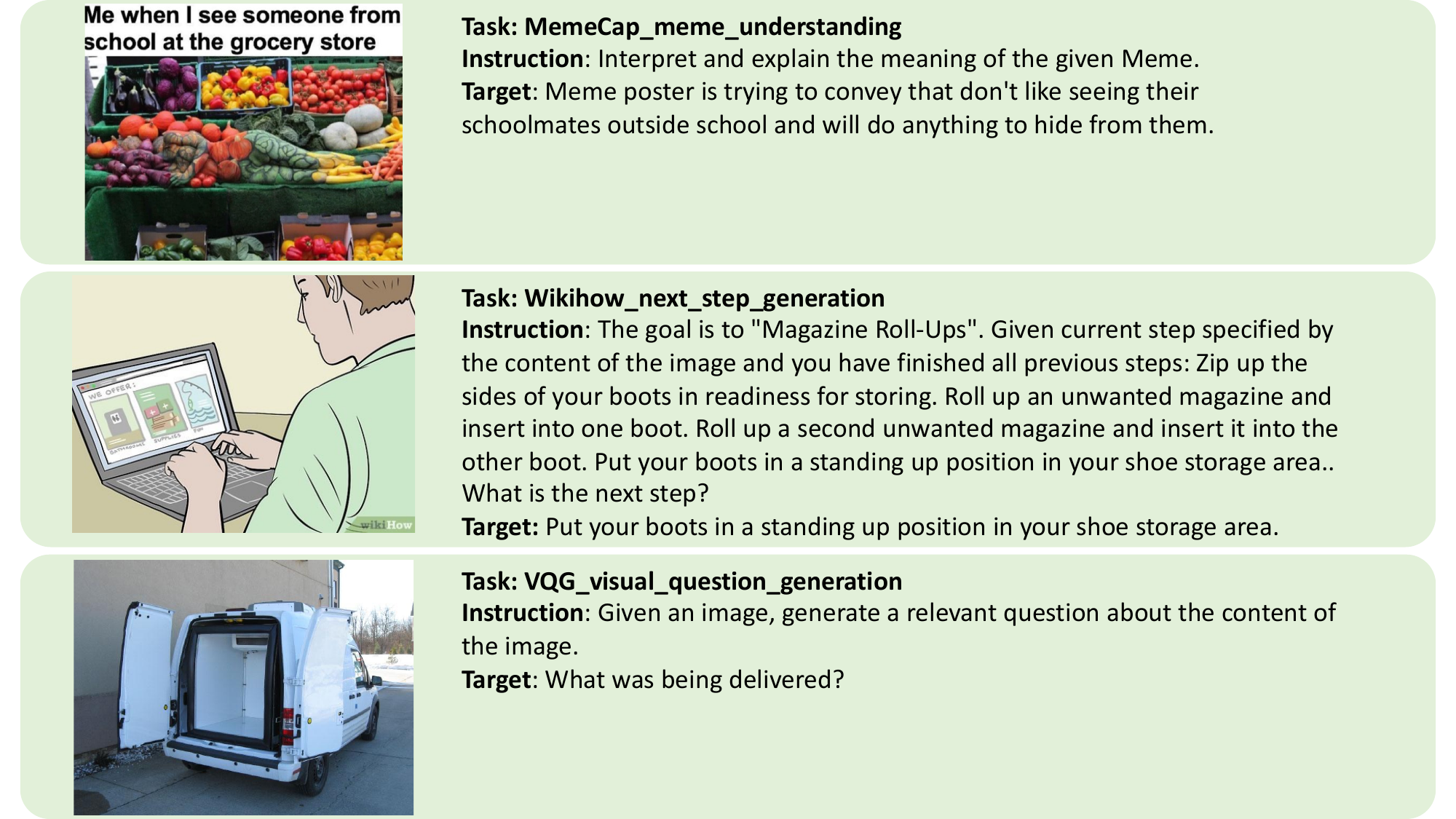}
   \caption{}
   \label{fig:}
\end{figure*}

\begin{figure*}[h!]
  \centering
   \includegraphics[width=\linewidth]{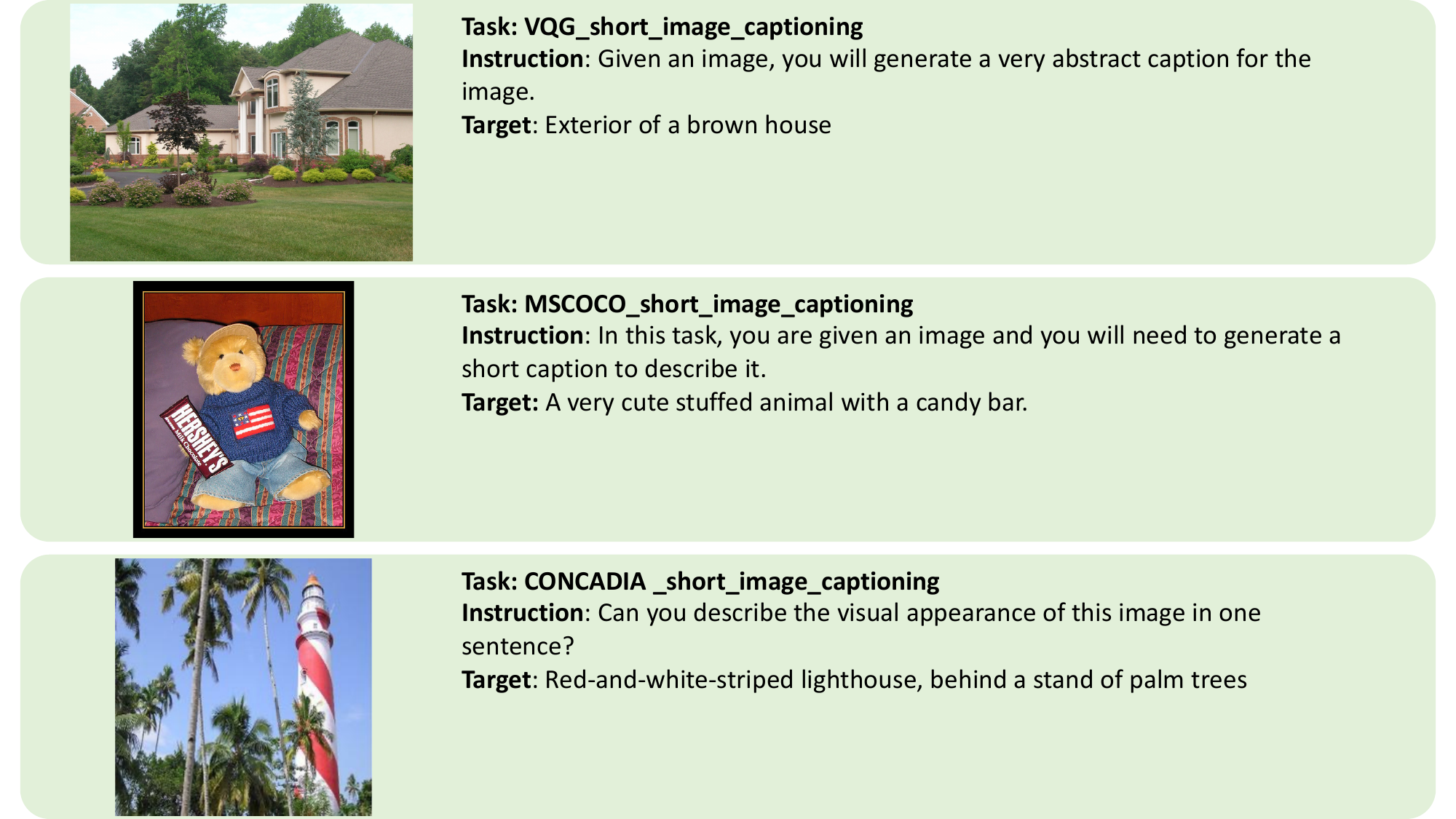}
   \caption{}
   \label{fig:}
\end{figure*}

\begin{figure*}[h!]
  \centering
   \includegraphics[width=\linewidth]{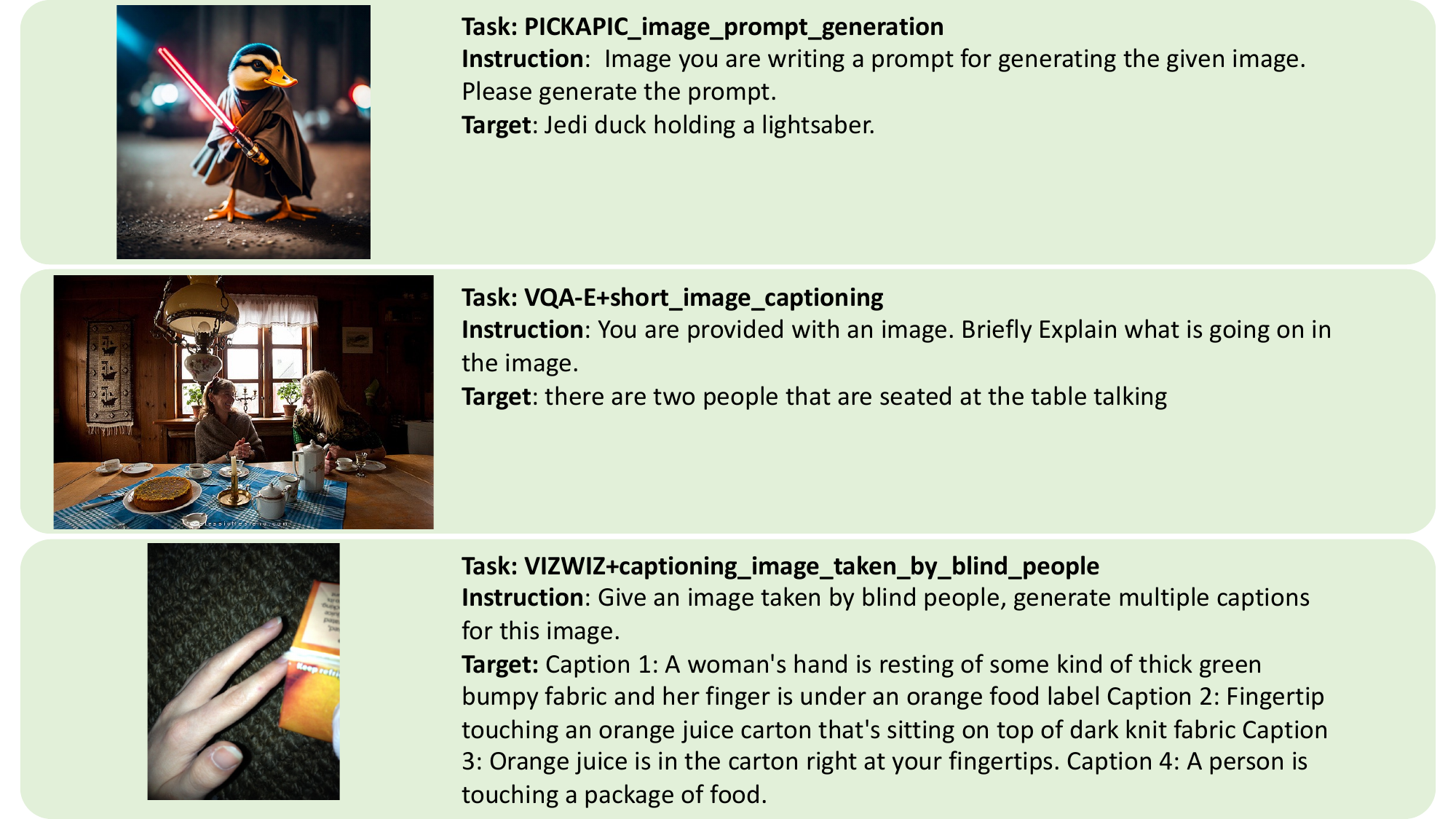}
   \caption{}
   \label{fig:}
\end{figure*}

\begin{figure*}[h!]
  \centering
   \includegraphics[width=\linewidth]{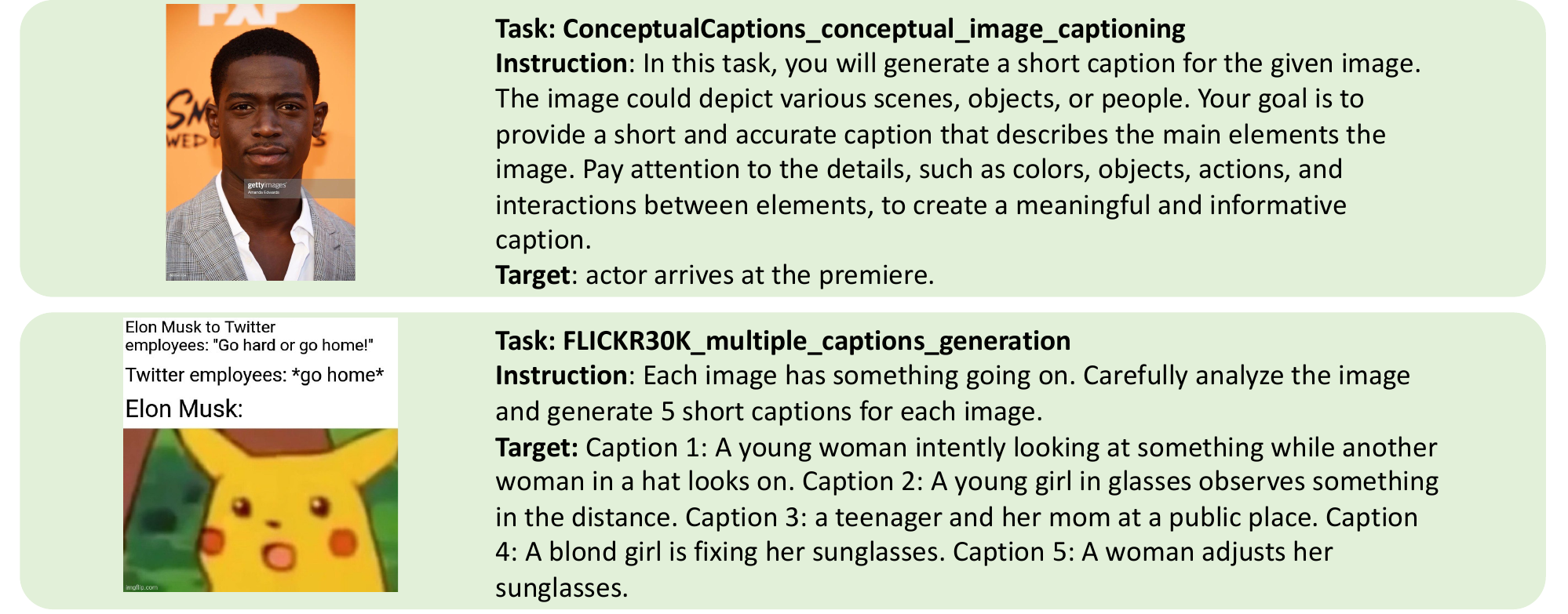}
   \caption{}
   \label{fig:}
\end{figure*}

\begin{figure*}[h!]
  \centering
   \includegraphics[width=\linewidth]{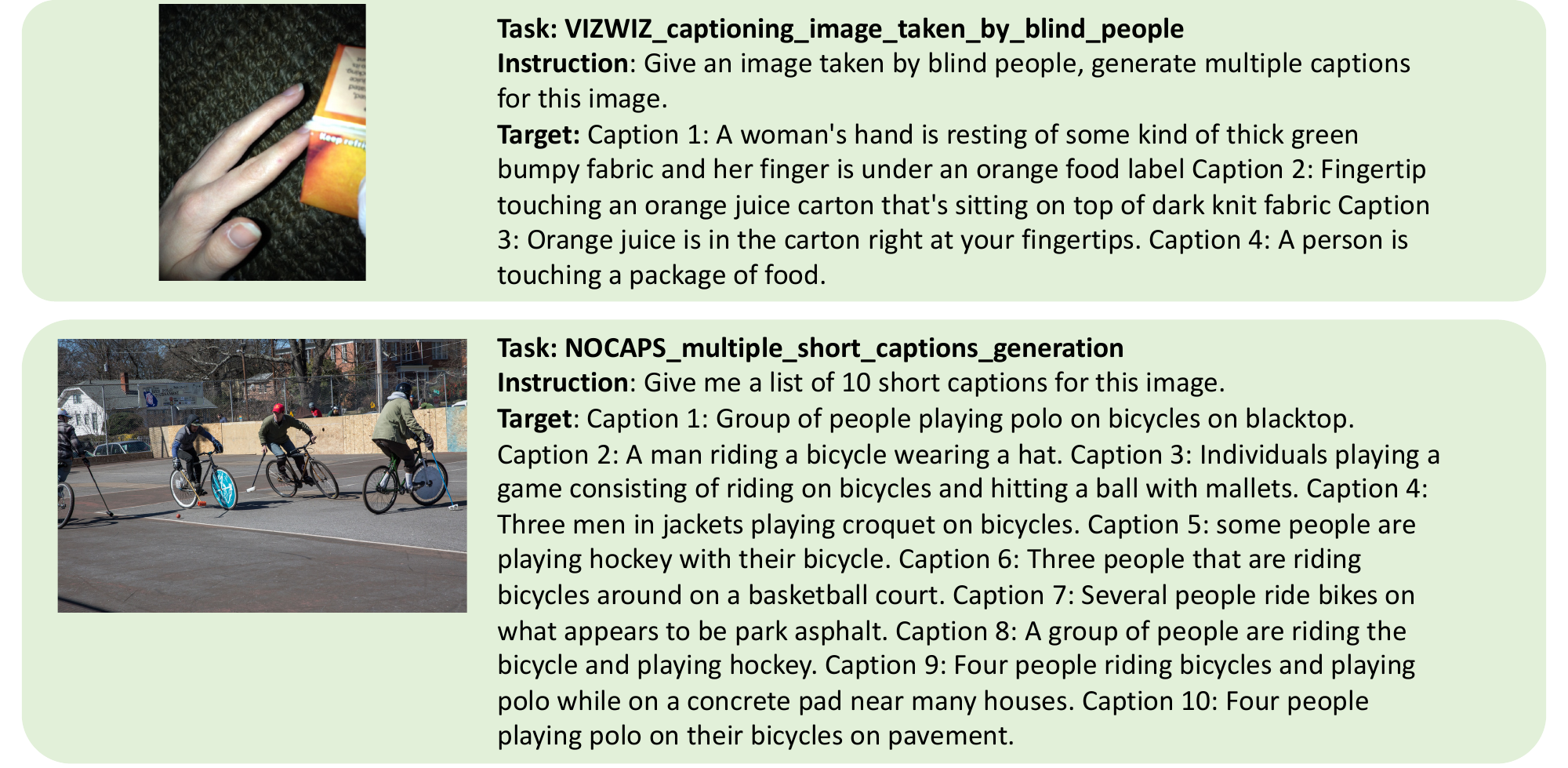}
   \caption{}
   \label{fig:}
\end{figure*}

\begin{figure*}[h!]
  \centering
   \includegraphics[width=\linewidth]{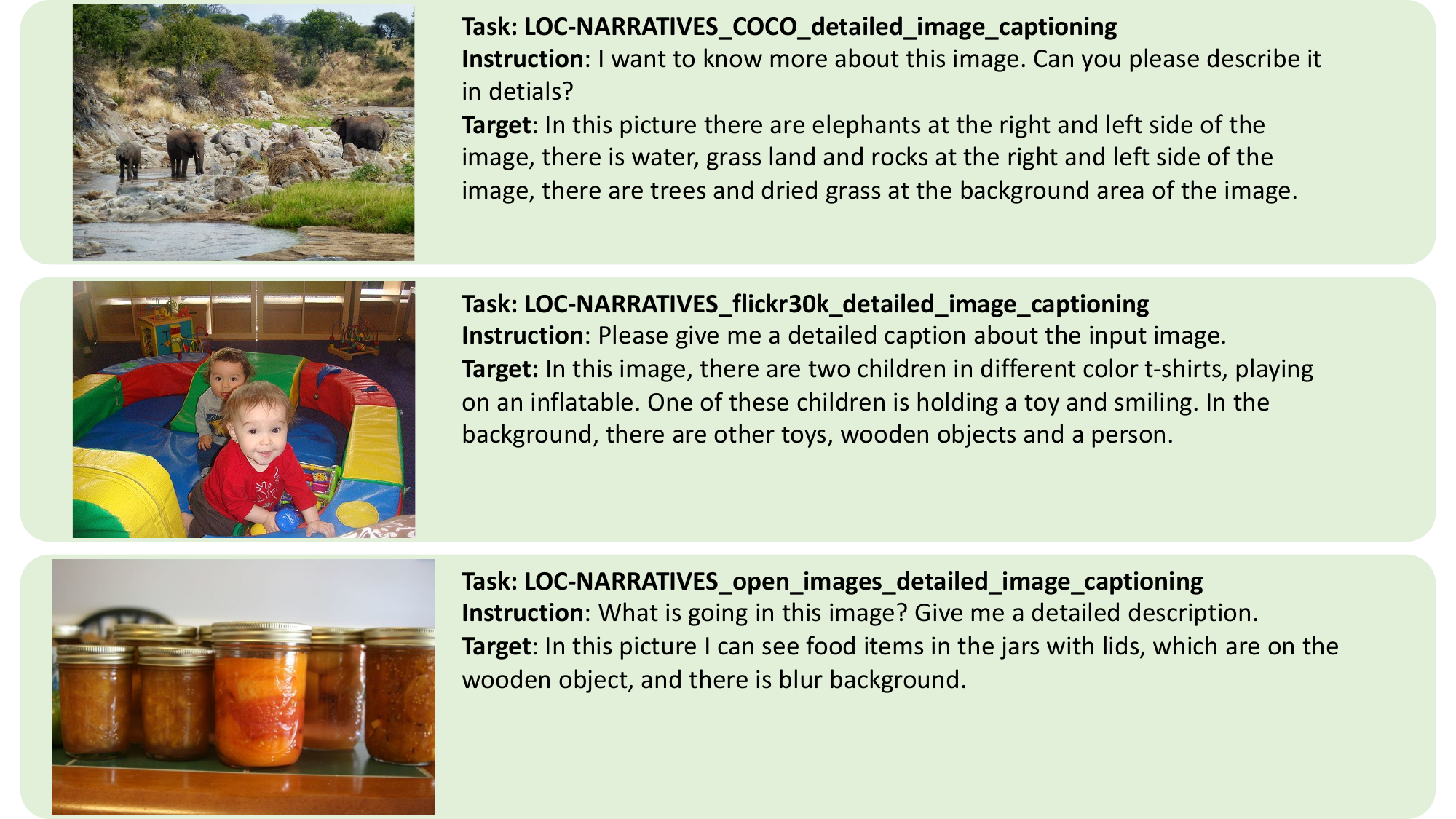}
   \caption{}
   \label{fig:}
\end{figure*}

\begin{figure*}[h!]
  \centering
   \includegraphics[width=\linewidth]{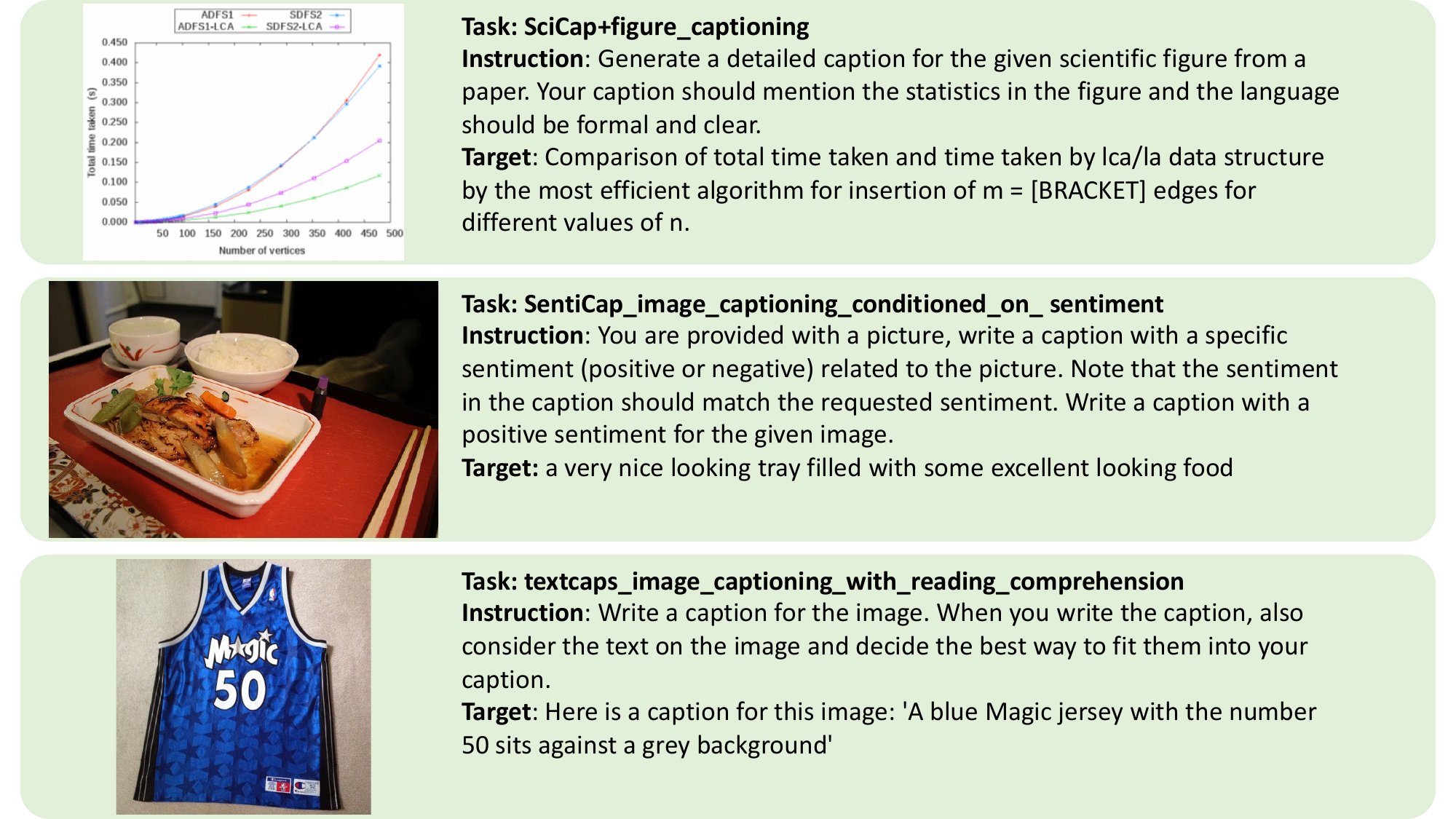}
   \caption{}
   \label{fig:}
\end{figure*}

\begin{figure*}[h!]
  \centering
   \includegraphics[width=\linewidth]{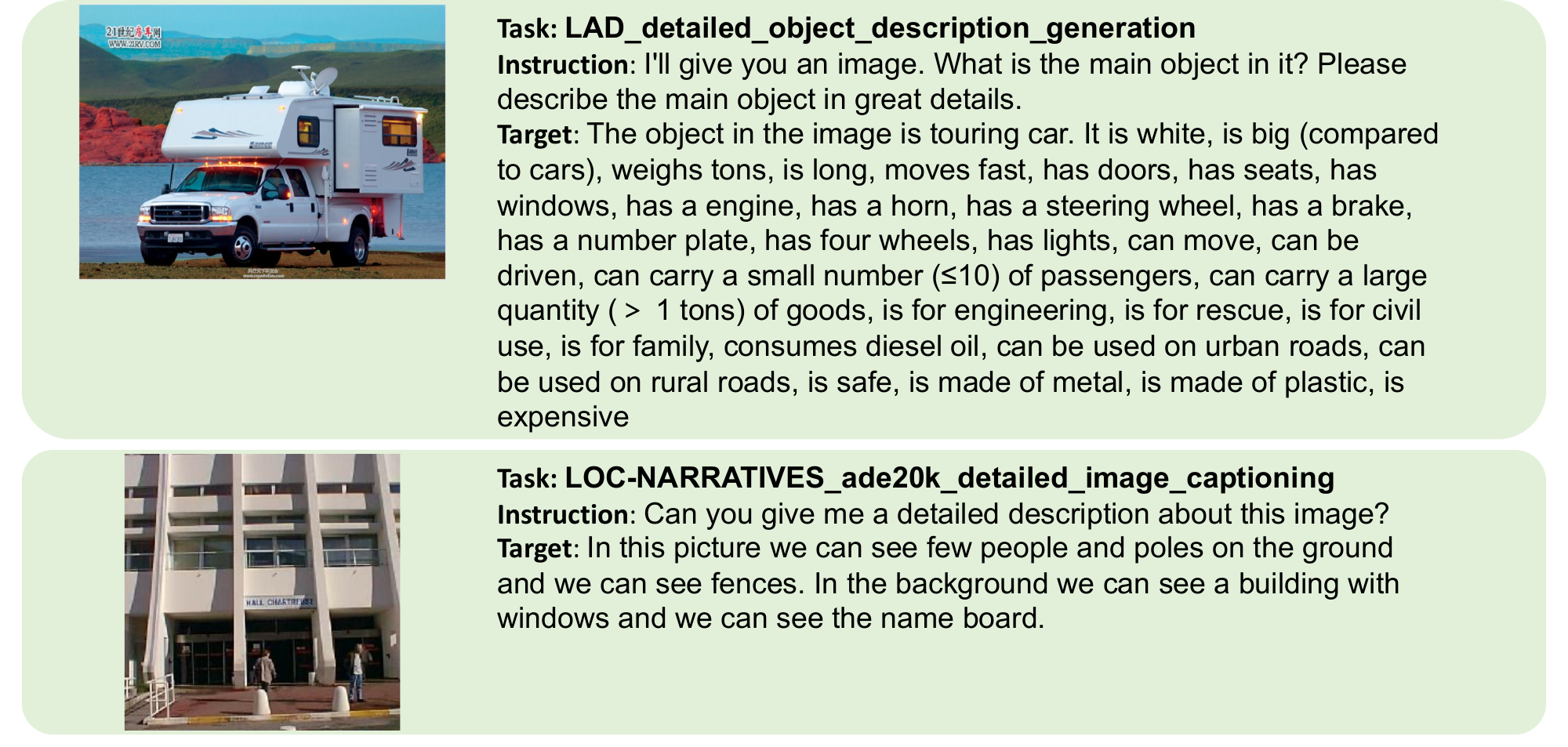}
   \caption{}
   \label{fig:}
\end{figure*}

\begin{figure*}[h!]
  \centering
   \includegraphics[width=\linewidth]{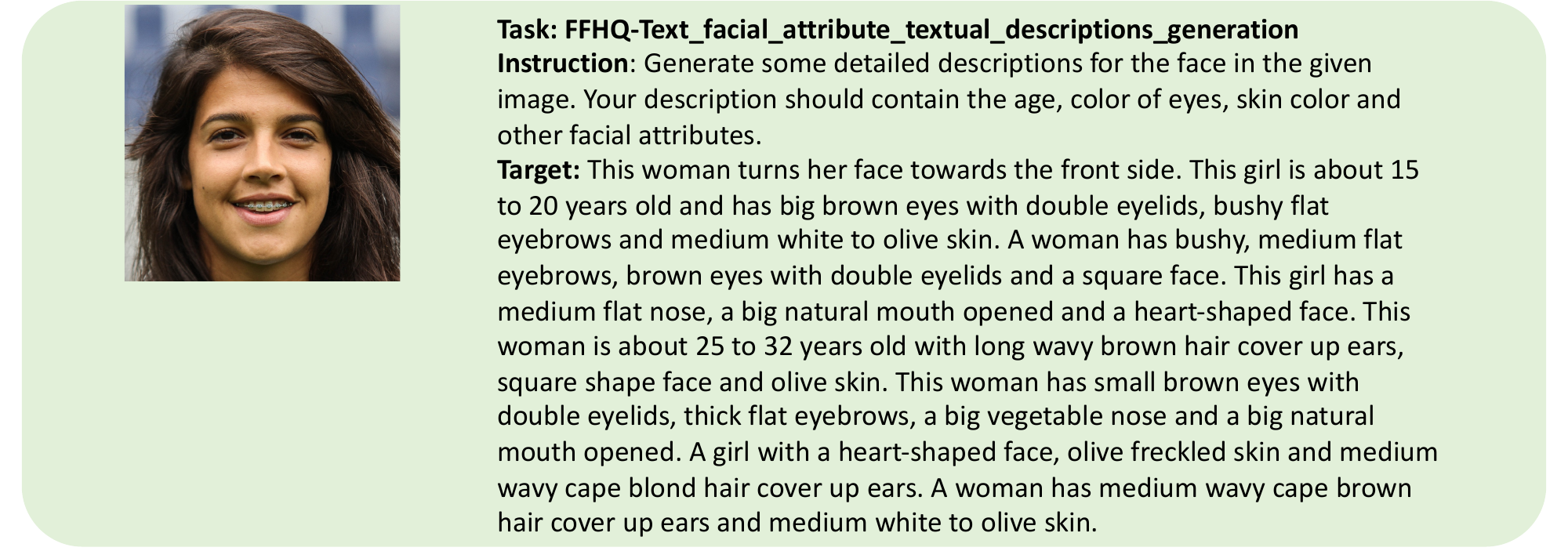}
   \caption{}
   \label{fig:}
\end{figure*}

\clearpage
\subsection{Classification Tasks}

\begin{figure*}[h!]
  \centering
   \includegraphics[width=\linewidth]{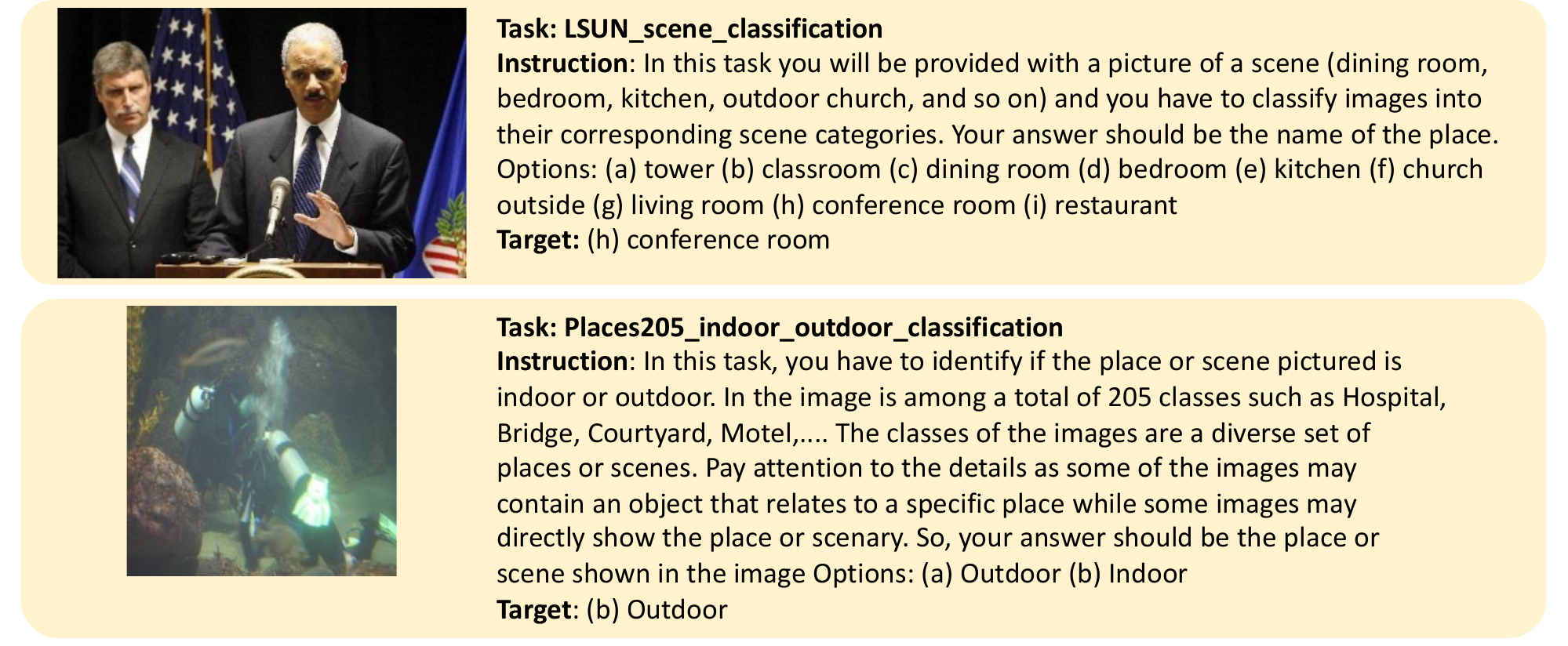}
   \caption{}
   \label{fig:}
\end{figure*}

\begin{figure*}[h!]
  \centering
   \includegraphics[width=\linewidth]{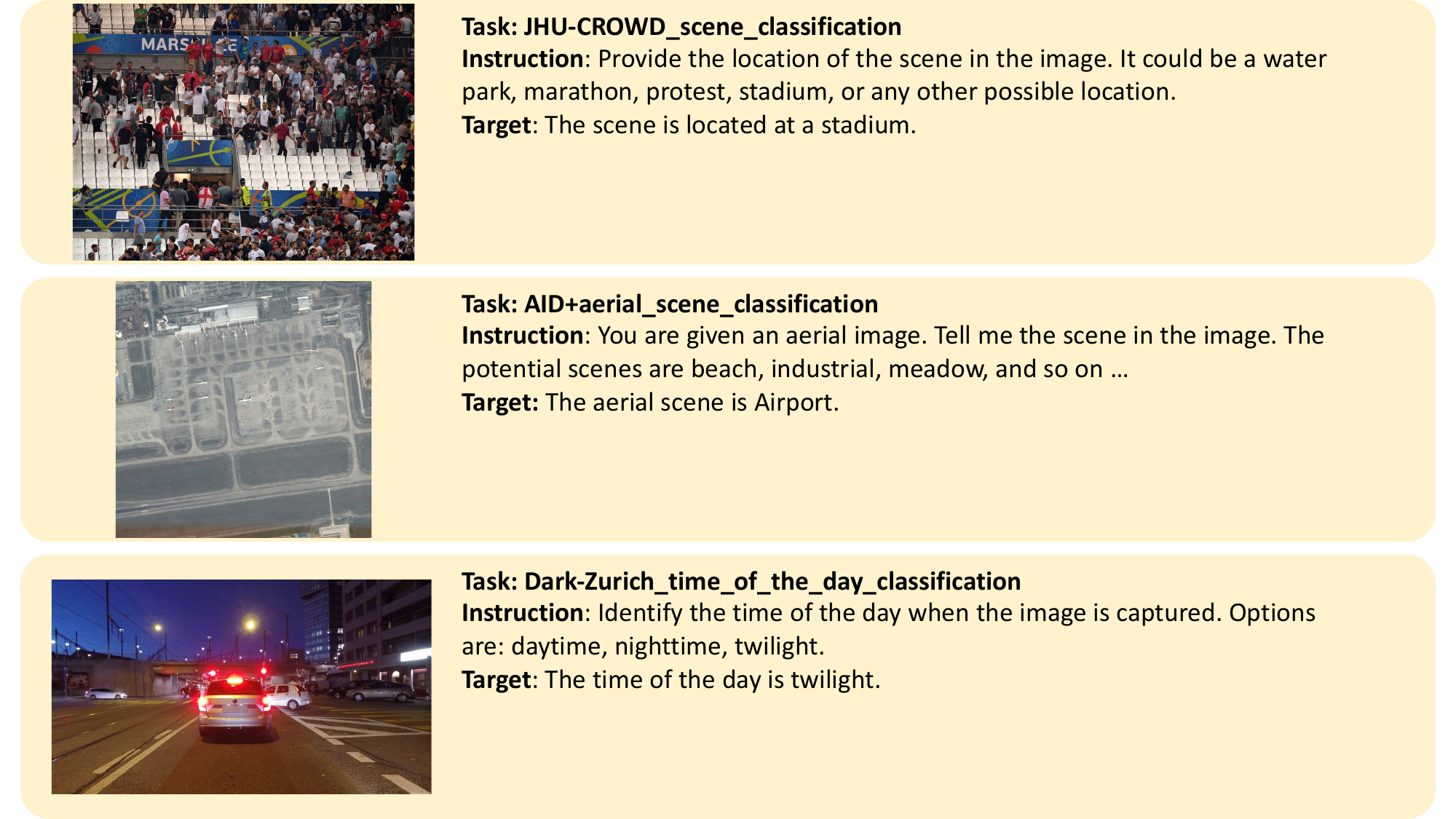}
   \caption{}
   \label{fig:}
\end{figure*}

\begin{figure*}[h!]
  \centering
   \includegraphics[width=\linewidth]{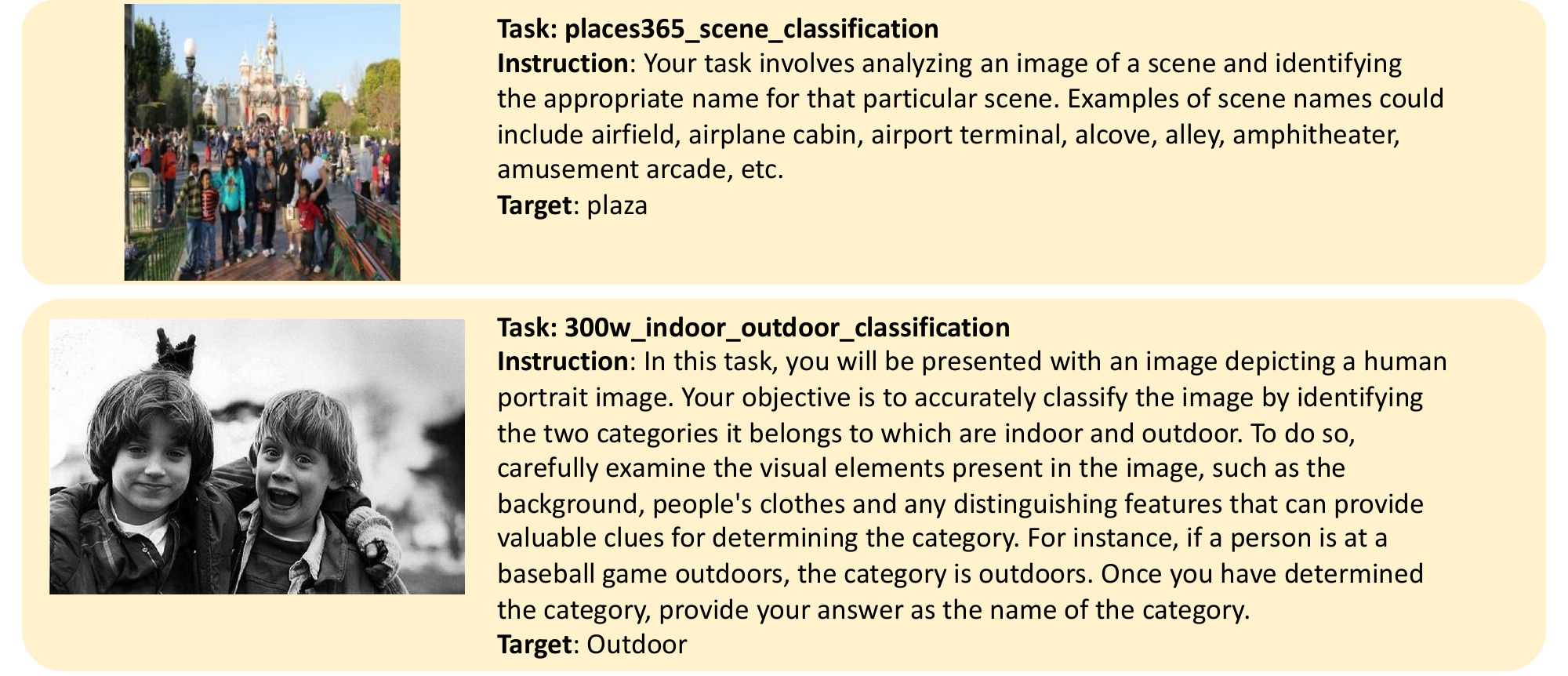}
   \caption{}
   \label{fig:}
\end{figure*}

\begin{figure*}[h!]
  \centering
   \includegraphics[width=\linewidth]{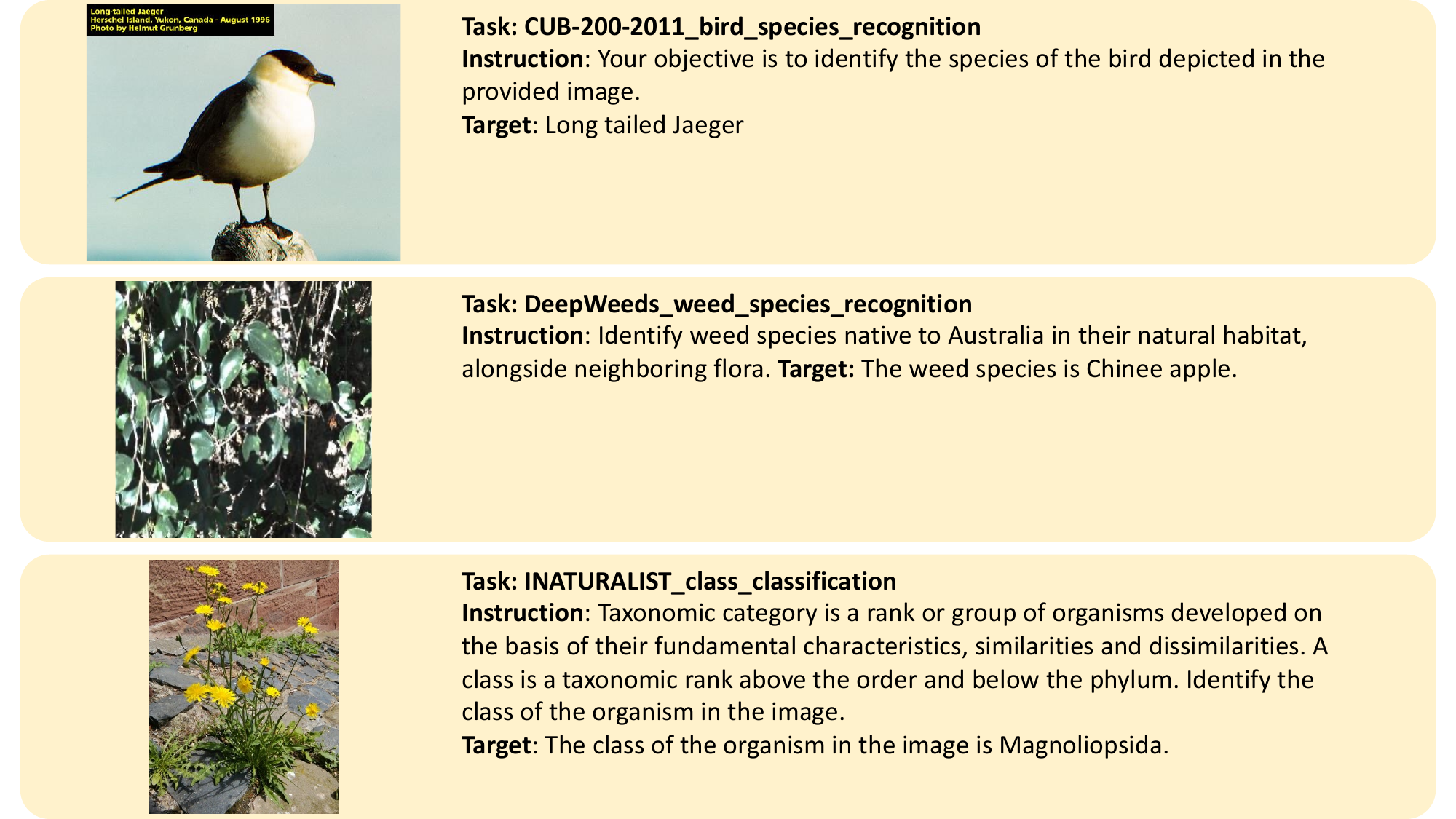}
   \caption{}
   \label{fig:}
\end{figure*}

\begin{figure*}[h!]
  \centering
   \includegraphics[width=\linewidth]{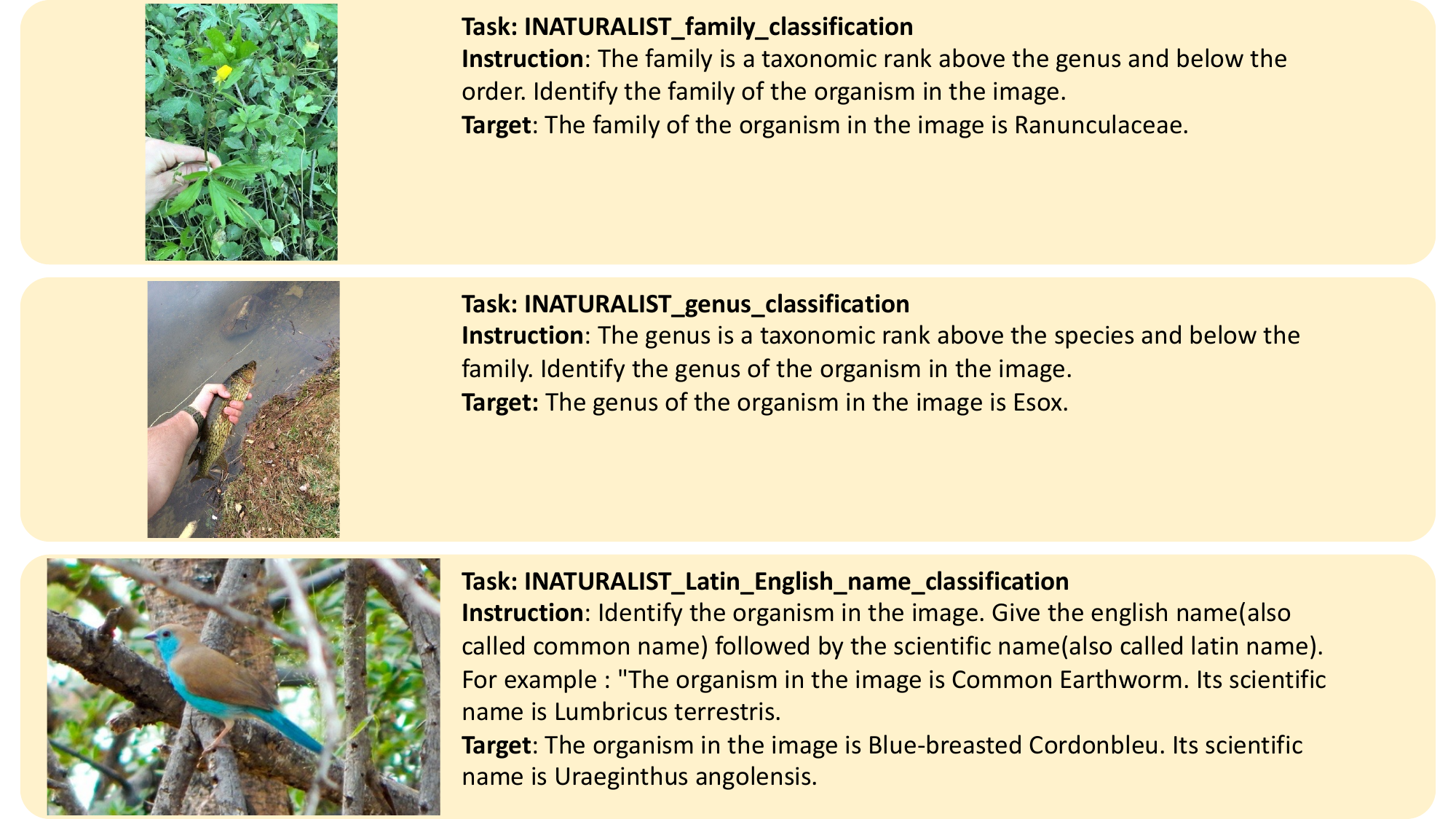}
   \caption{}
   \label{fig:}
\end{figure*}

\begin{figure*}[h!]
  \centering
   \includegraphics[width=\linewidth]{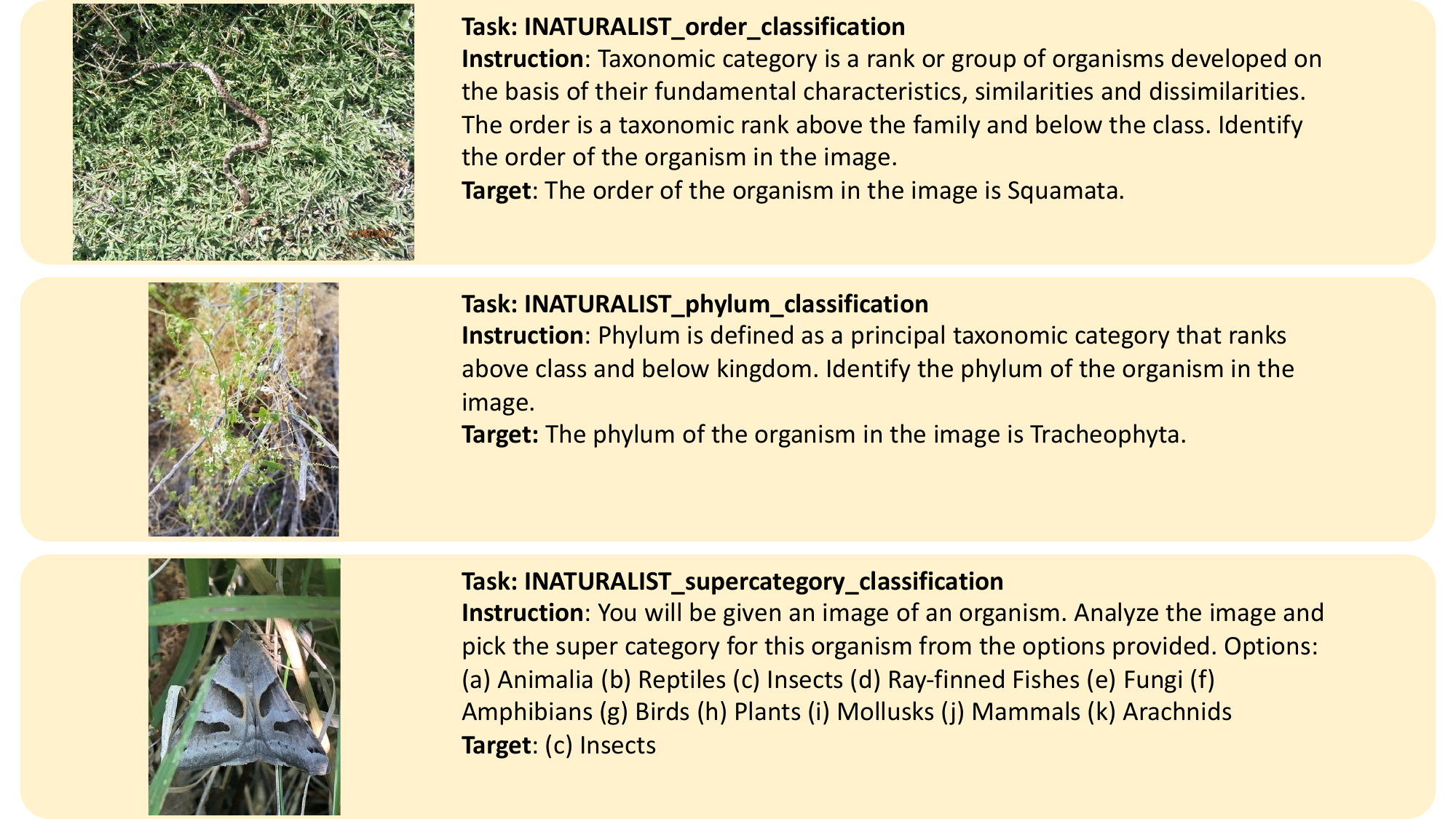}
   \caption{}
   \label{fig:}
\end{figure*}

\begin{figure*}[h!]
  \centering
   \includegraphics[width=\linewidth]{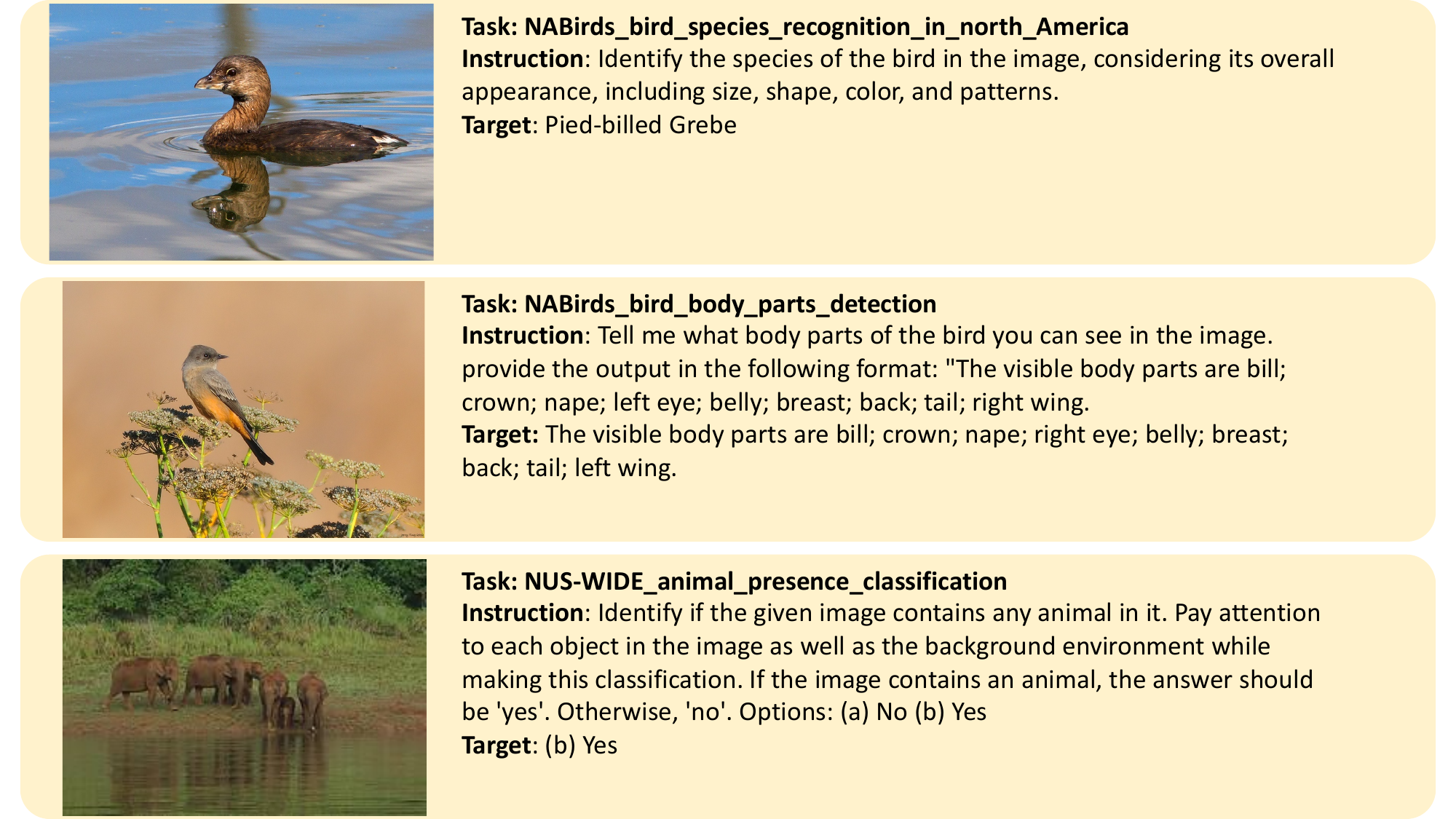}
   \caption{}
   \label{fig:}
\end{figure*}

\begin{figure*}[h!]
  \centering
   \includegraphics[width=\linewidth]{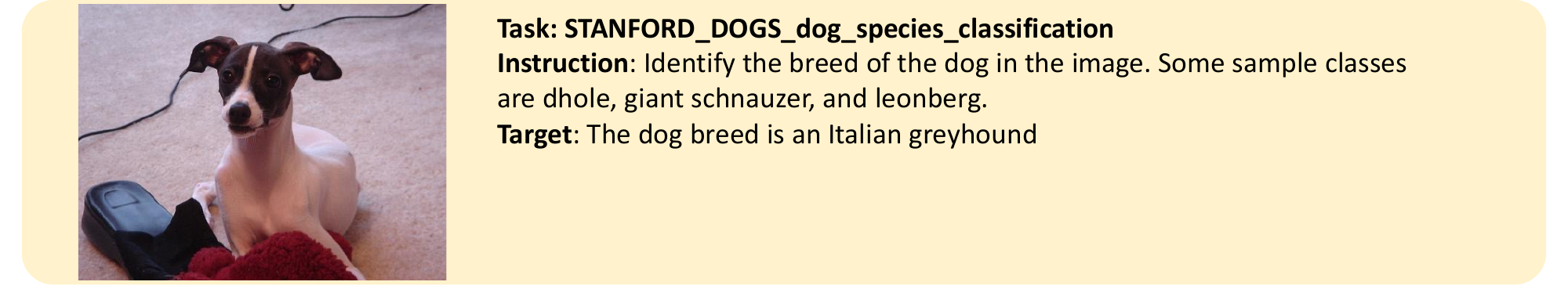}
   \caption{}
   \label{fig:}
\end{figure*}

\begin{figure*}[h!]
  \centering
   \includegraphics[width=\linewidth]{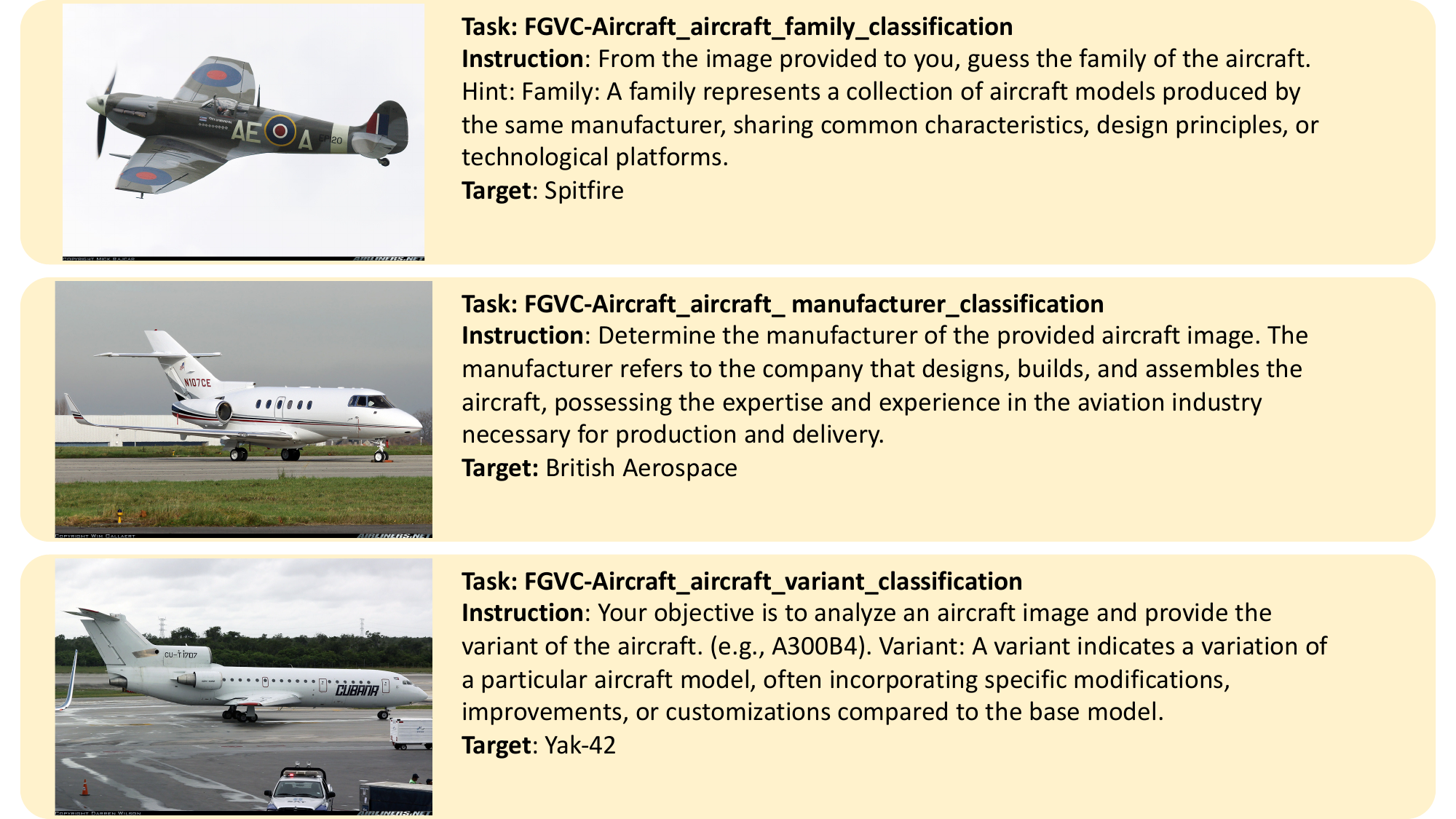}
   \caption{}
   \label{fig:}
\end{figure*}

\begin{figure*}[h!]
  \centering
   \includegraphics[width=\linewidth]{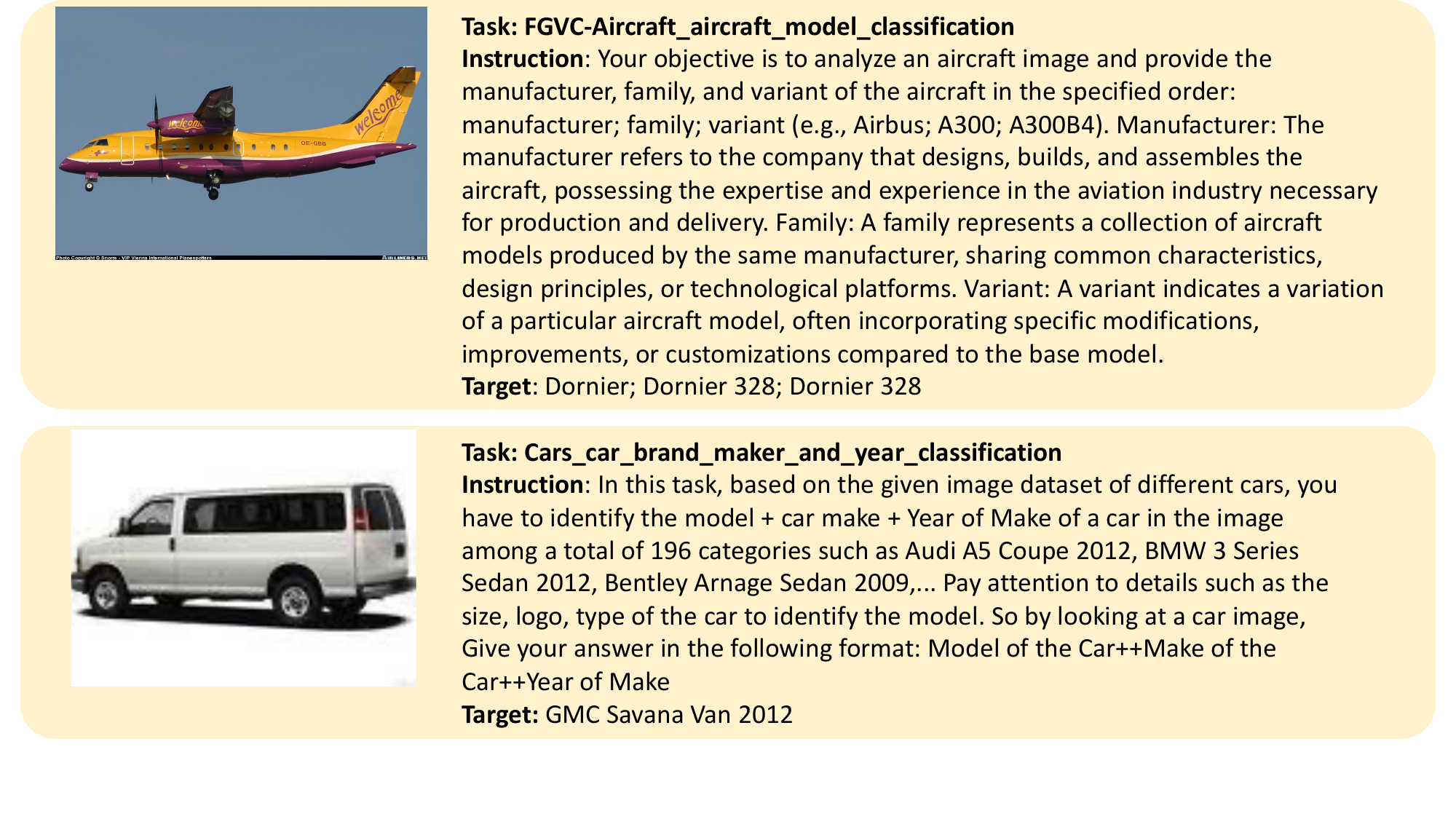}
   \caption{}
   \label{fig:}
\end{figure*}

\begin{figure*}[h!]
  \centering
   \includegraphics[width=\linewidth]{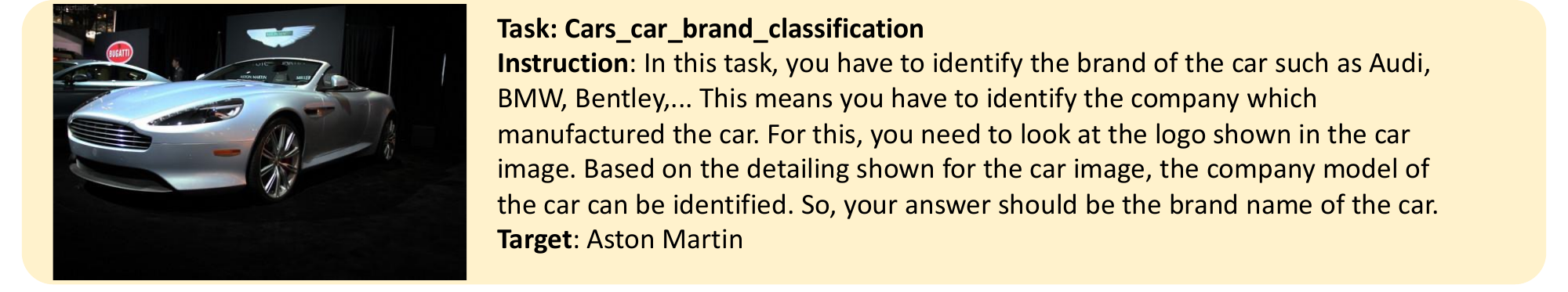}
   \caption{}
   \label{fig:}
\end{figure*}

\begin{figure*}[h!]
  \centering
   \includegraphics[width=\linewidth]{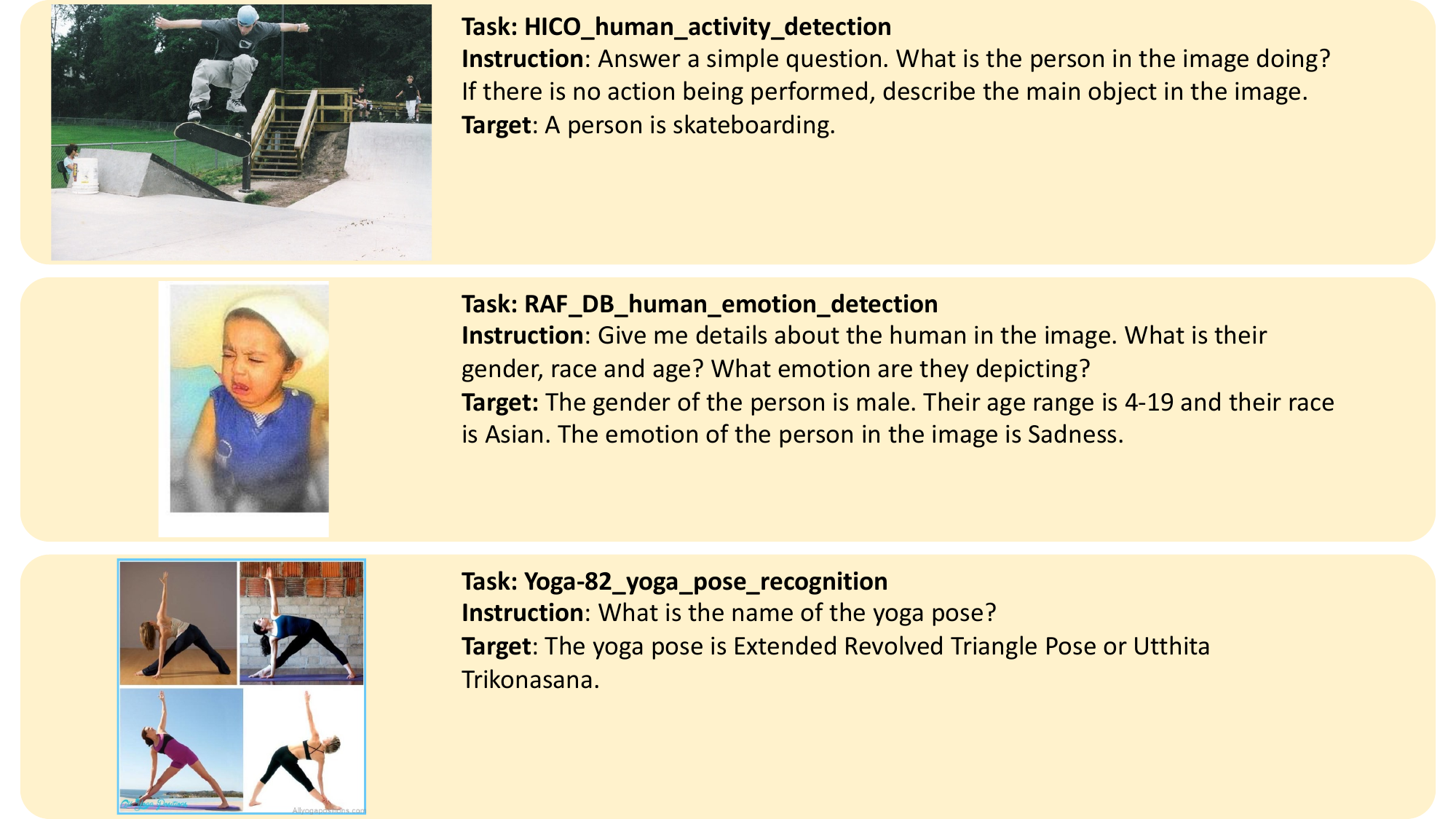}
   \caption{}
   \label{fig:}
\end{figure*}

\begin{figure*}[h!]
  \centering
   \includegraphics[width=\linewidth]{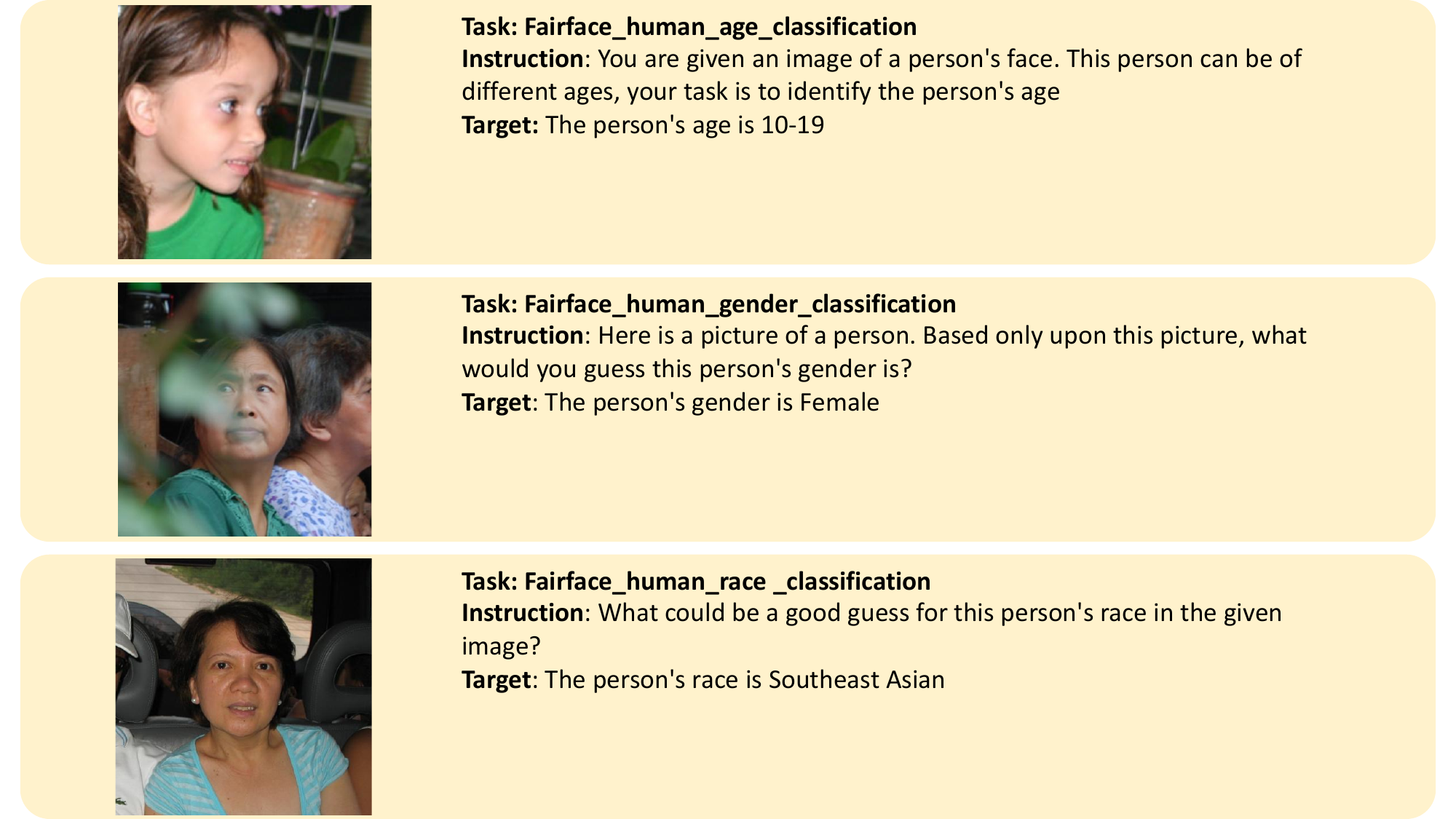}
   \caption{}
   \label{fig:}
\end{figure*}

\begin{figure*}[h!]
  \centering
   \includegraphics[width=\linewidth]{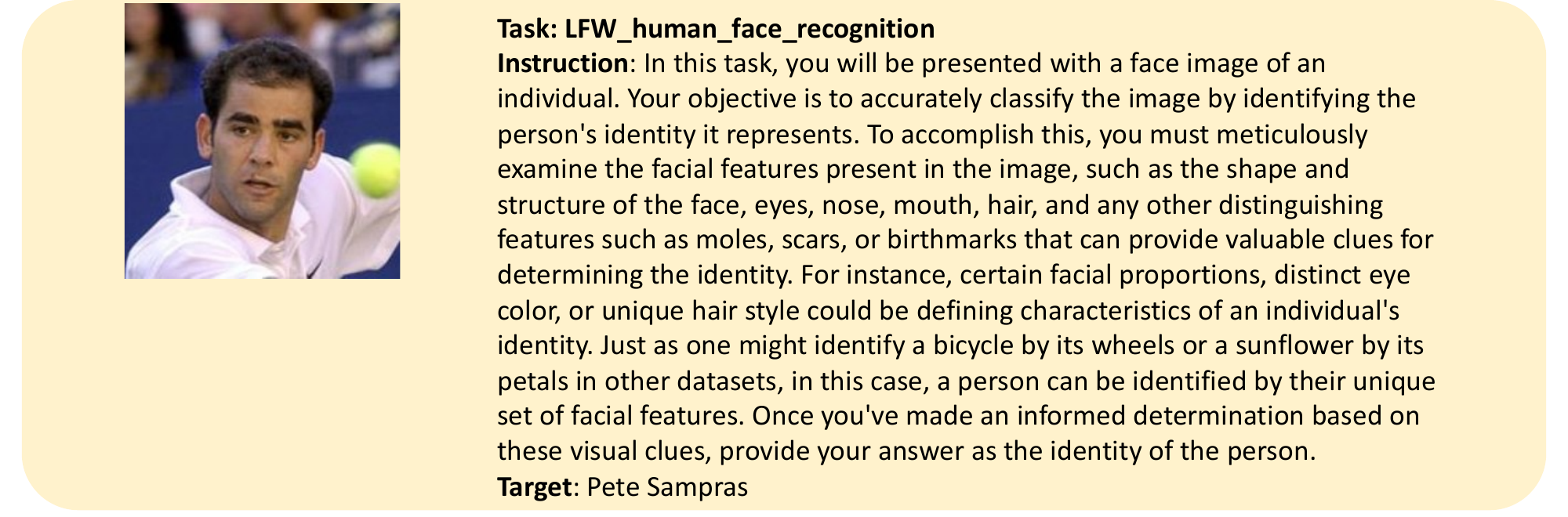}
   \caption{}
   \label{fig:}
\end{figure*}

\begin{figure*}[h!]
  \centering
   \includegraphics[width=\linewidth]{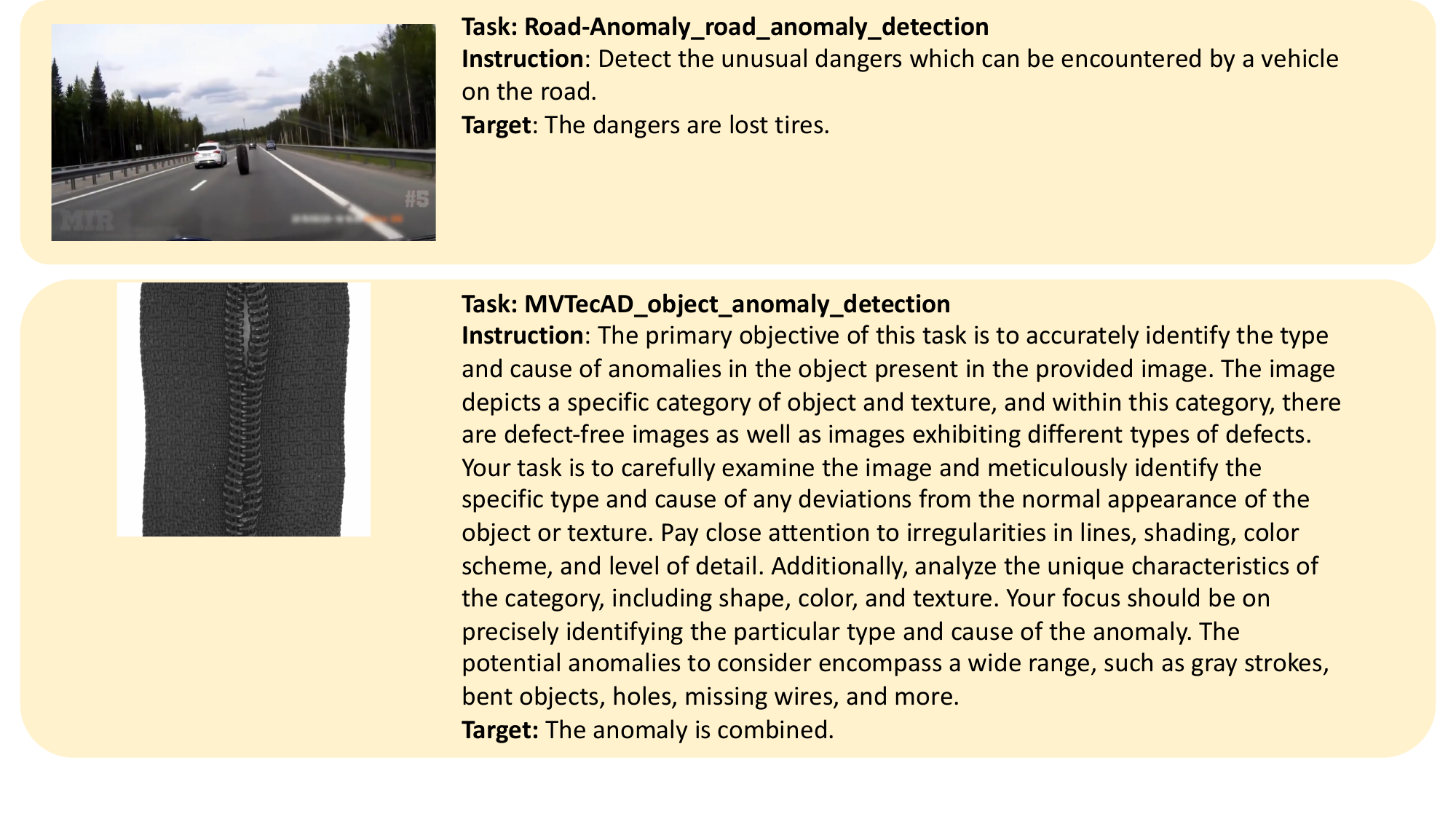}
   \caption{}
   \label{fig:}
\end{figure*}

\begin{figure*}[h!]
  \centering
   \includegraphics[width=\linewidth]{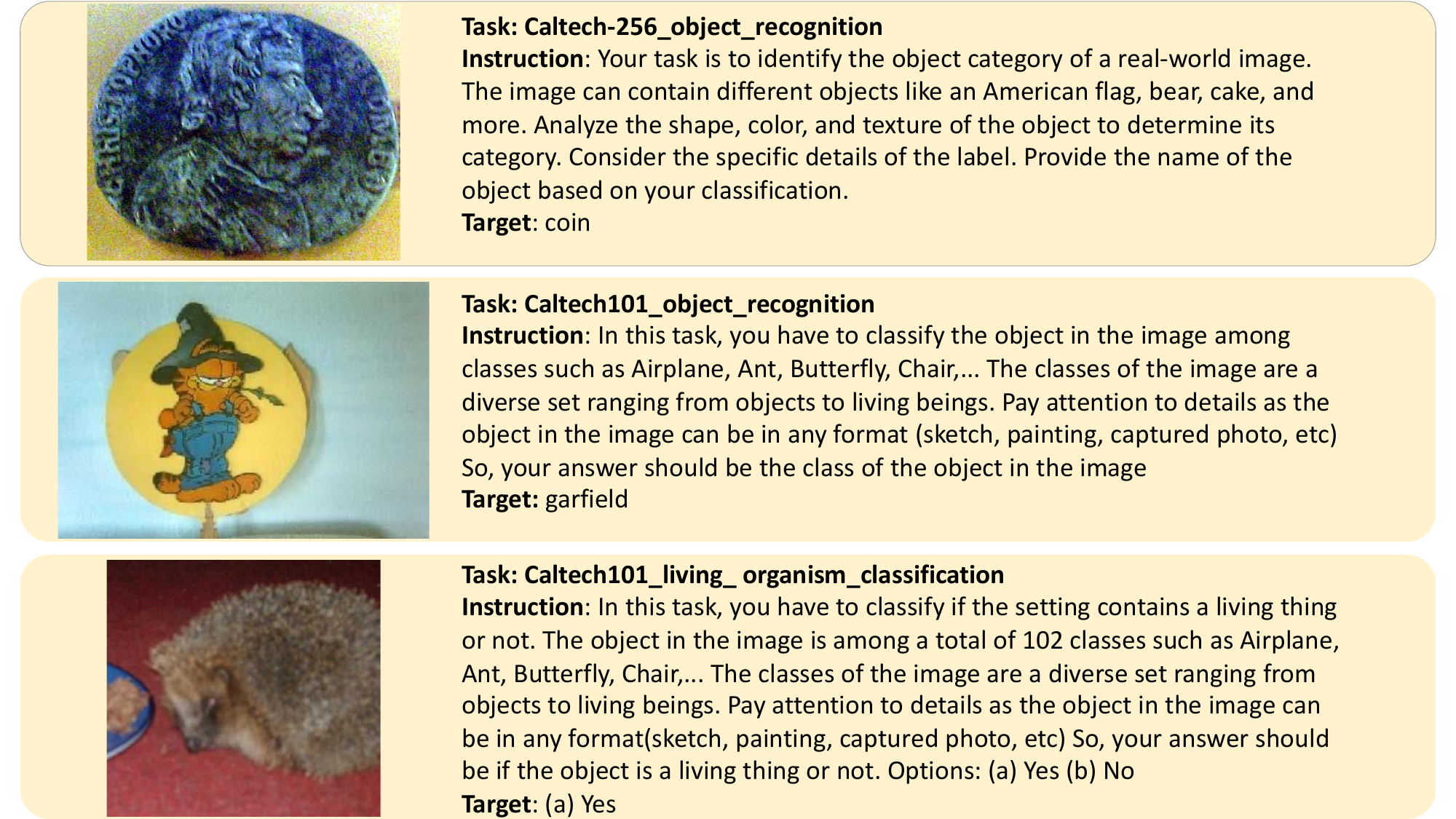}
   \caption{}
   \label{fig:}
\end{figure*}

\begin{figure*}[h!]
  \centering
   \includegraphics[width=\linewidth]{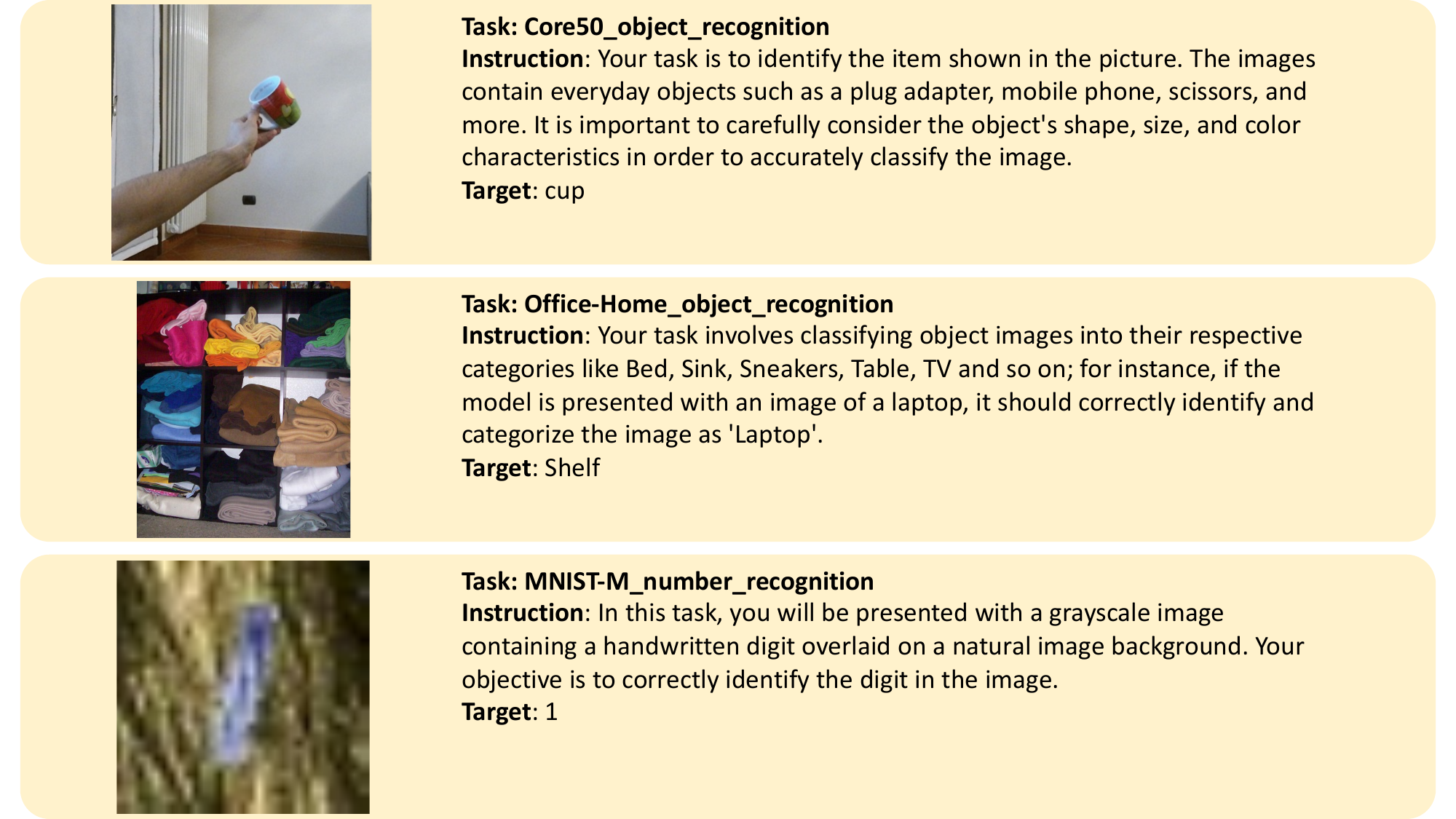}
   \caption{}
   \label{fig:}
\end{figure*}

\begin{figure*}[h!]
  \centering
   \includegraphics[width=\linewidth]{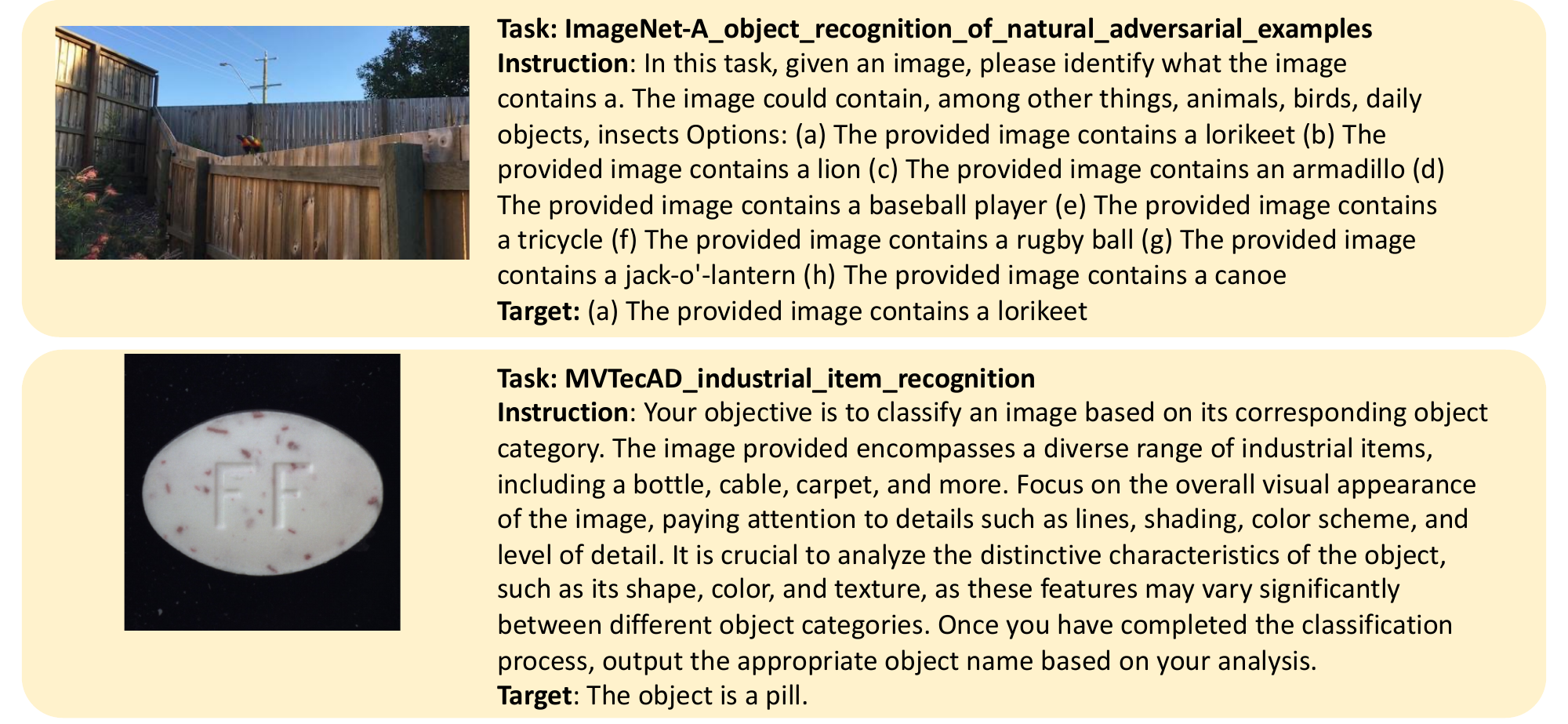}
   \caption{}
   \label{fig:}
\end{figure*}

\begin{figure*}[h!]
  \centering
   \includegraphics[width=\linewidth]{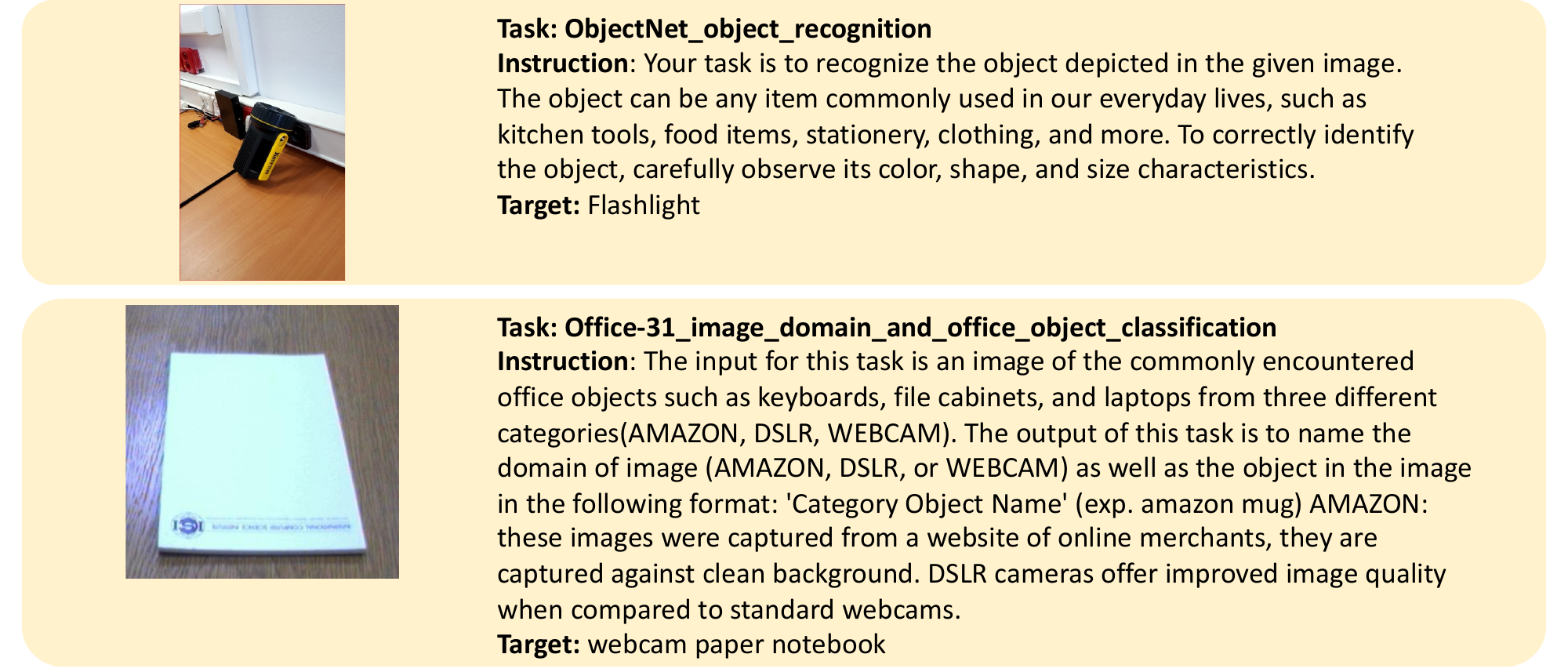}
   \caption{}
   \label{fig:}
\end{figure*}

\begin{figure*}[h!]
  \centering
   \includegraphics[width=\linewidth]{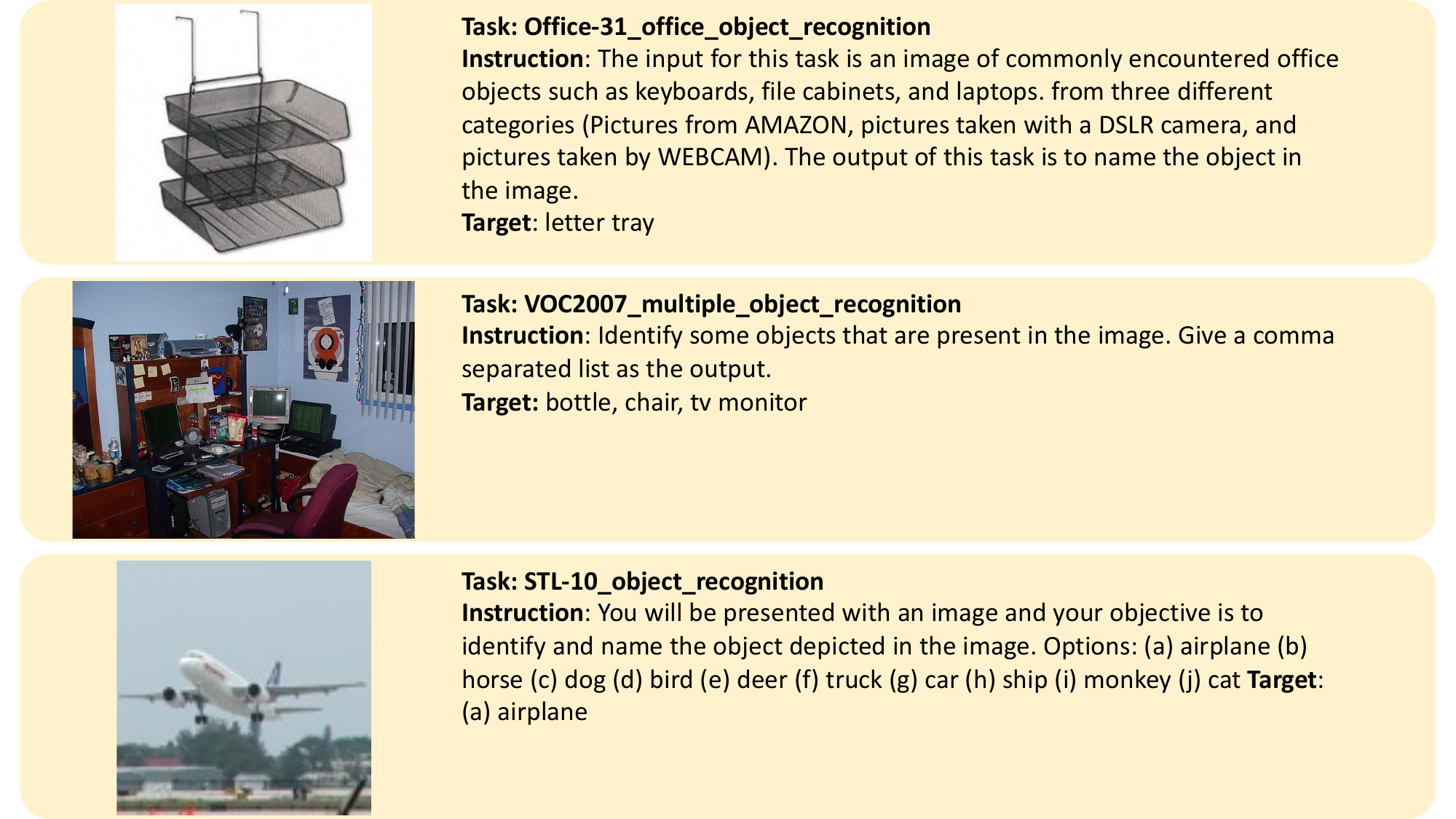}
   \caption{}
   \label{fig:}
\end{figure*}

\begin{figure*}[h!]
  \centering
   \includegraphics[width=\linewidth]{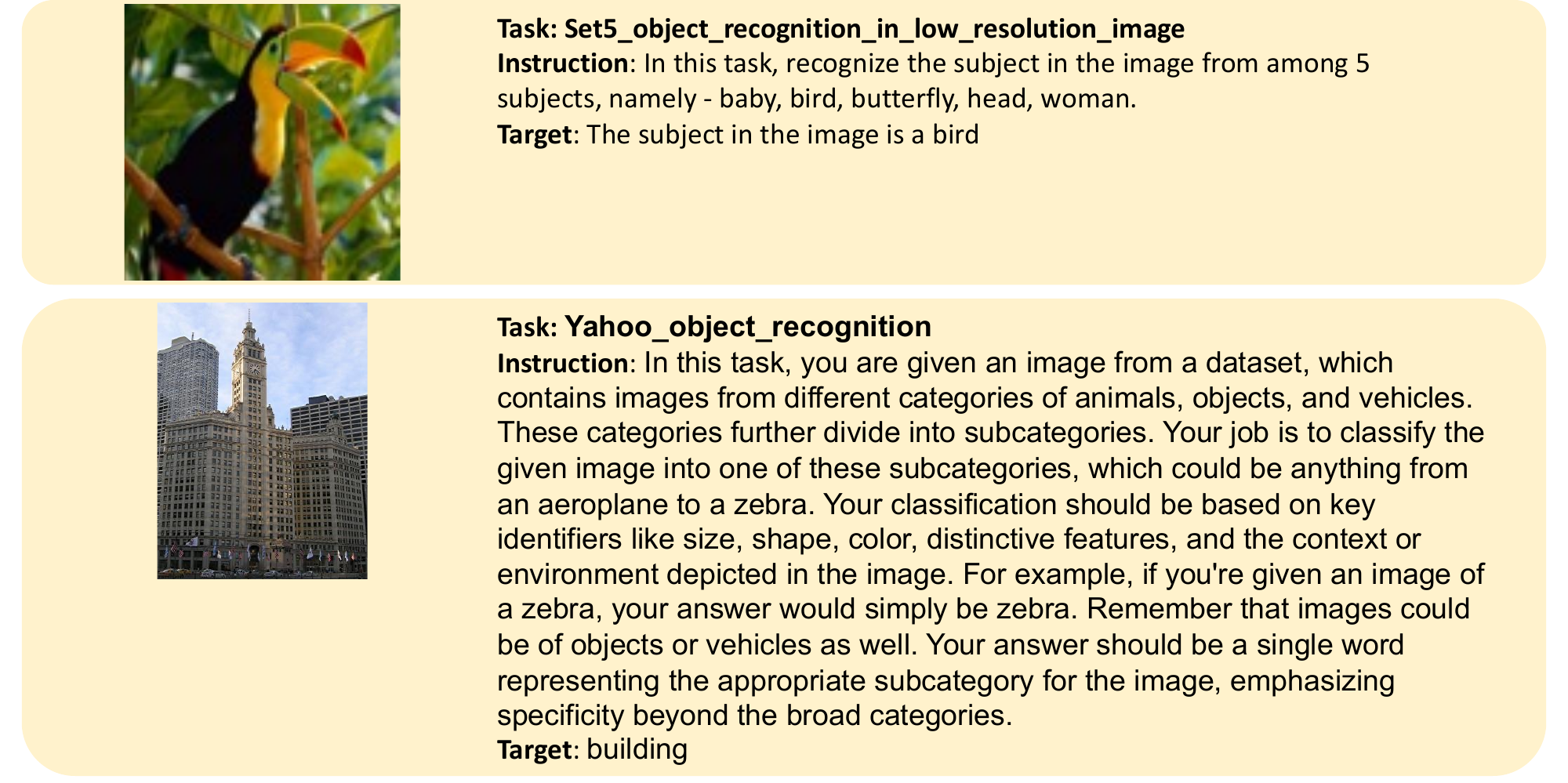}
   \caption{}
   \label{fig:}
\end{figure*}

\begin{figure*}[h!]
  \centering
   \includegraphics[width=\linewidth]{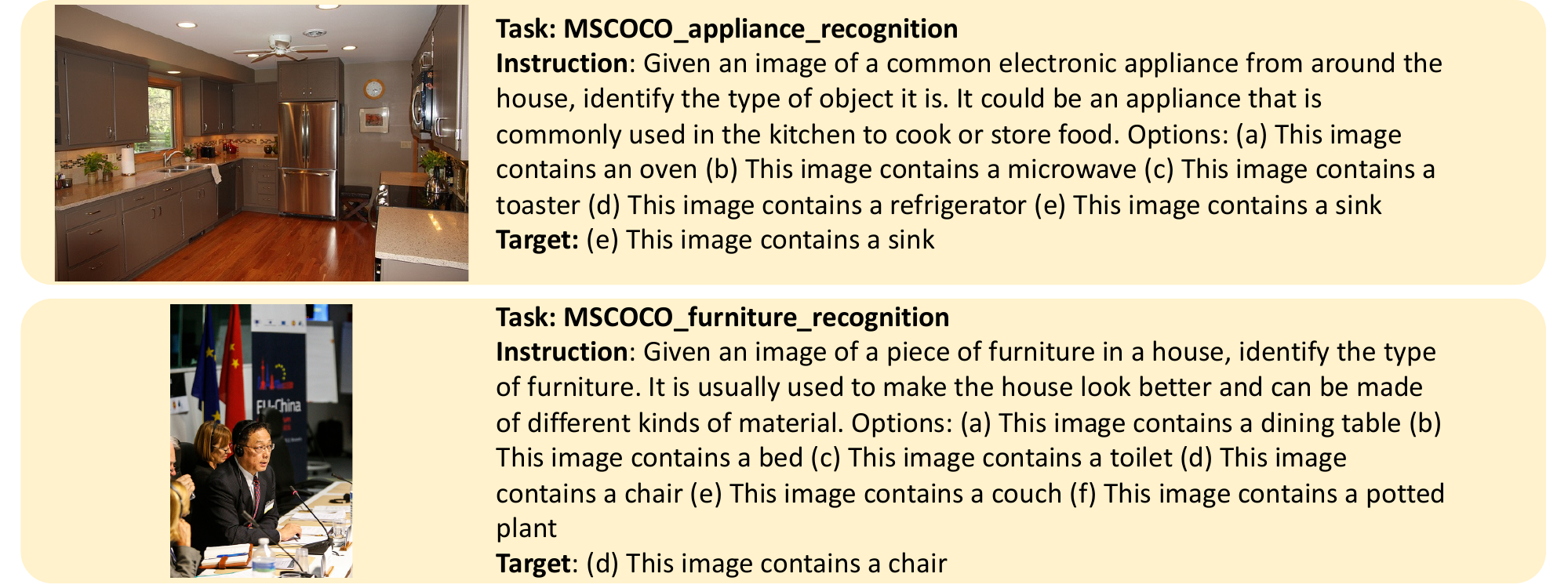}
   \caption{}
   \label{fig:}
\end{figure*}

\begin{figure*}[h!]
  \centering
   \includegraphics[width=\linewidth]{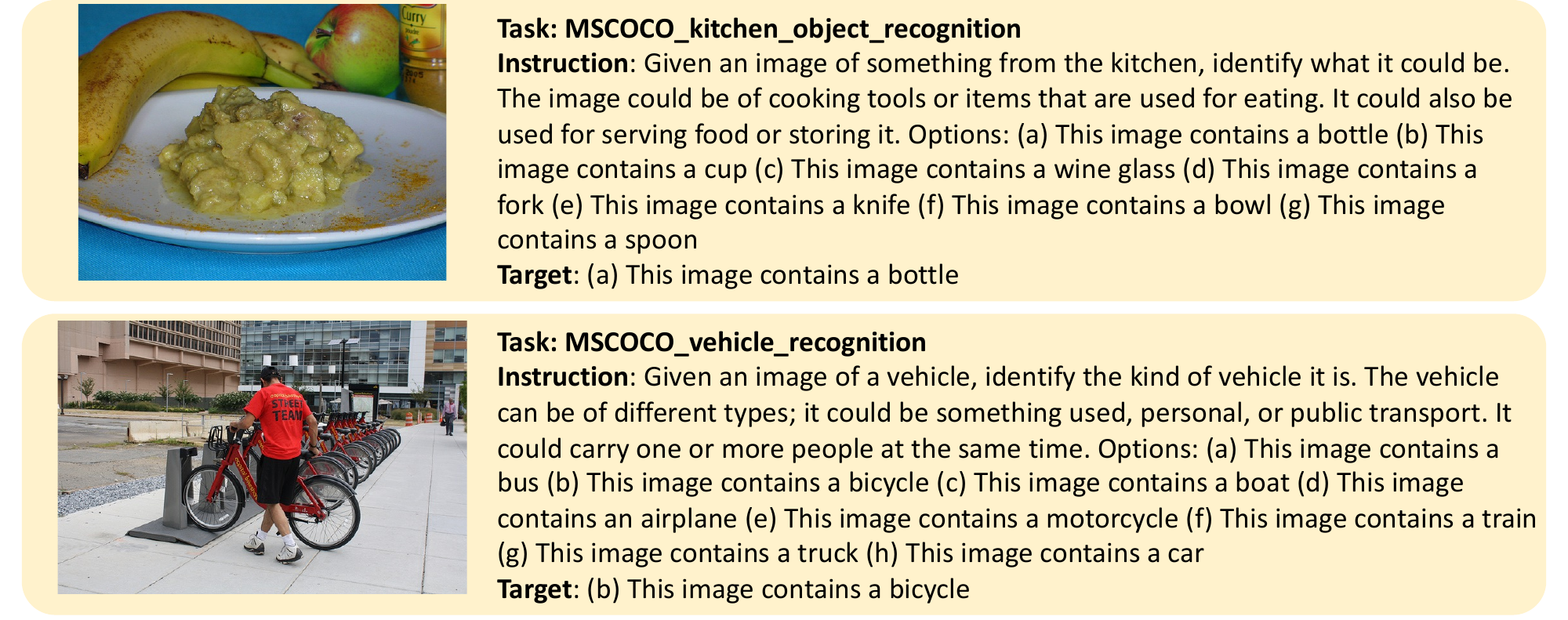}
   \caption{}
   \label{fig:}
\end{figure*}

\begin{figure*}[h!]
  \centering
   \includegraphics[width=\linewidth]{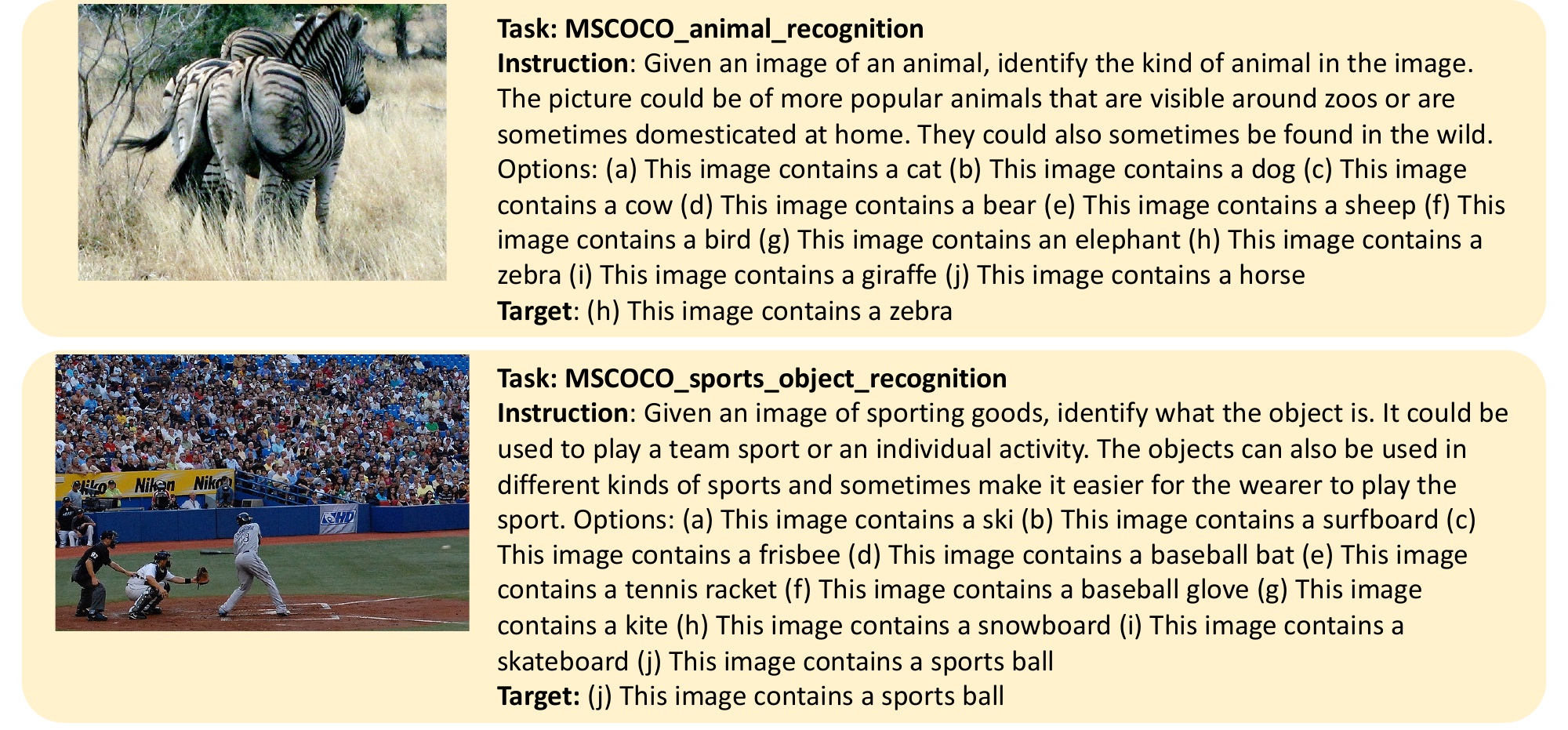}
   \caption{}
   \label{fig:}
\end{figure*}

\begin{figure*}[h!]
  \centering
   \includegraphics[width=\linewidth]{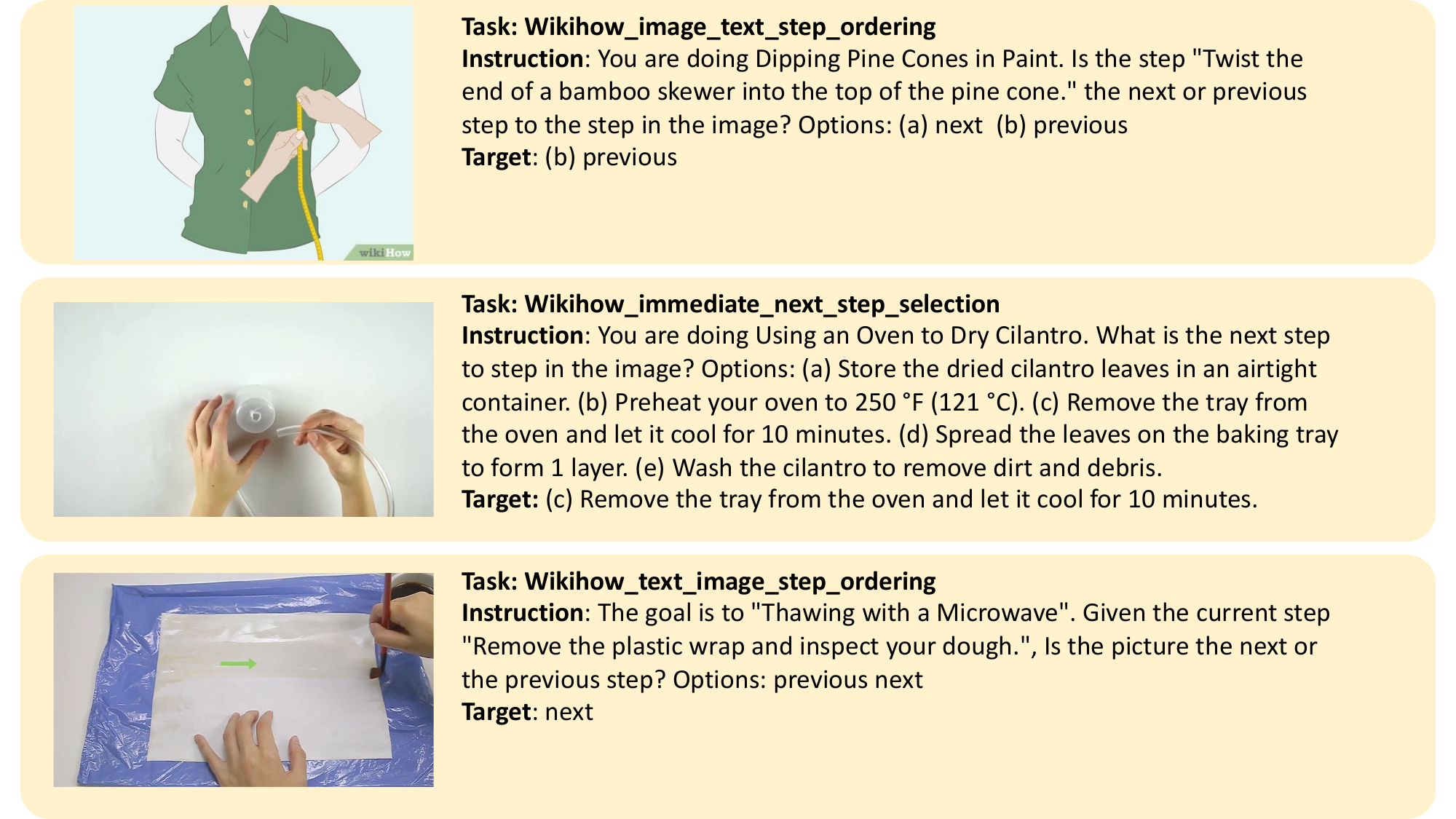}
   \caption{}
   \label{fig:}
\end{figure*}

\begin{figure*}[h!]
  \centering
   \includegraphics[width=\linewidth]{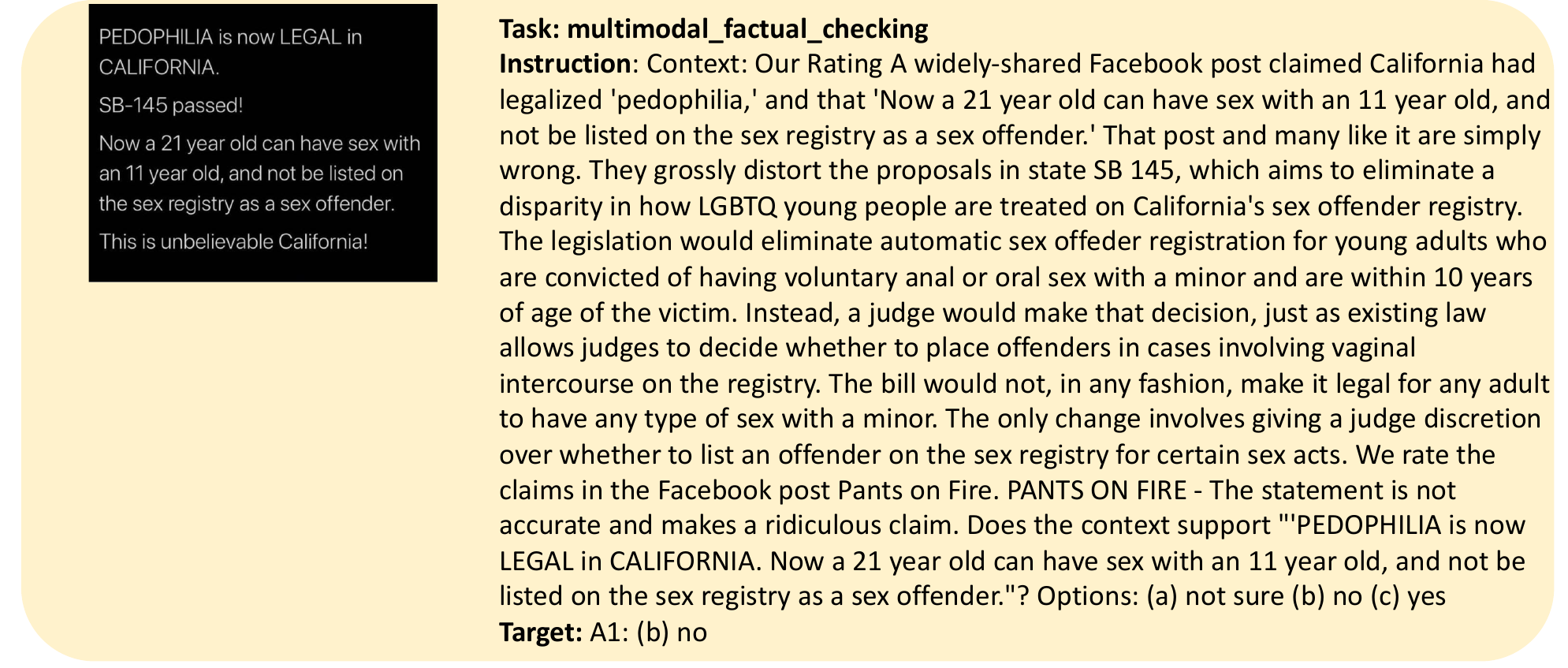}
   \caption{}
   \label{fig:}
\end{figure*}

\begin{figure*}[h!]
  \centering
   \includegraphics[width=\linewidth]{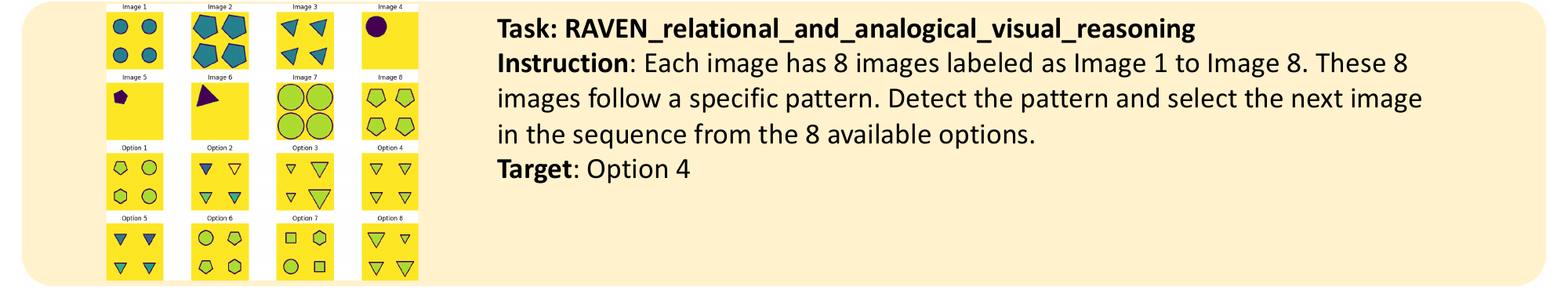}
   \caption{}
   \label{fig:}
\end{figure*}

\begin{figure*}[h!]
  \centering
   \includegraphics[width=\linewidth]{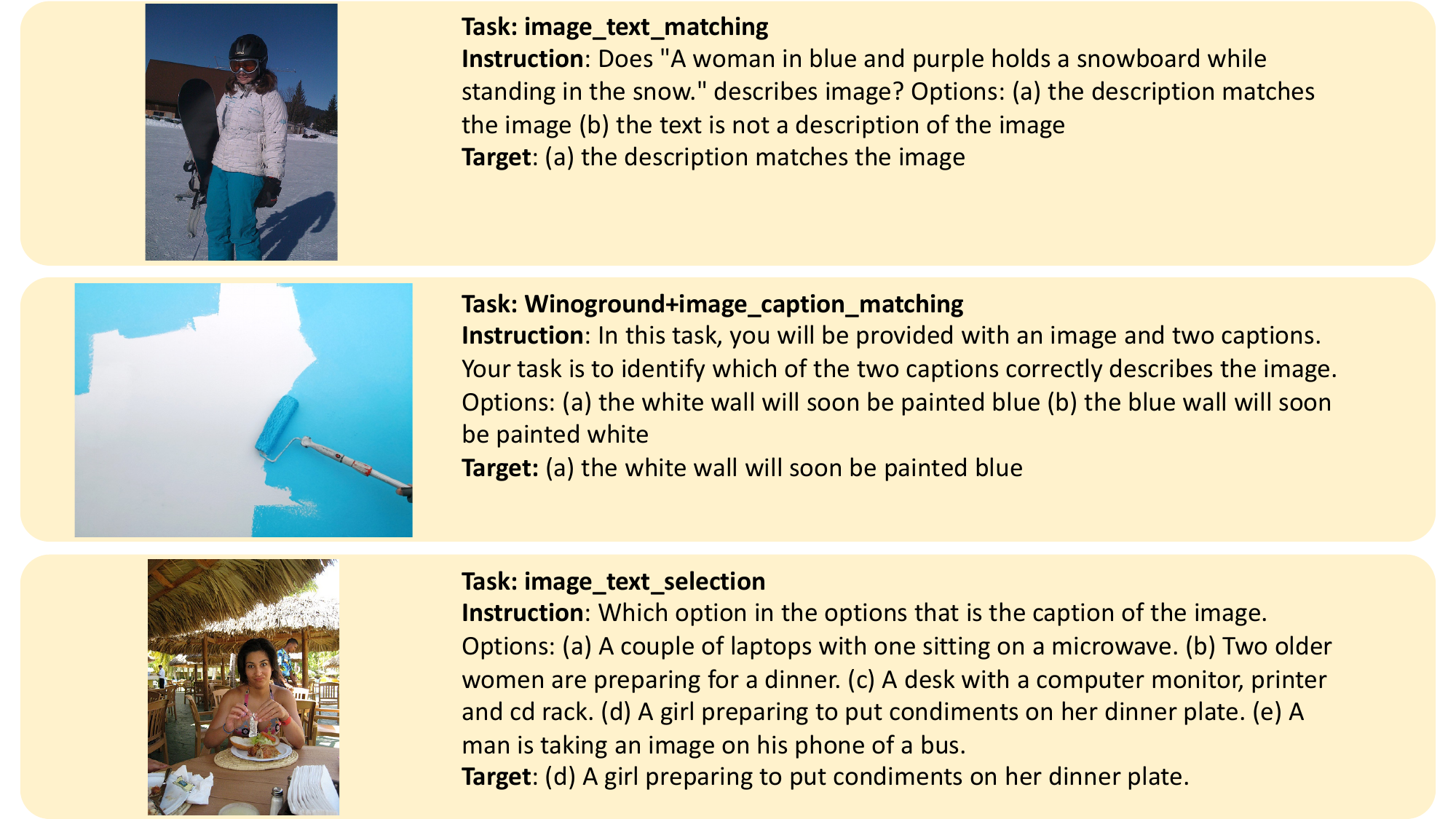}
   \caption{}
   \label{fig:}
\end{figure*}

\begin{figure*}[h!]
  \centering
   \includegraphics[width=\linewidth]{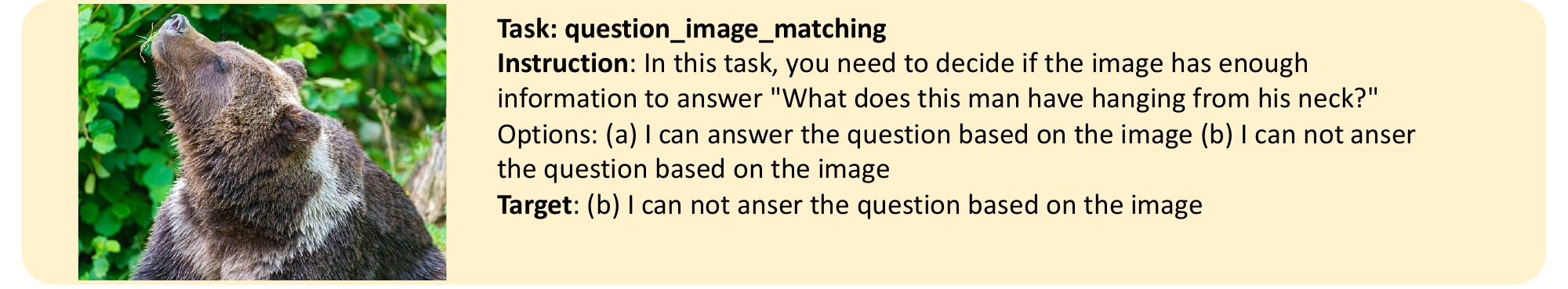}
   \caption{}
   \label{fig:}
\end{figure*}

\begin{figure*}[h!]
  \centering
   \includegraphics[width=\linewidth]{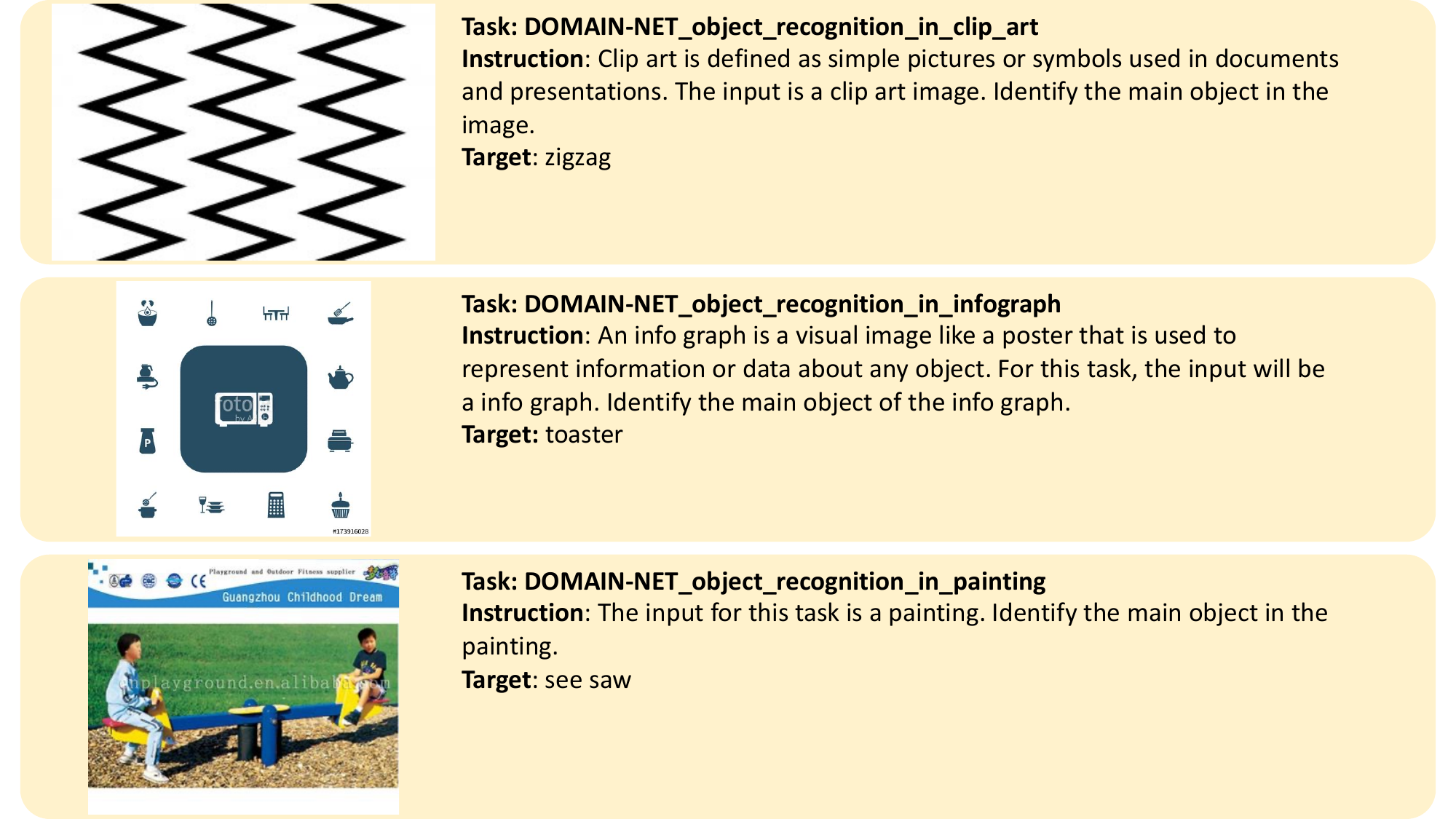}
   \caption{}
   \label{fig:}
\end{figure*}

\begin{figure*}[h!]
  \centering
   \includegraphics[width=\linewidth]{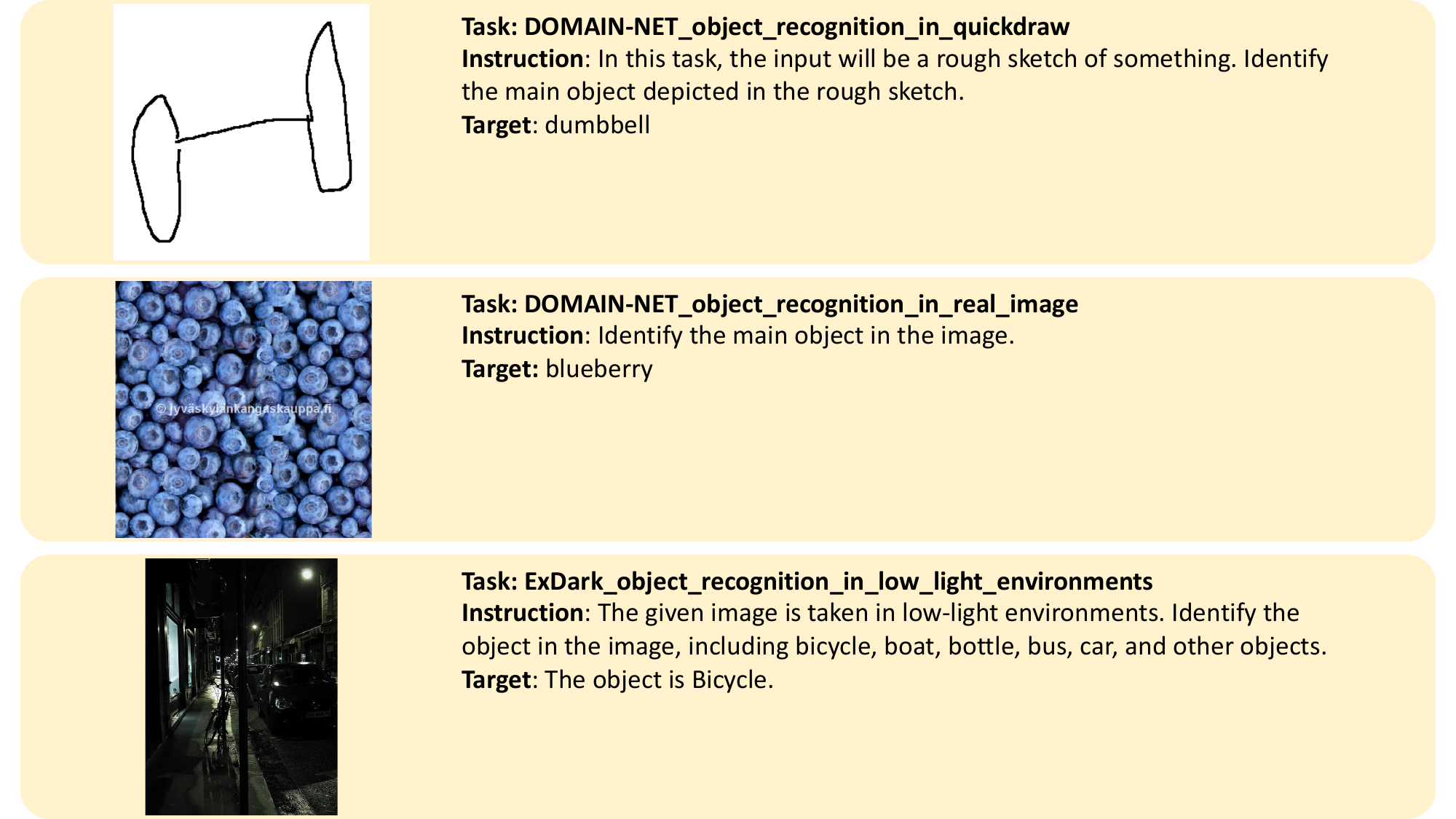}
   \caption{}
   \label{fig:}
\end{figure*}

\begin{figure*}[h!]
  \centering
   \includegraphics[width=\linewidth]{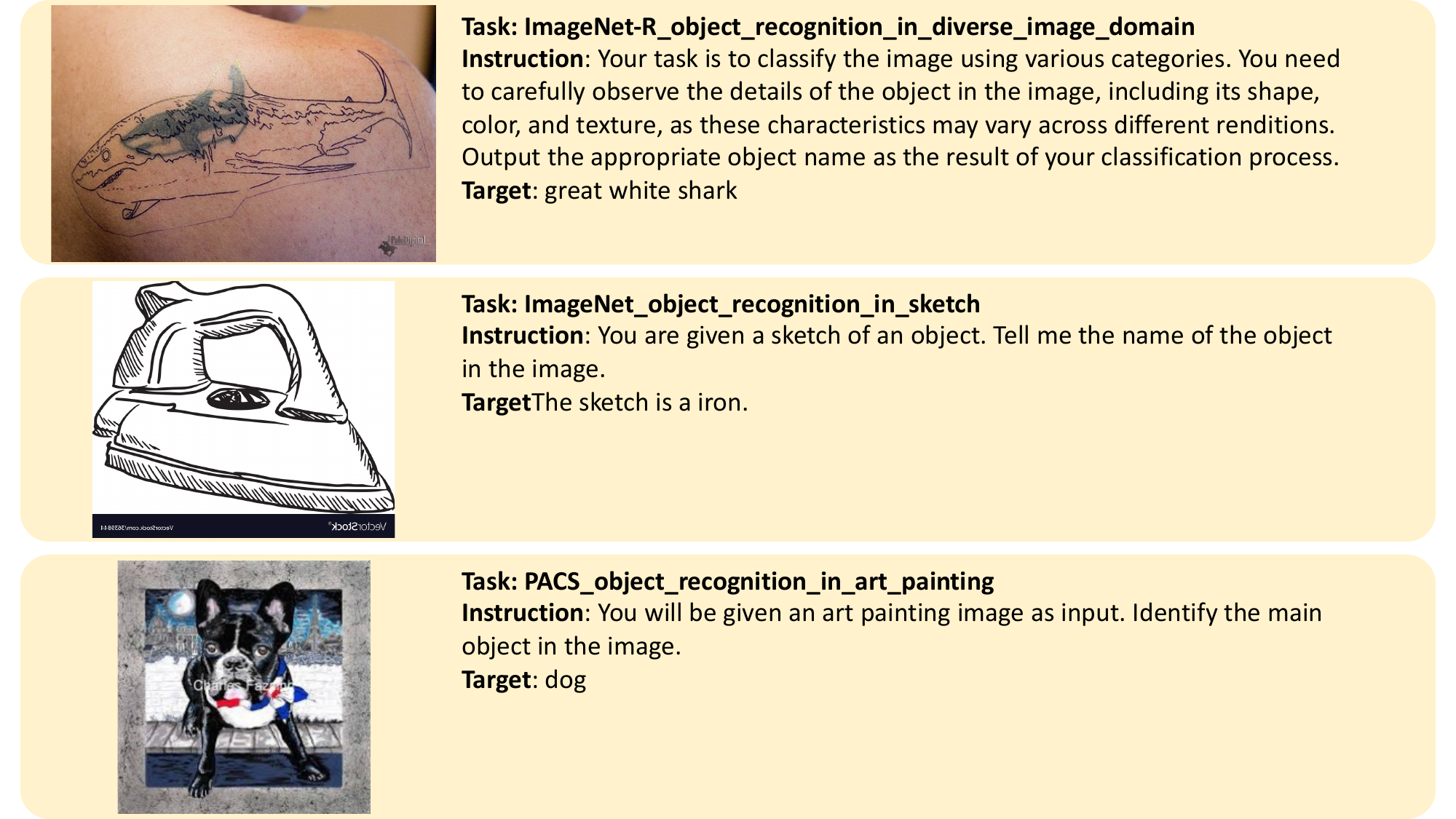}
   \caption{}
   \label{fig:}
\end{figure*}

\begin{figure*}[h!]
  \centering
   \includegraphics[width=\linewidth]{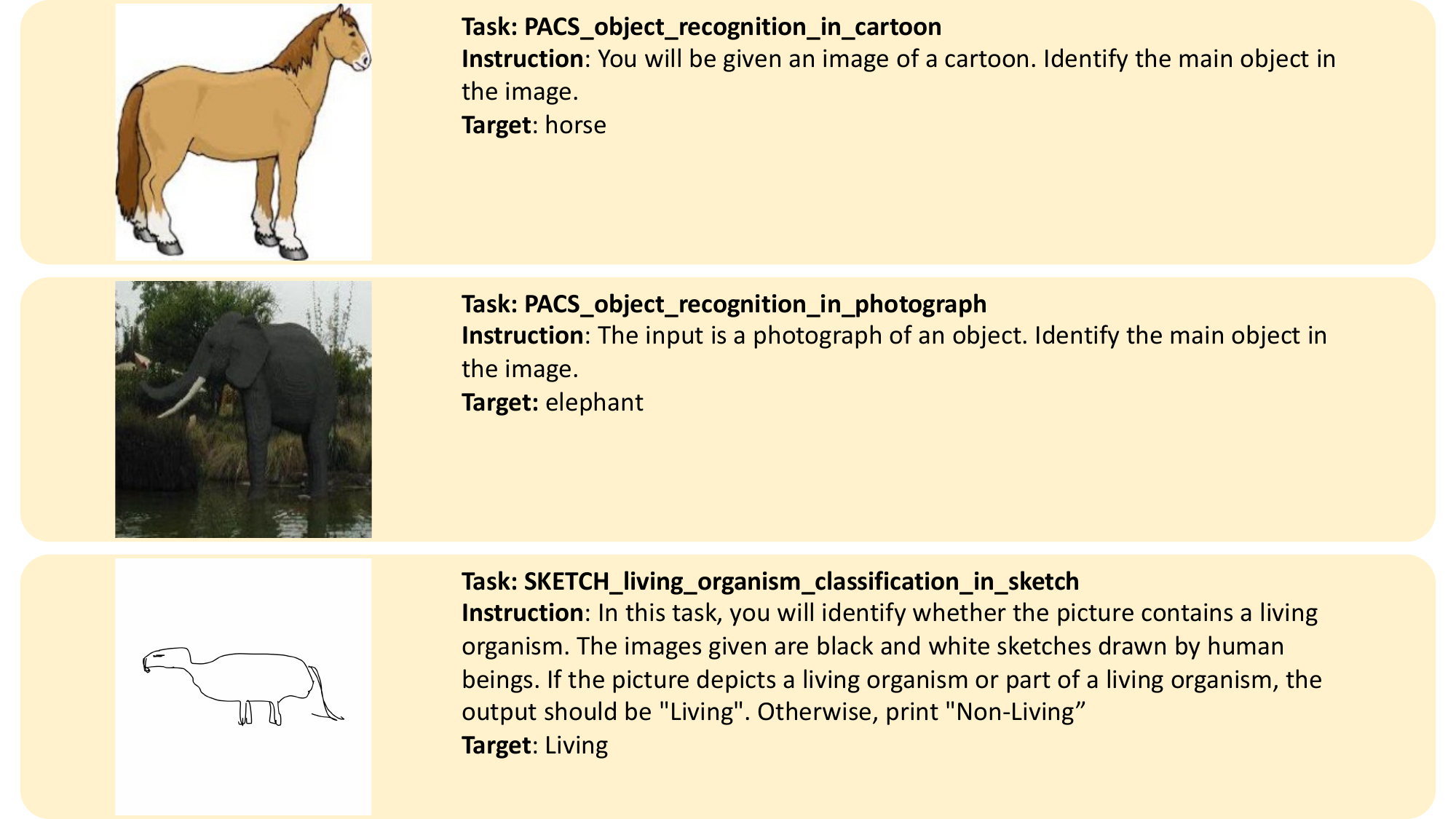}
   \caption{}
   \label{fig:}
\end{figure*}

\begin{figure*}[h!]
  \centering
   \includegraphics[width=\linewidth]{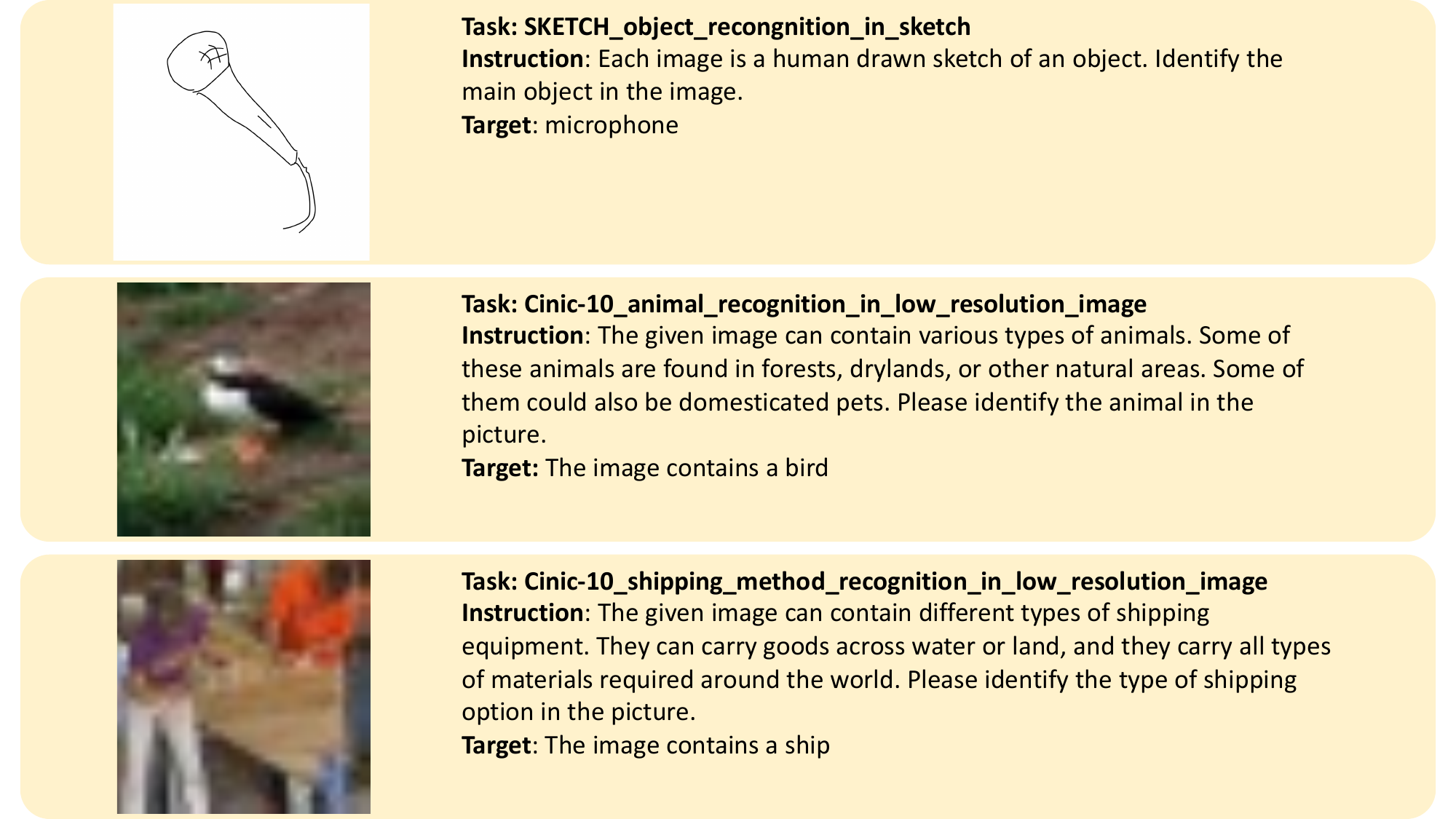}
   \caption{}
   \label{fig:}
\end{figure*}

\begin{figure*}[h!]
  \centering
   \includegraphics[width=\linewidth]{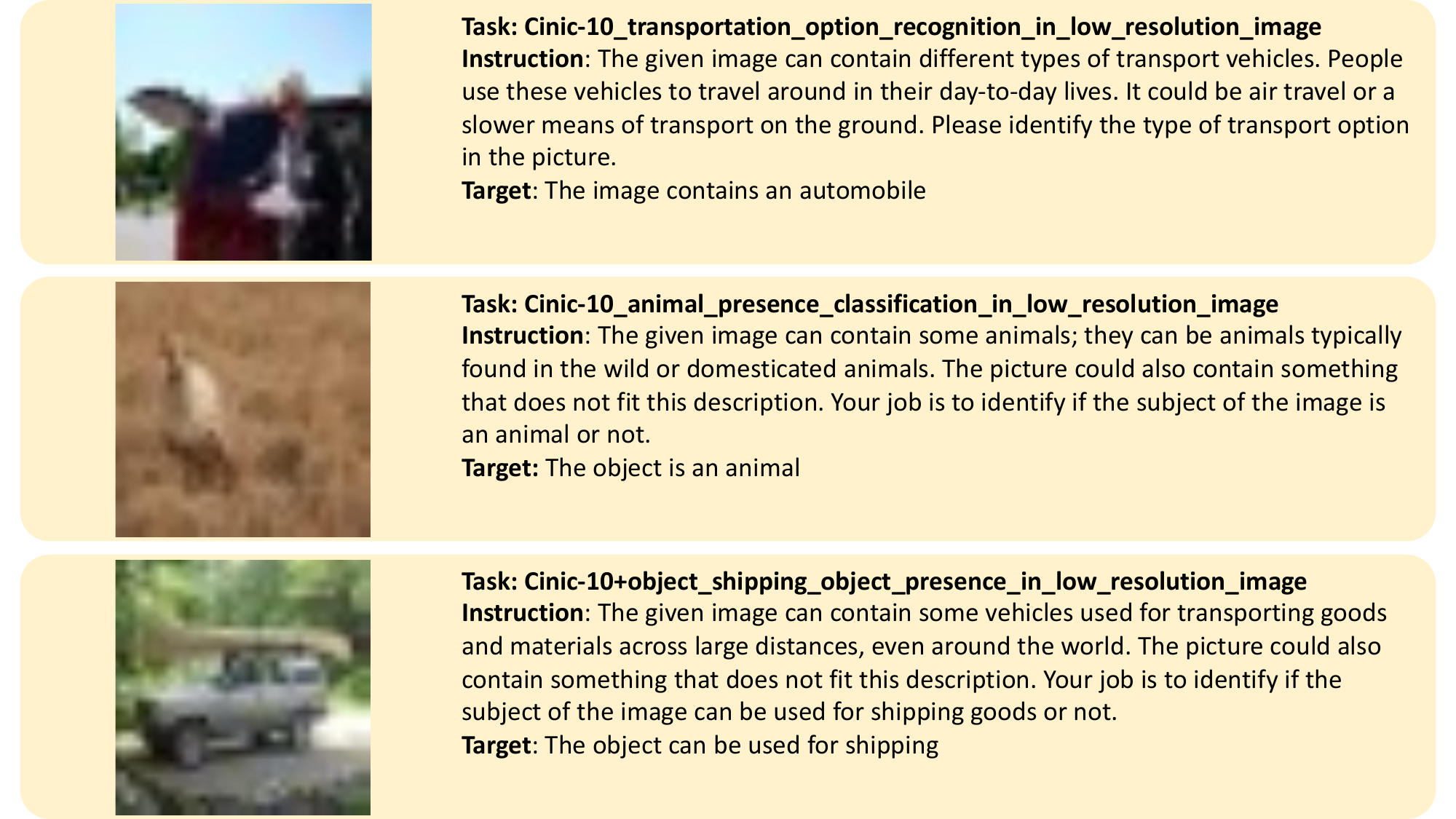}
   \caption{}
   \label{fig:}
\end{figure*}

\begin{figure*}[h!]
  \centering
   \includegraphics[width=\linewidth]{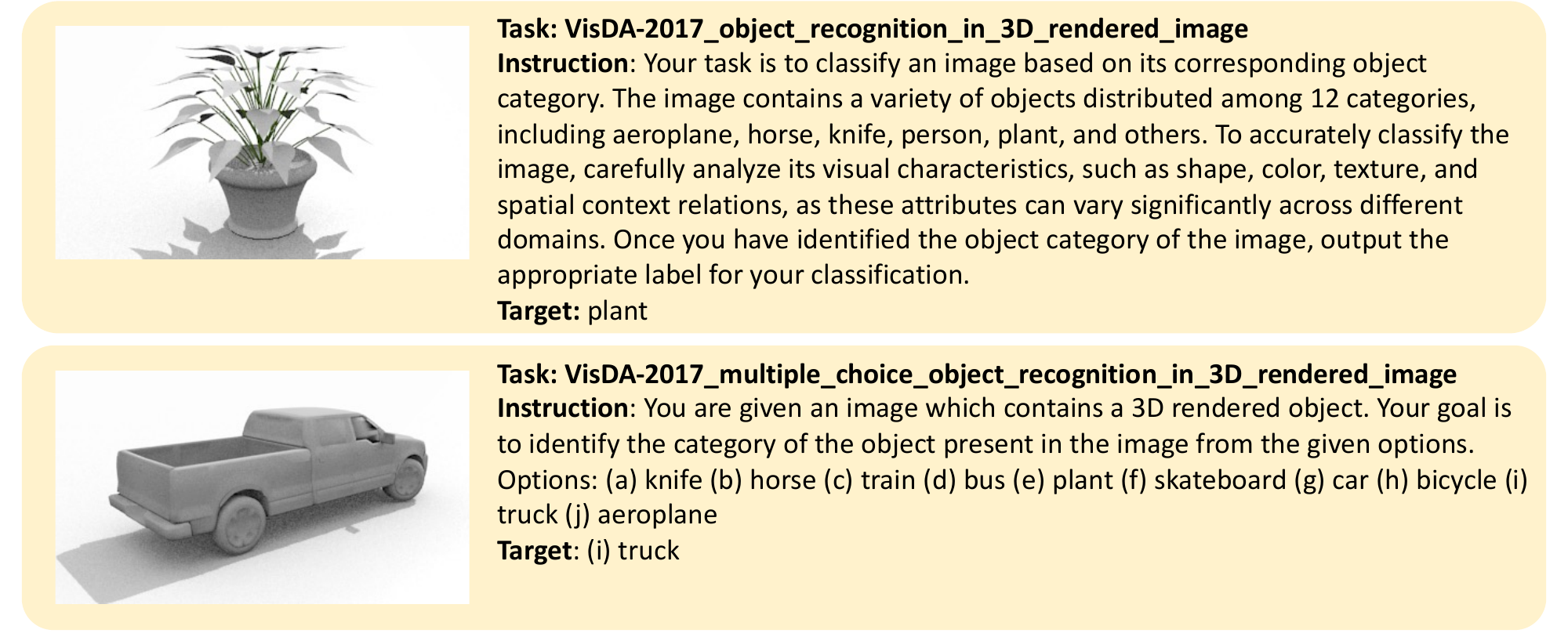}
   \caption{}
   \label{fig:}
\end{figure*}

\begin{figure*}[h!]
  \centering
   \includegraphics[width=\linewidth]{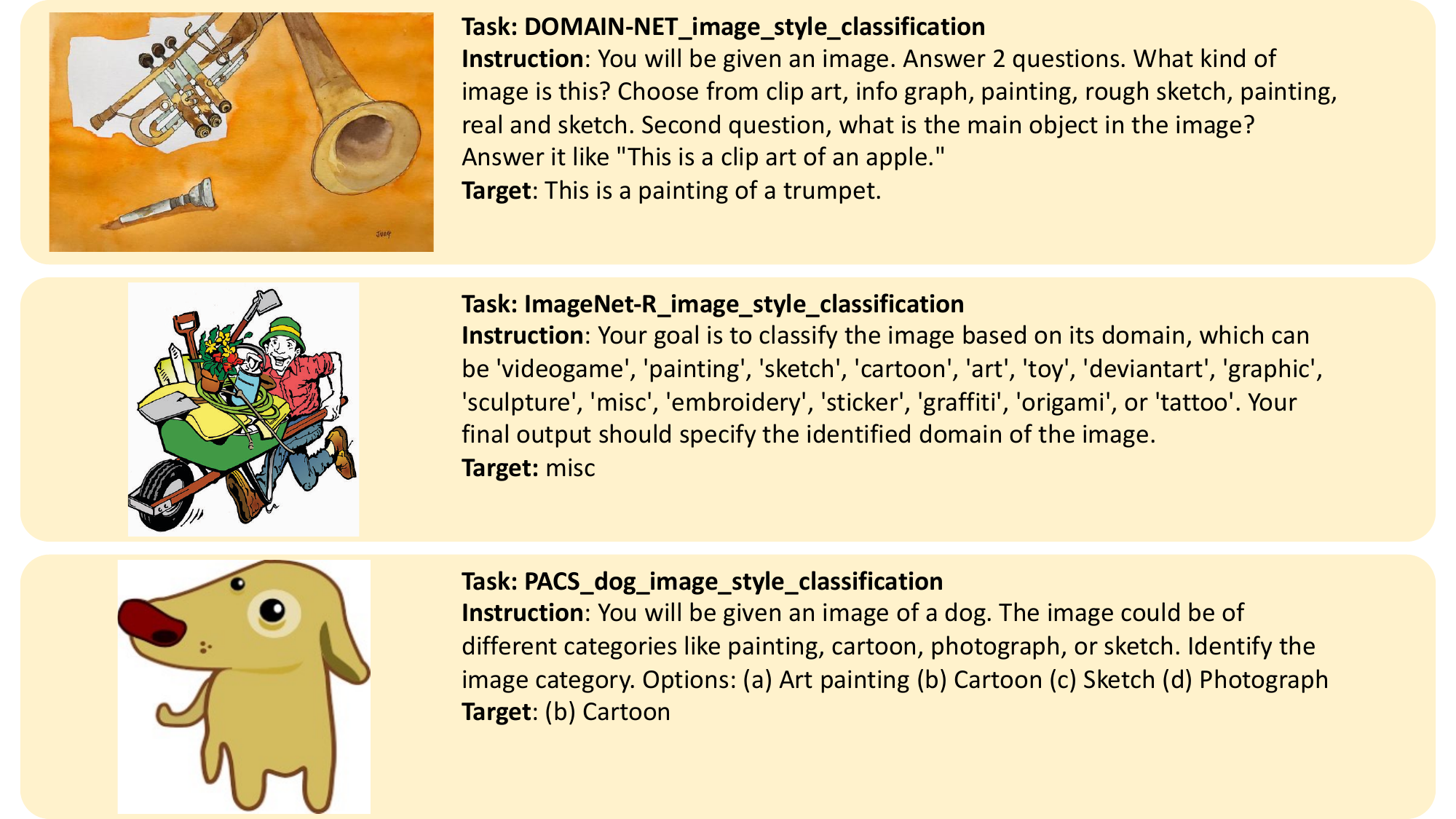}
   \caption{}
   \label{fig:}
\end{figure*}

\begin{figure*}[h!]
  \centering
   \includegraphics[width=\linewidth]{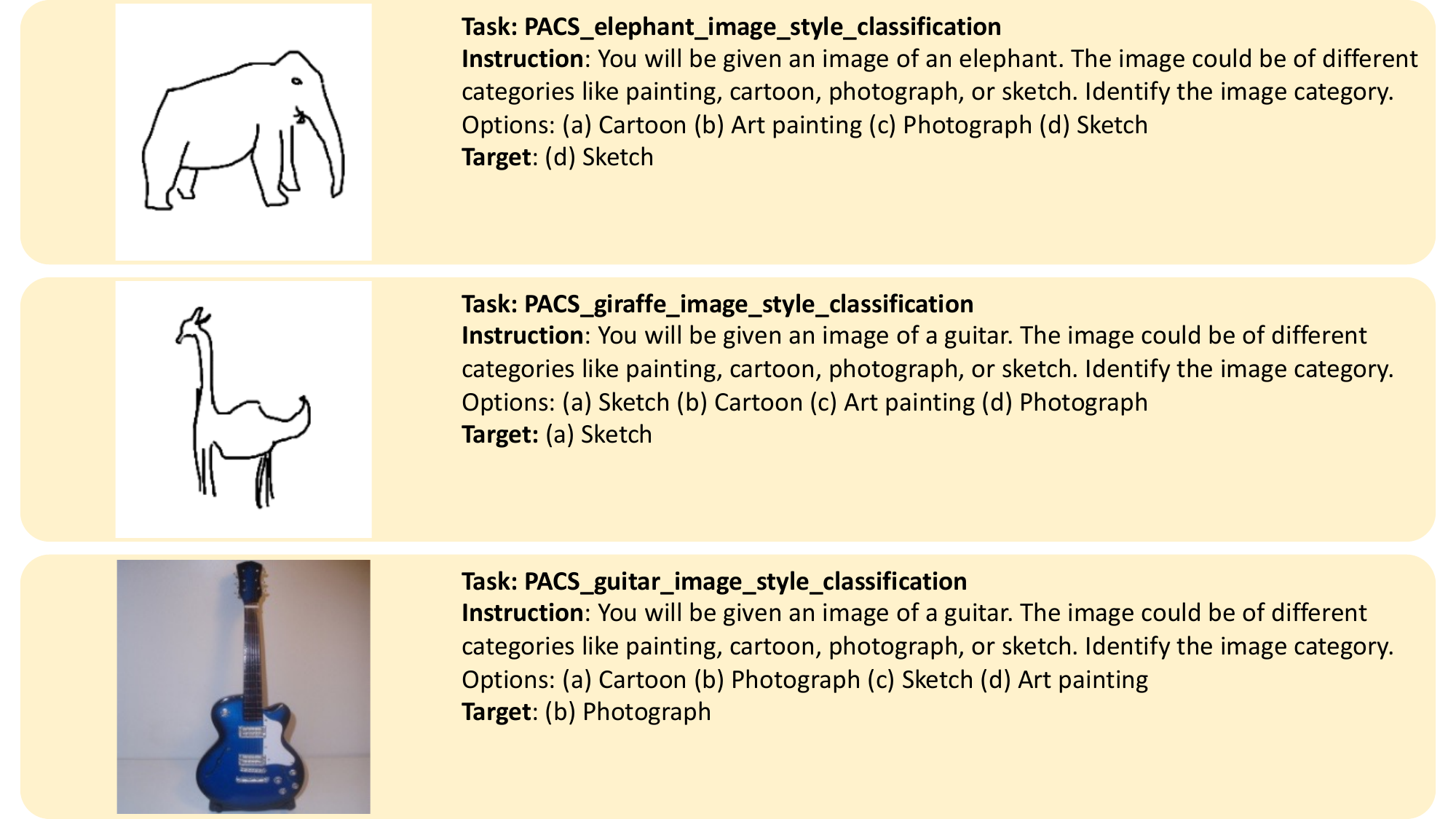}
   \caption{}
   \label{fig:}
\end{figure*}

\begin{figure*}[h!]
  \centering
   \includegraphics[width=\linewidth]{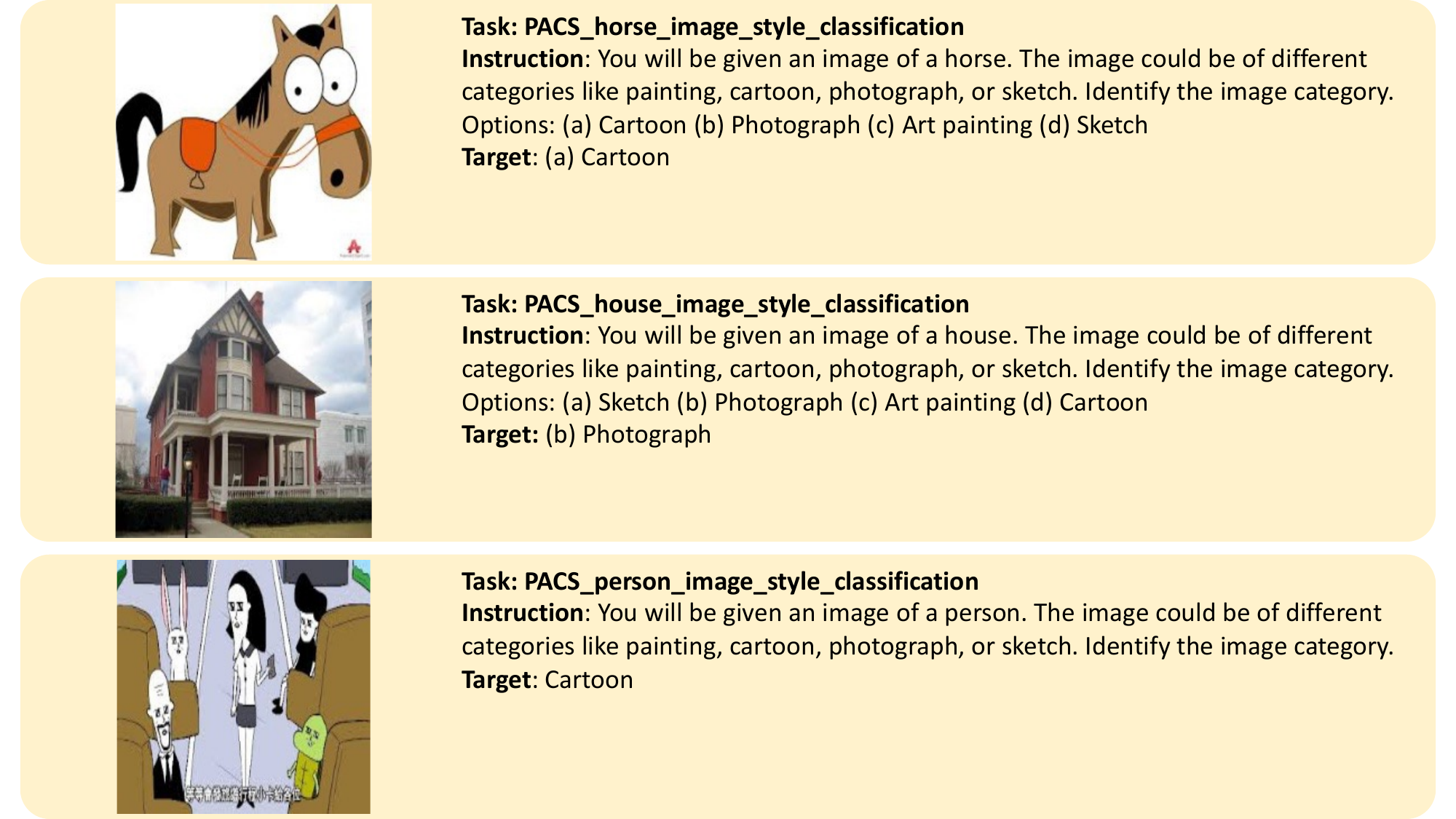}
   \caption{}
   \label{fig:}
\end{figure*}

\begin{figure*}[h!]
  \centering
   \includegraphics[width=\linewidth]{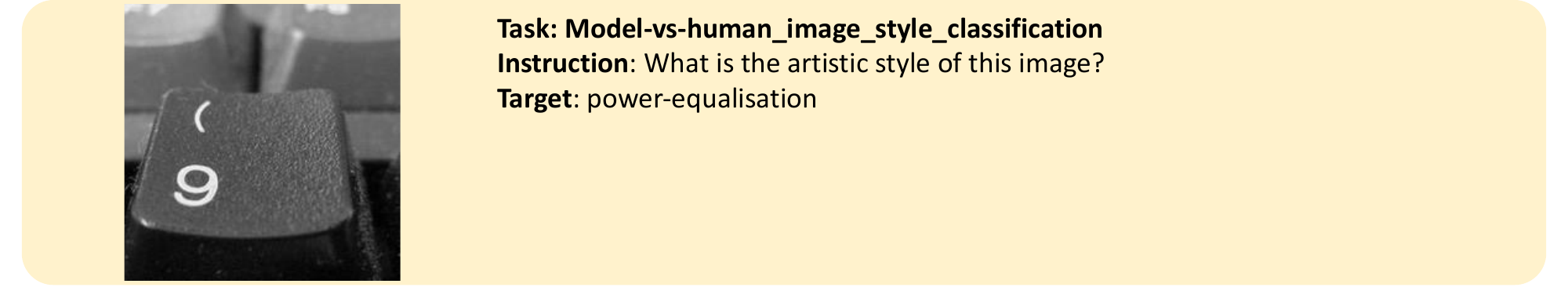}
   \caption{}
   \label{fig:}
\end{figure*}

\begin{figure*}[h!]
  \centering
   \includegraphics[width=\linewidth]{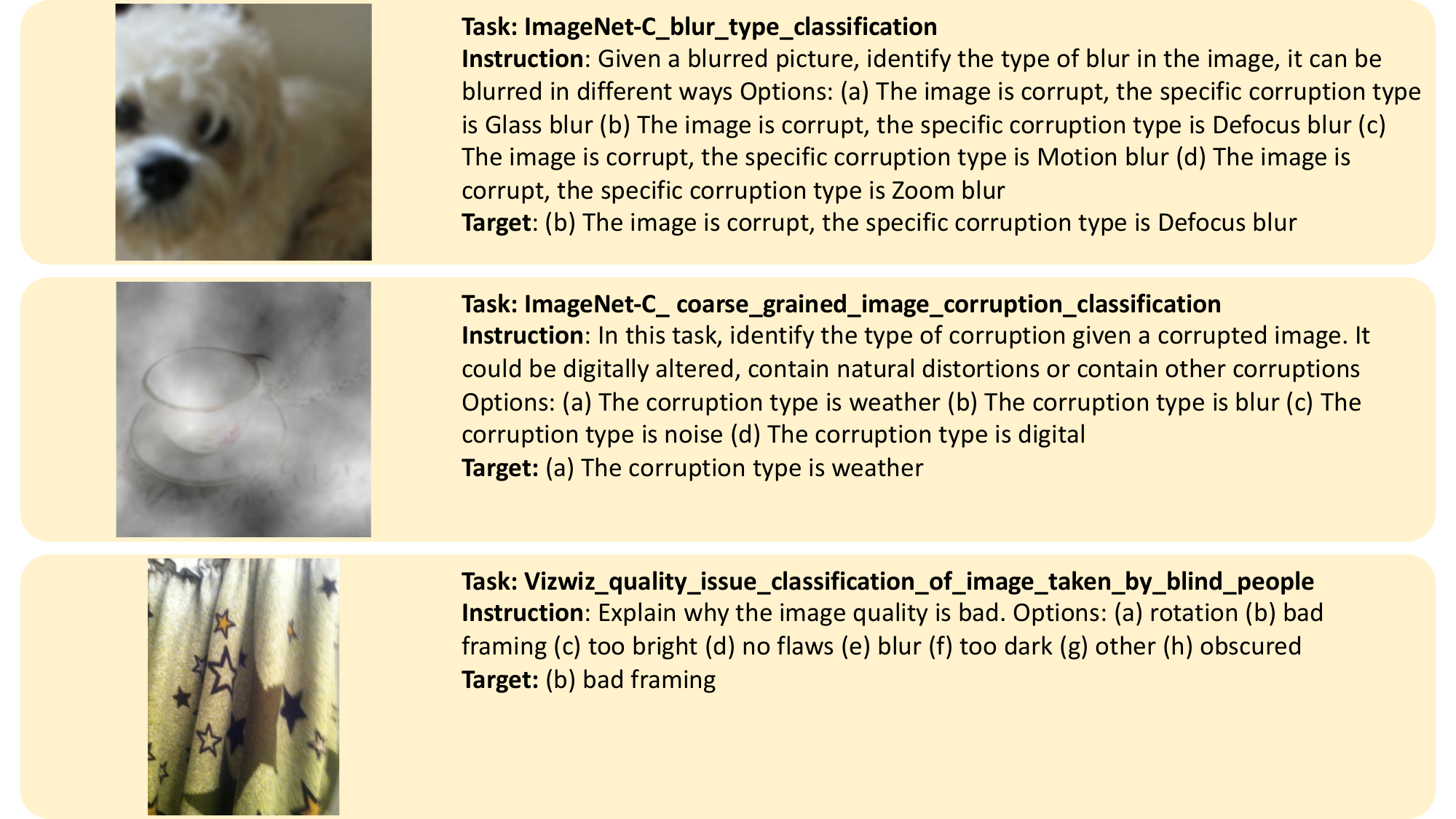}
   \caption{}
   \label{fig:}
\end{figure*}

\begin{figure*}[h!]
  \centering
   \includegraphics[width=\linewidth]{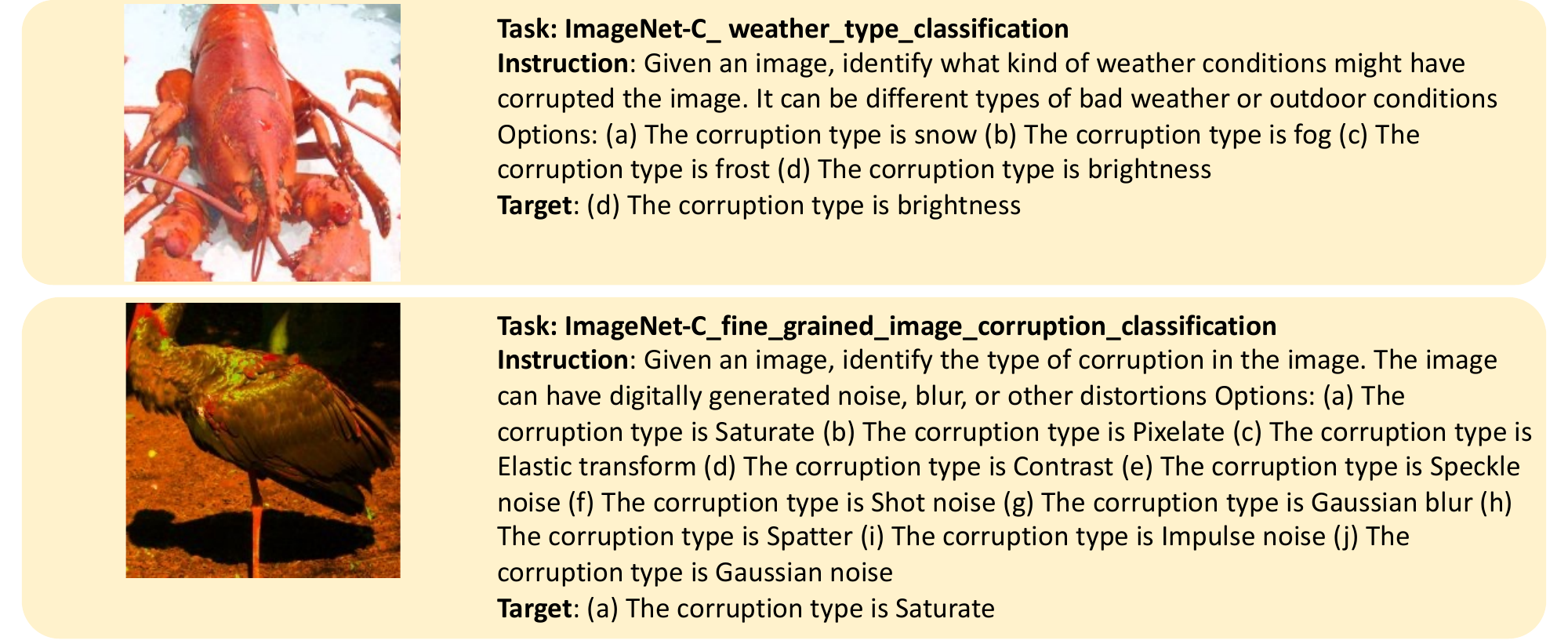}
   \caption{}
   \label{fig:}
\end{figure*}

\begin{figure*}[h!]
  \centering
   \includegraphics[width=\linewidth]{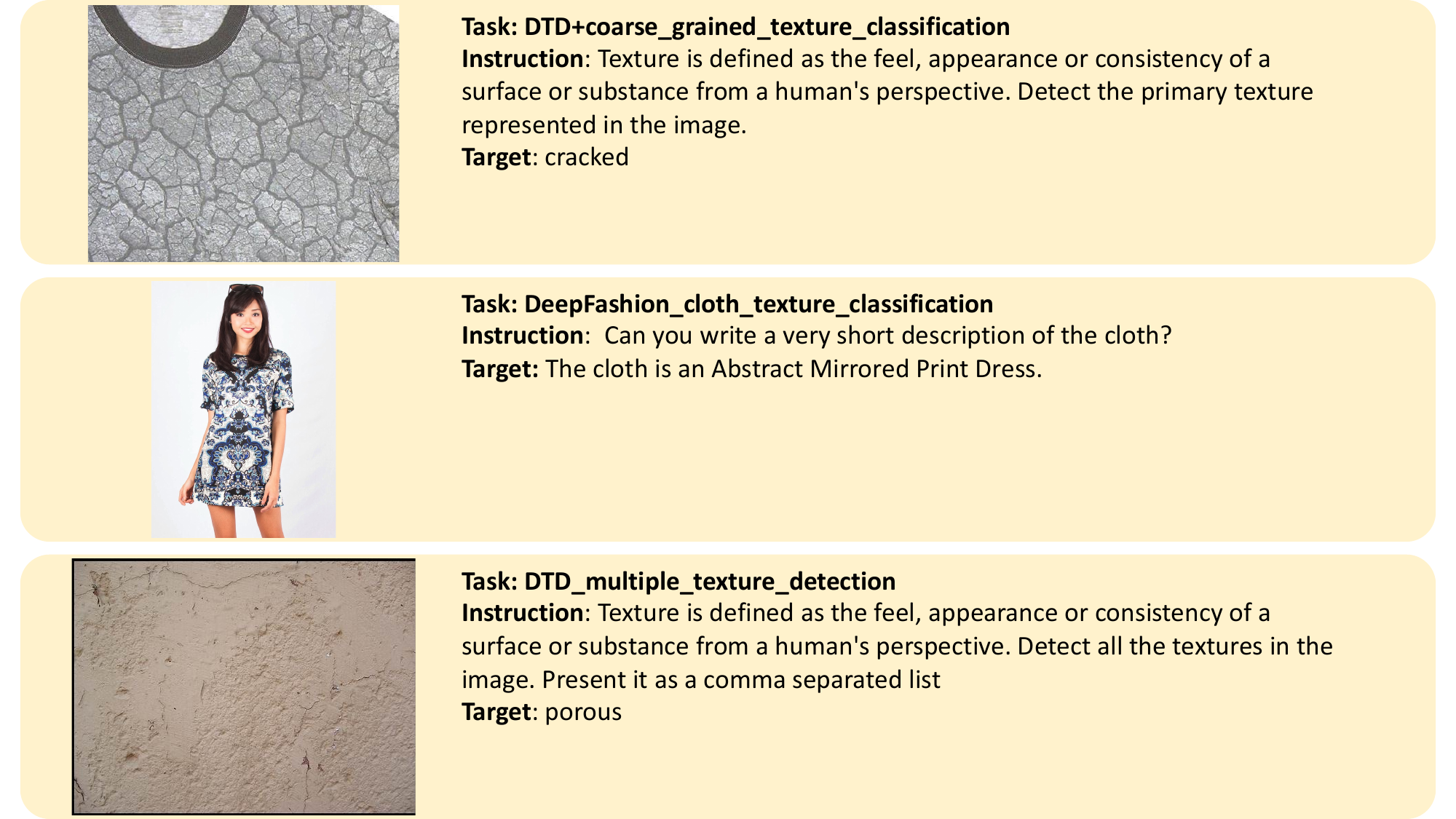}
   \caption{}
   \label{fig:}
\end{figure*}

\clearpage
\subsection{VQA Tasks}

\begin{figure*}[h!]
  \centering
   \includegraphics[width=\linewidth]{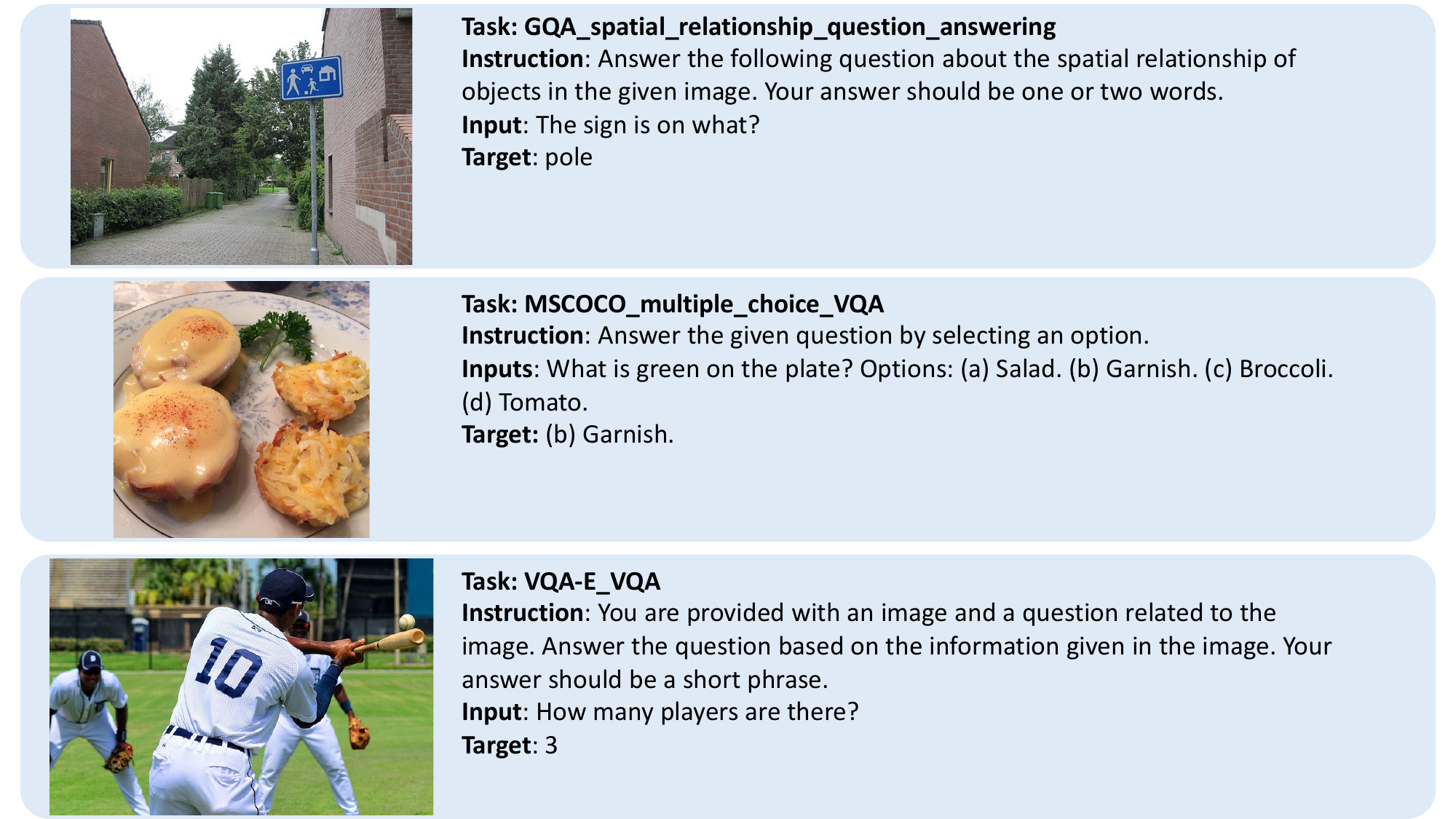}
   \caption{}
   \label{fig:}
\end{figure*}

\begin{figure*}[h!]
  \centering
   \includegraphics[width=\linewidth]{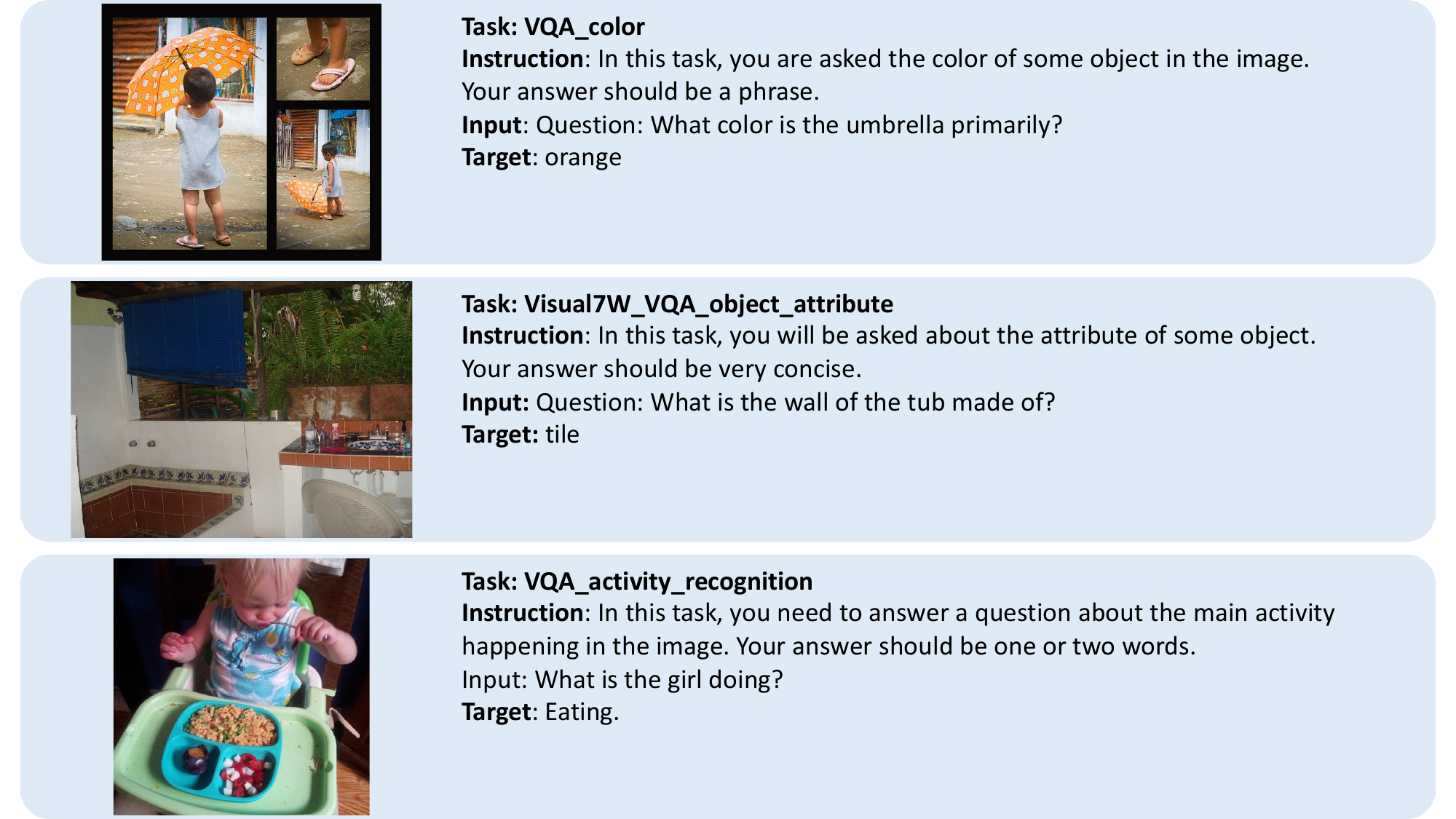}
   \caption{}
   \label{fig:}
\end{figure*}

\begin{figure*}[h!]
  \centering
   \includegraphics[width=\linewidth]{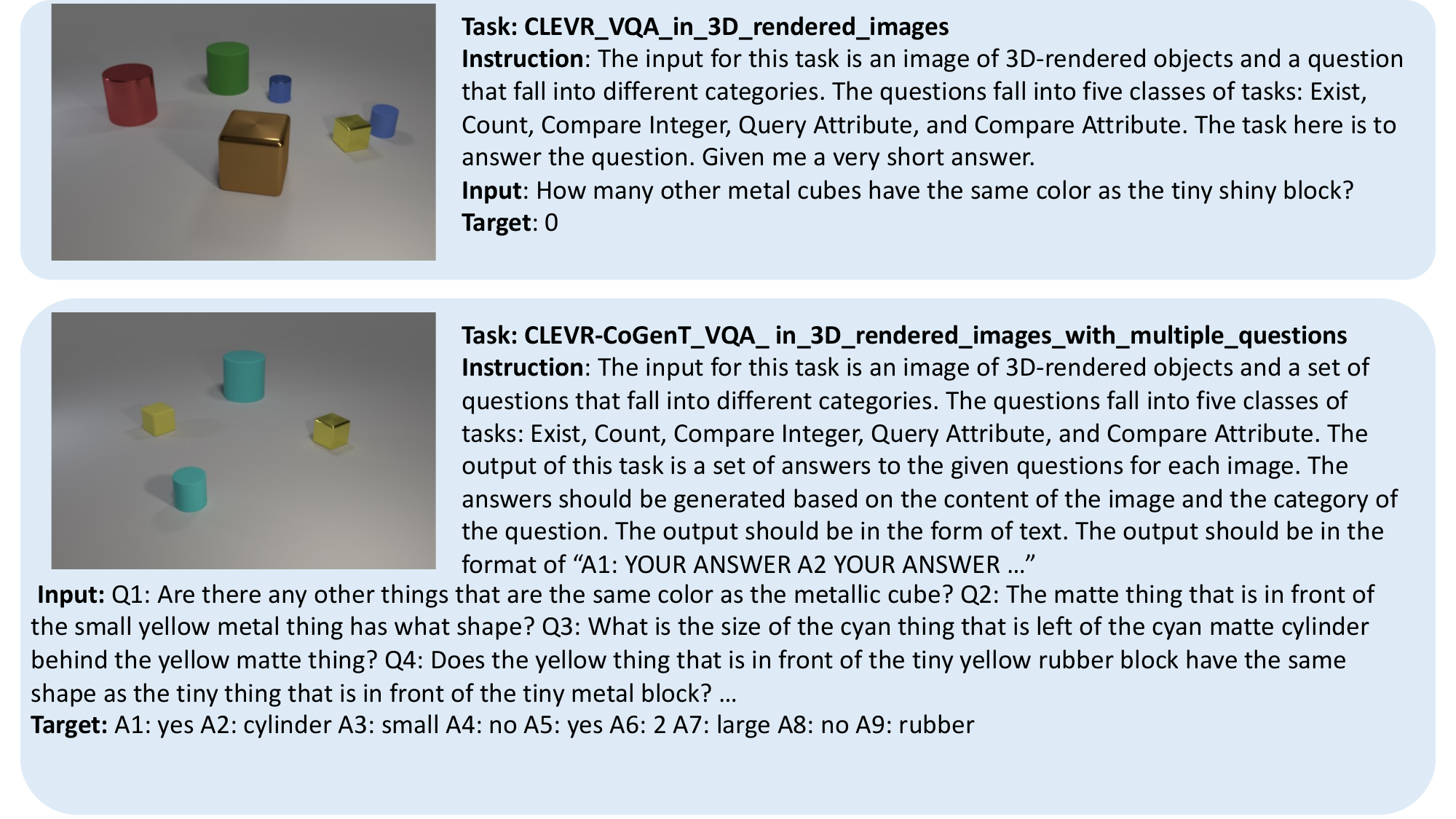}
   \caption{}
   \label{fig:}
\end{figure*}

\begin{figure*}[h!]
  \centering
   \includegraphics[width=\linewidth]{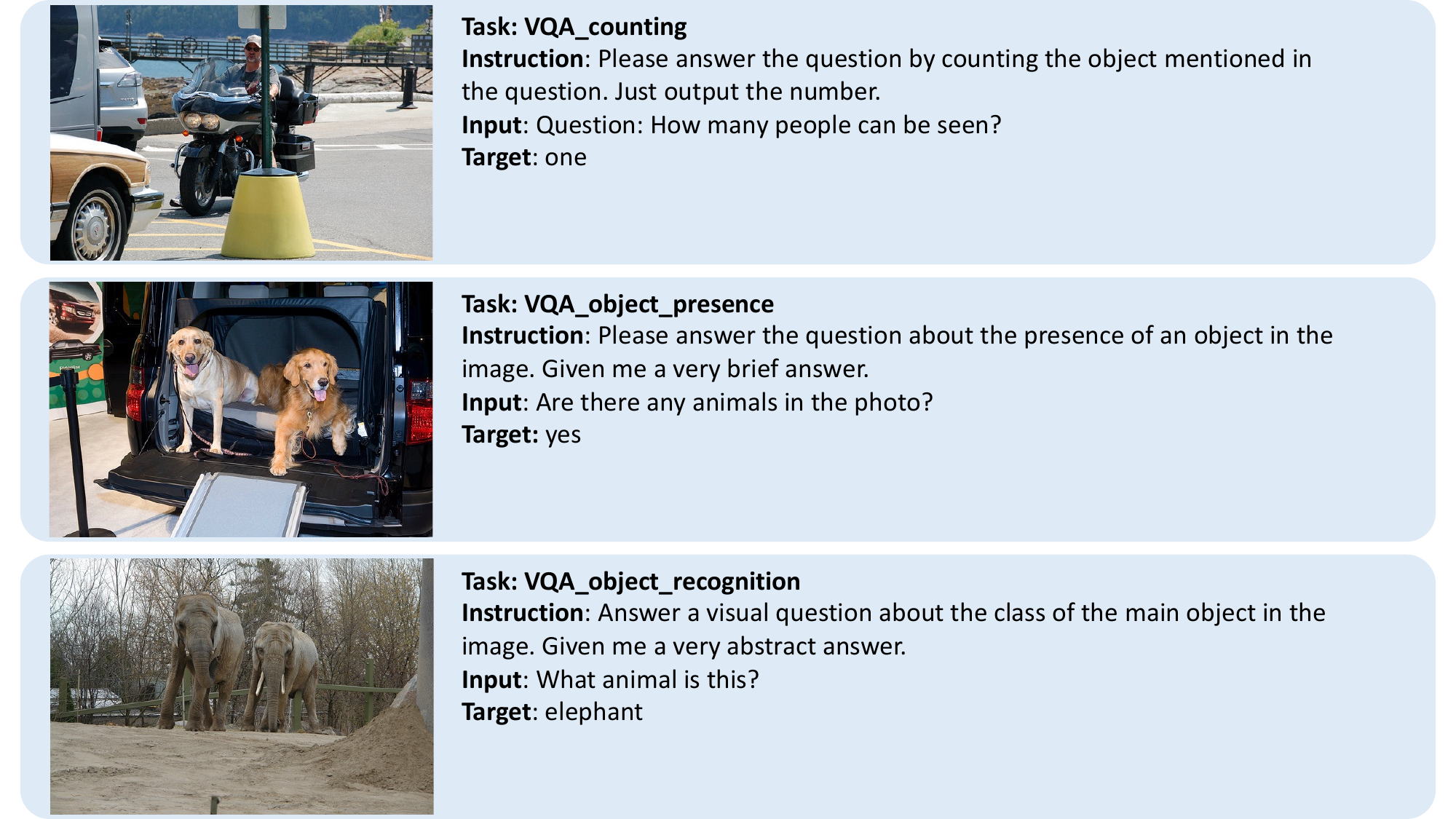}
   \caption{}
   \label{fig:}
\end{figure*}

\begin{figure*}[h!]
  \centering
   \includegraphics[width=\linewidth]{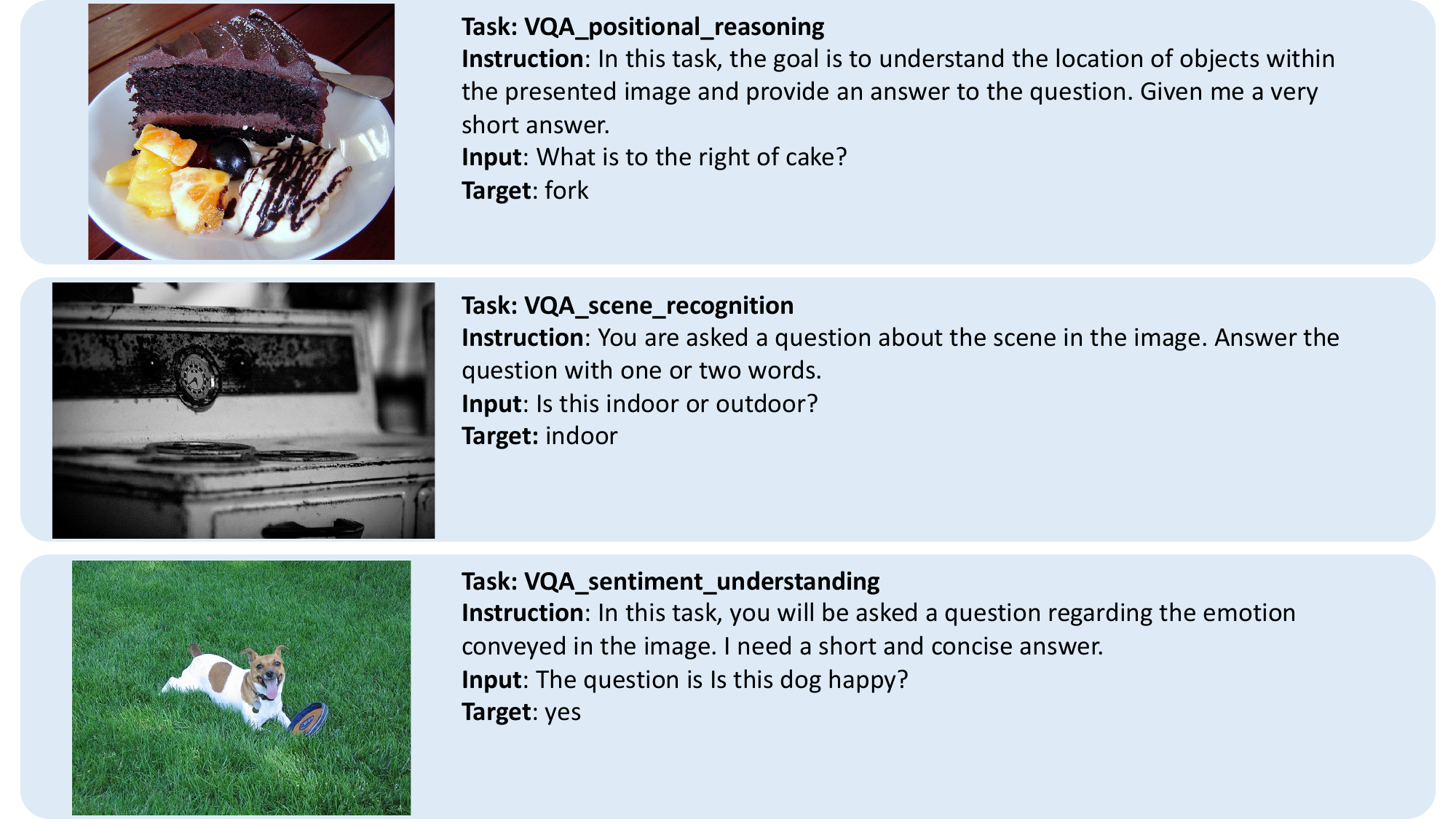}
   \caption{}
   \label{fig:}
\end{figure*}

\begin{figure*}[h!]
  \centering
   \includegraphics[width=\linewidth]{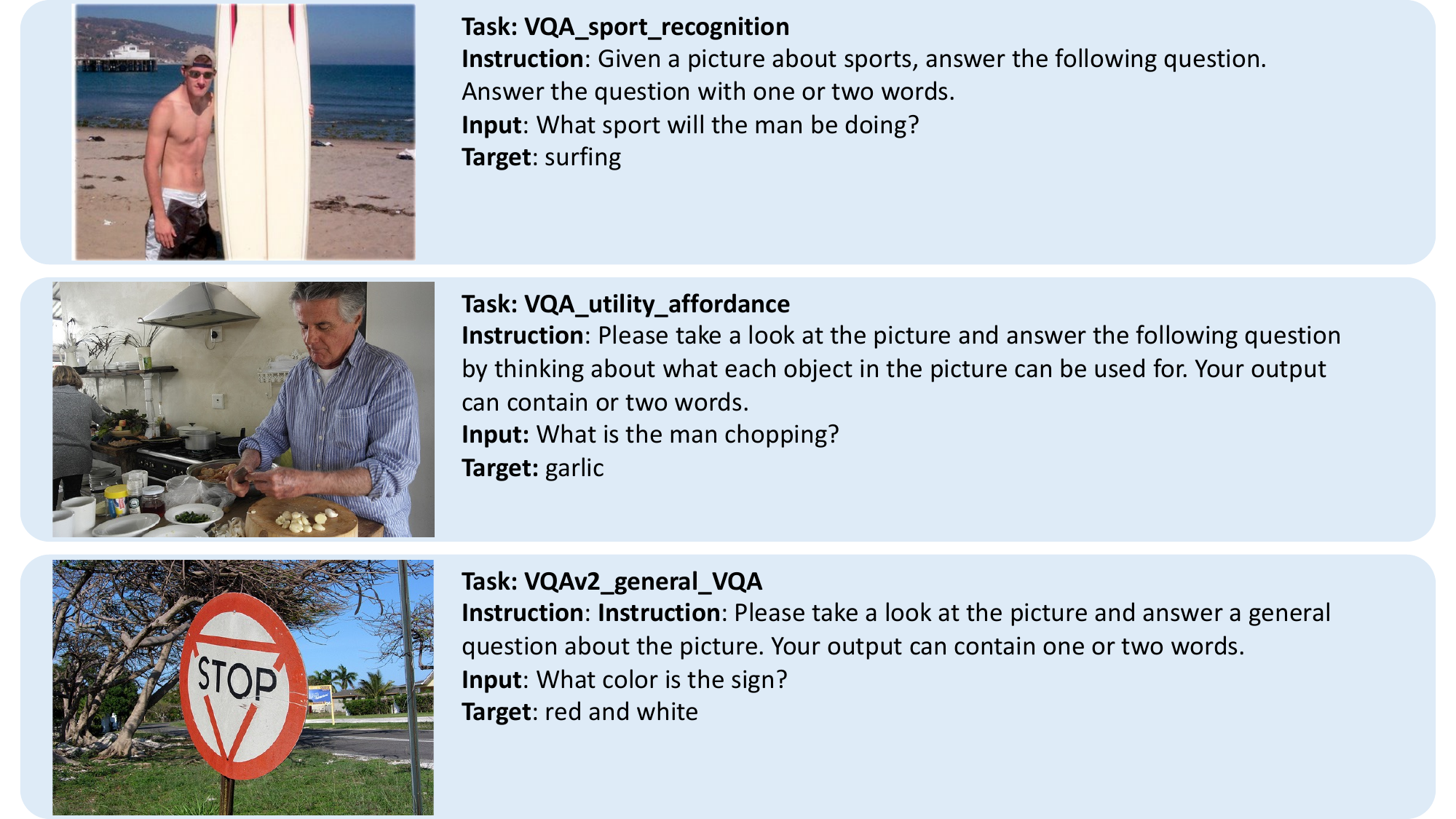}
   \caption{}
   \label{fig:}
\end{figure*}

\begin{figure*}[h!]
  \centering
   \includegraphics[width=\linewidth]{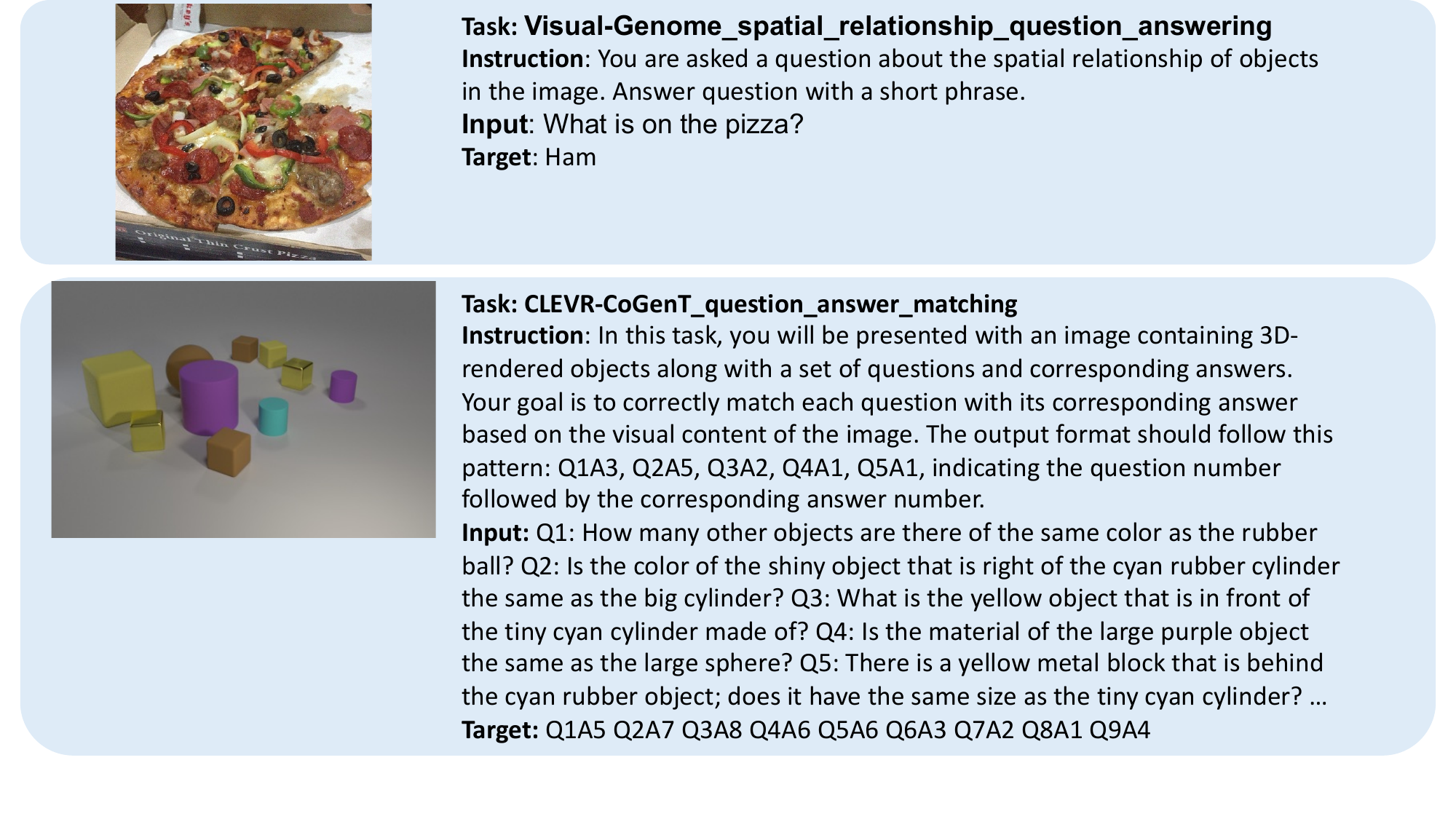}
   \caption{}
   \label{fig:}
\end{figure*}

\begin{figure*}[h!]
  \centering
   \includegraphics[width=\linewidth]{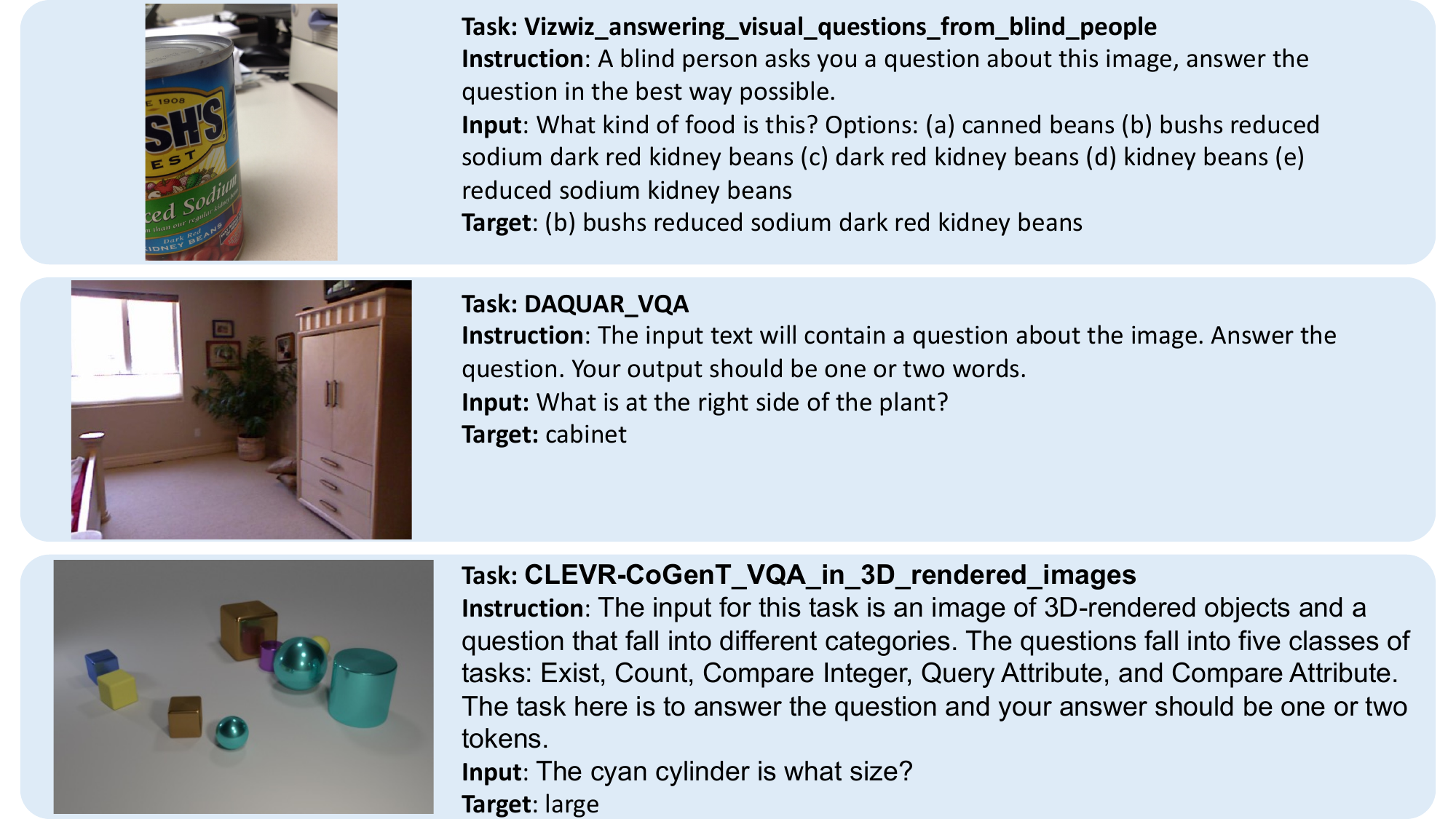}
   \caption{}
   \label{fig:}
\end{figure*}

\begin{figure*}[h!]
  \centering
   \includegraphics[width=\linewidth]{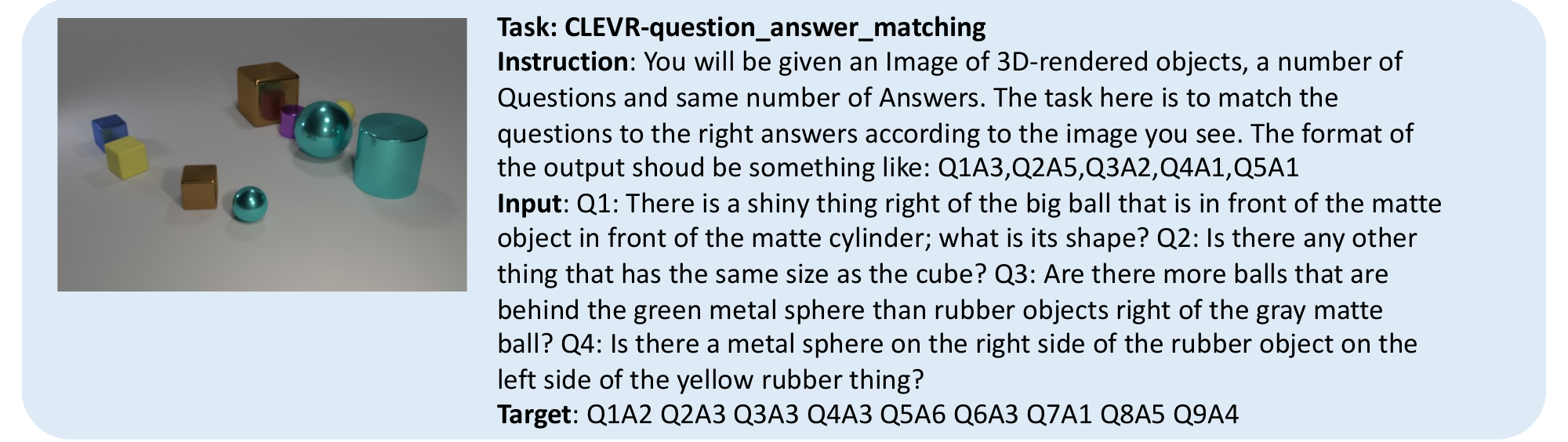}
   \caption{}
   \label{fig:}
\end{figure*}

\begin{figure*}[h!]
  \centering
   \includegraphics[width=\linewidth]{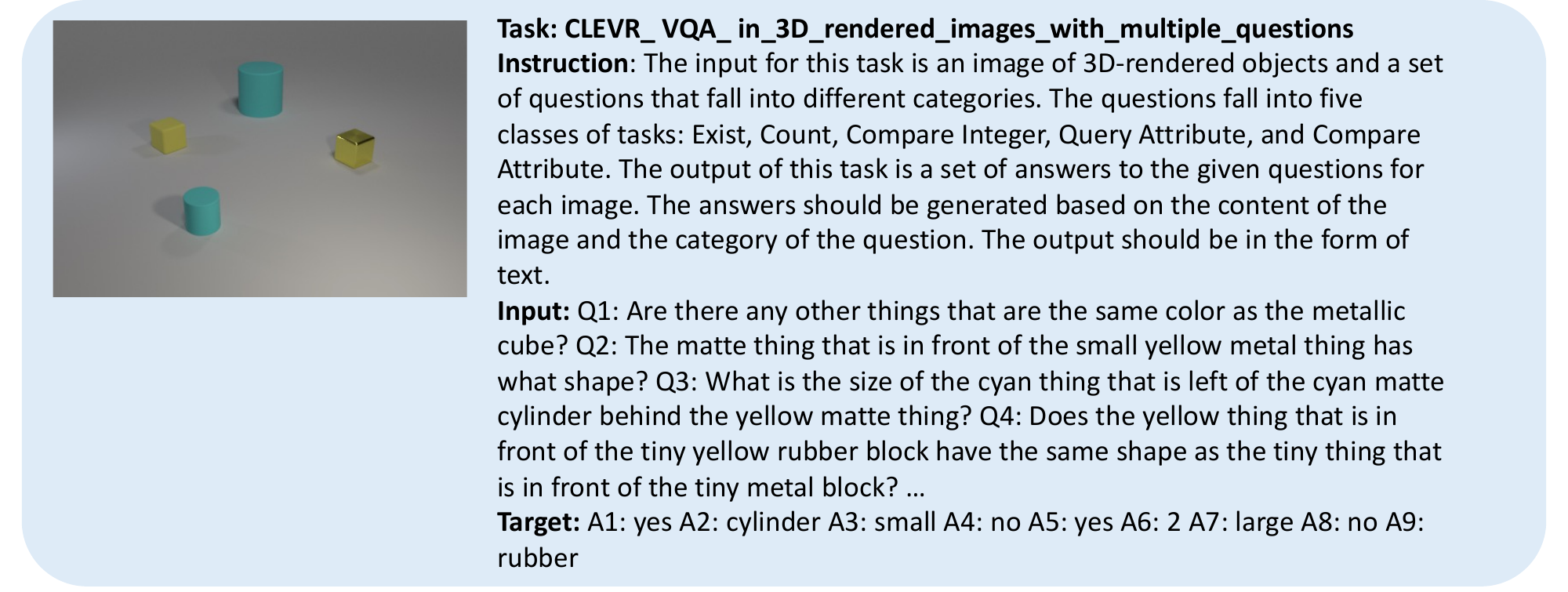}
   \caption{}
   \label{fig:}
\end{figure*}

\begin{figure*}[h!]
  \centering
   \includegraphics[width=\linewidth]{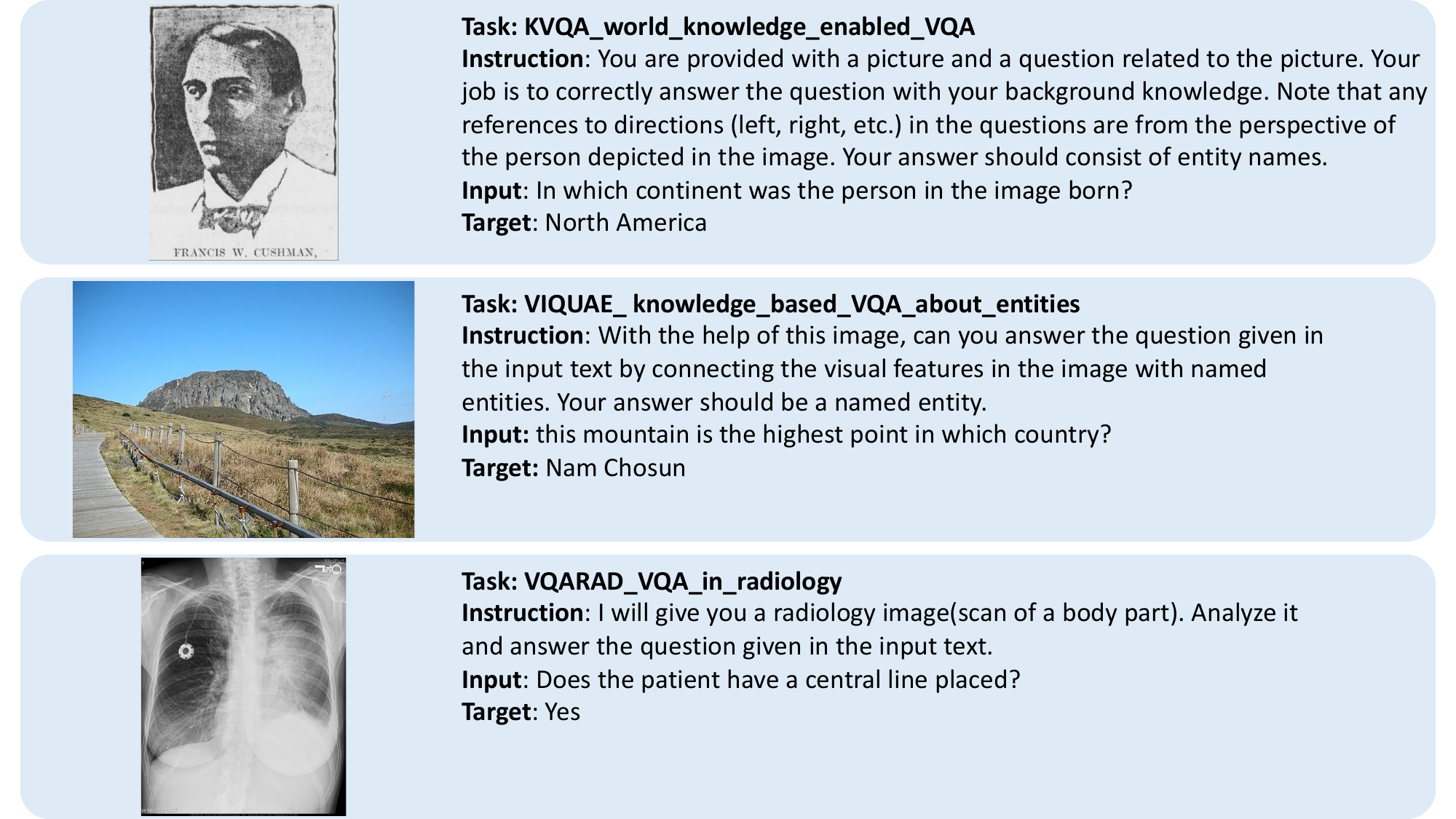}
   \caption{}
   \label{fig:}
\end{figure*}

\begin{figure*}[h!]
  \centering
   \includegraphics[width=\linewidth]{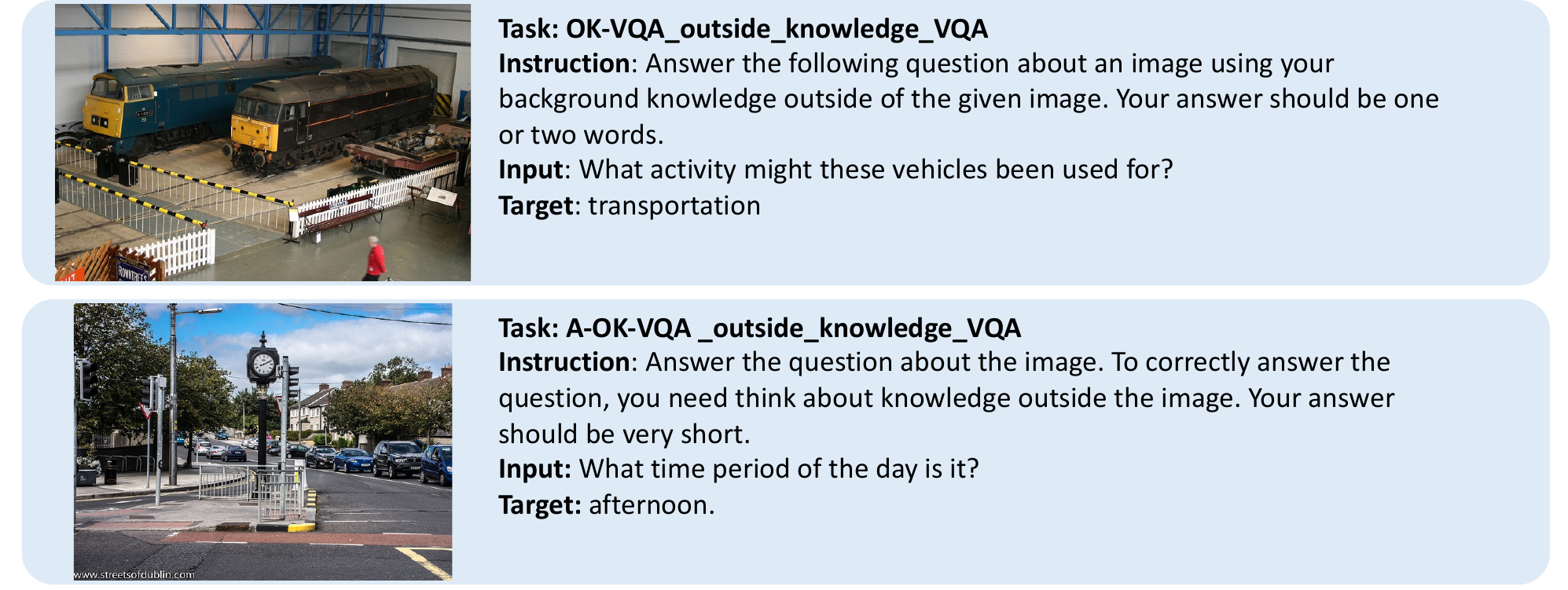}
   \caption{}
   \label{fig:}
\end{figure*}

\begin{figure*}[h!]
  \centering
   \includegraphics[width=\linewidth]{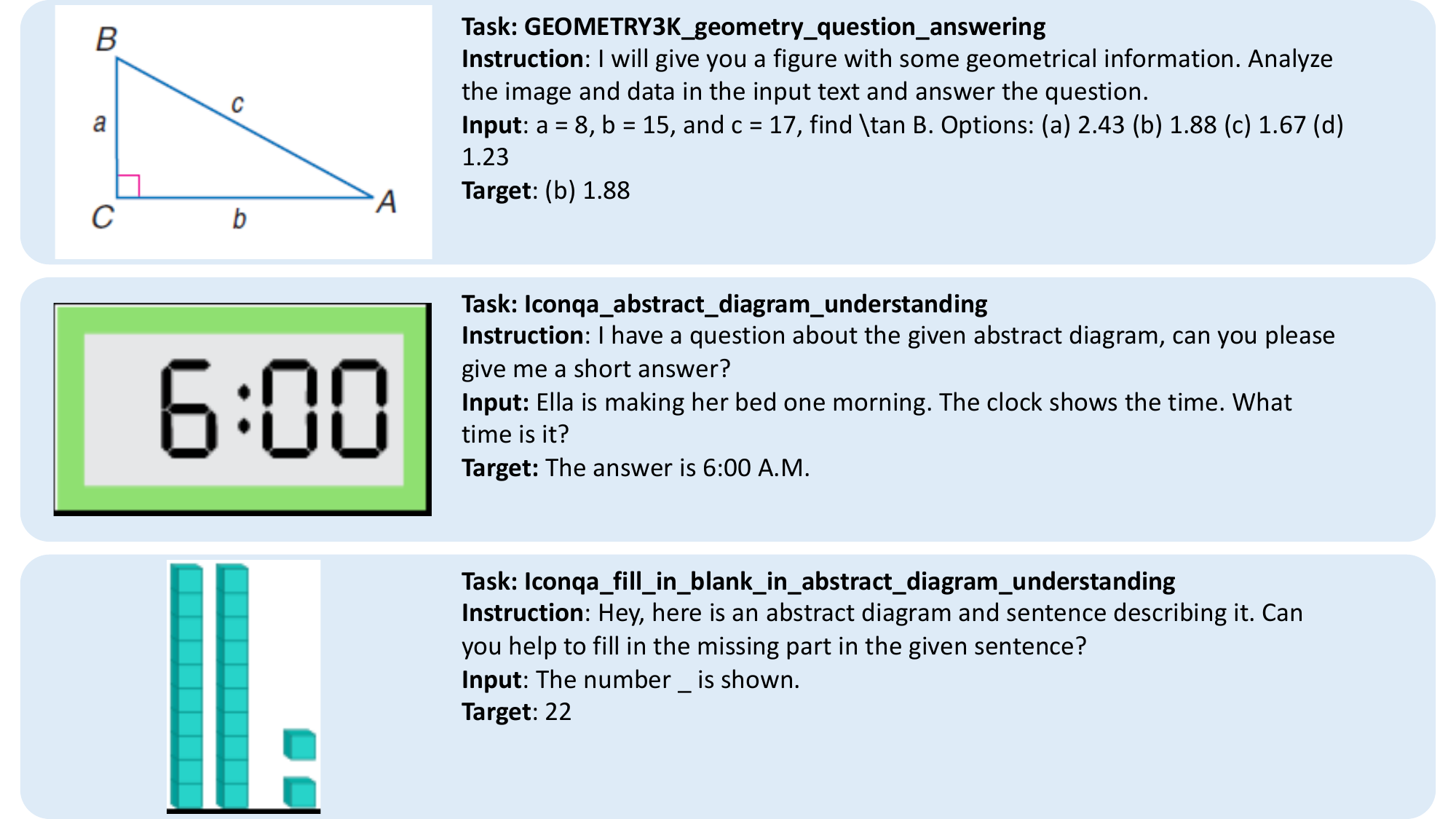}
   \caption{}
   \label{fig:}
\end{figure*}

\begin{figure*}[h!]
  \centering
   \includegraphics[width=\linewidth]{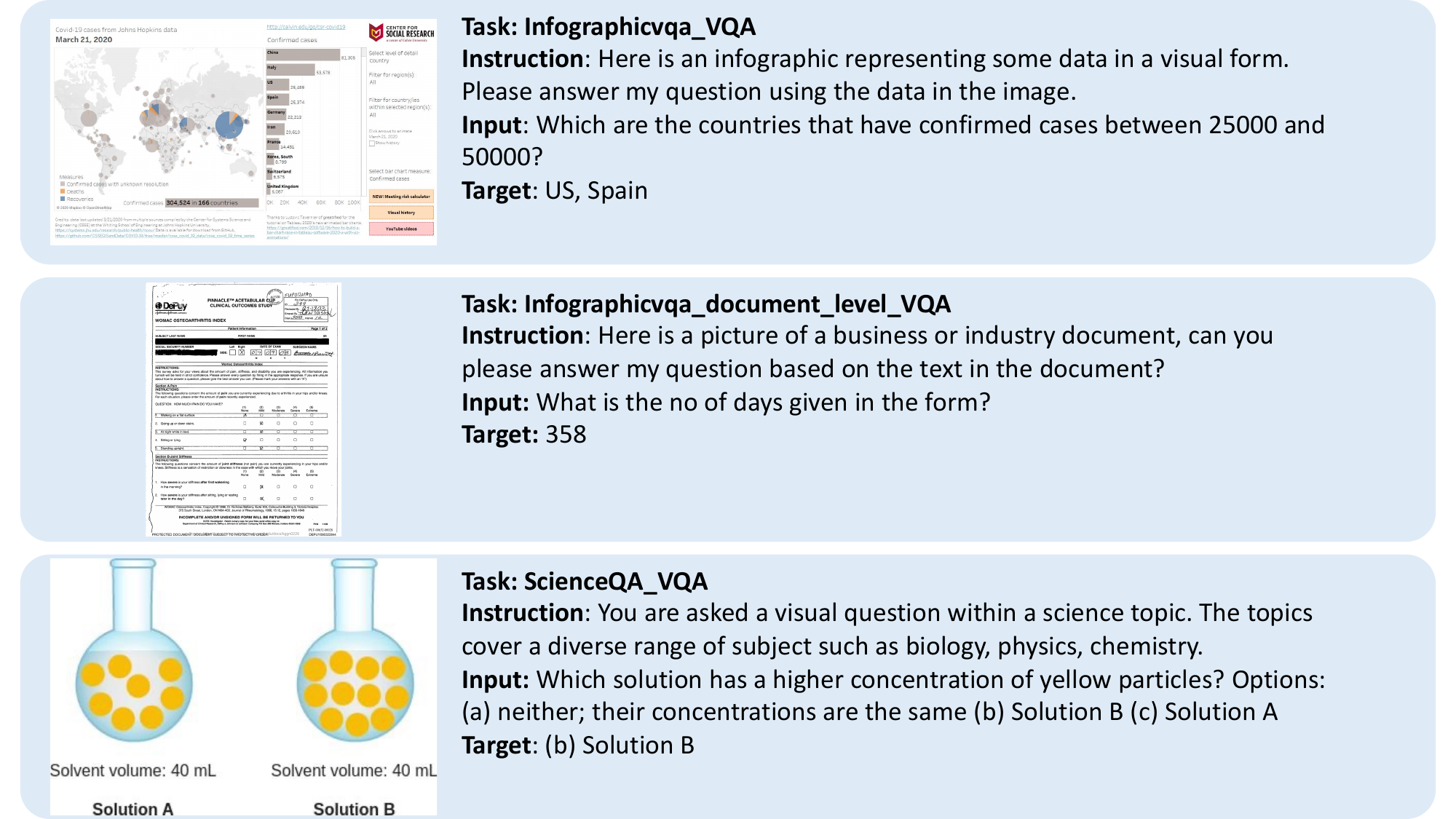}
   \caption{}
   \label{fig:}
\end{figure*}

\begin{figure*}[h!]
  \centering
   \includegraphics[width=\linewidth]{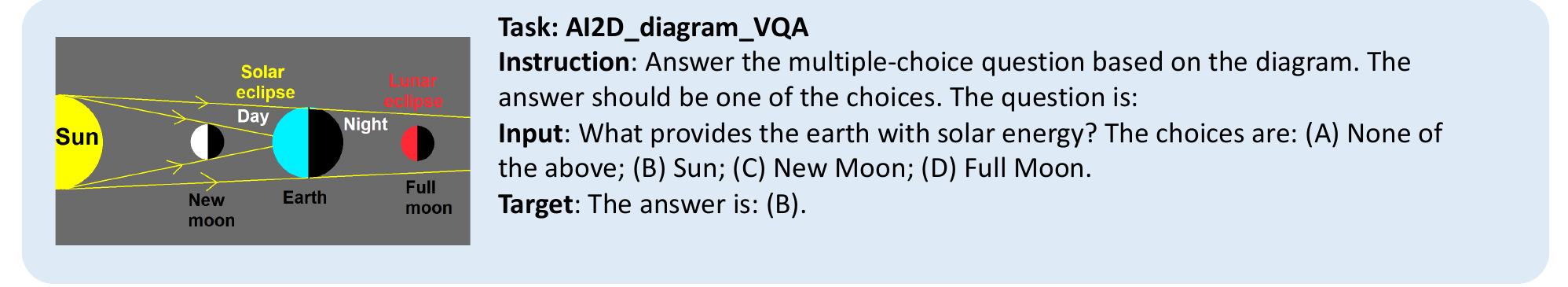}
   \caption{}
   \label{fig:}
\end{figure*}

\begin{figure*}[h!]
  \centering
   \includegraphics[width=\linewidth]{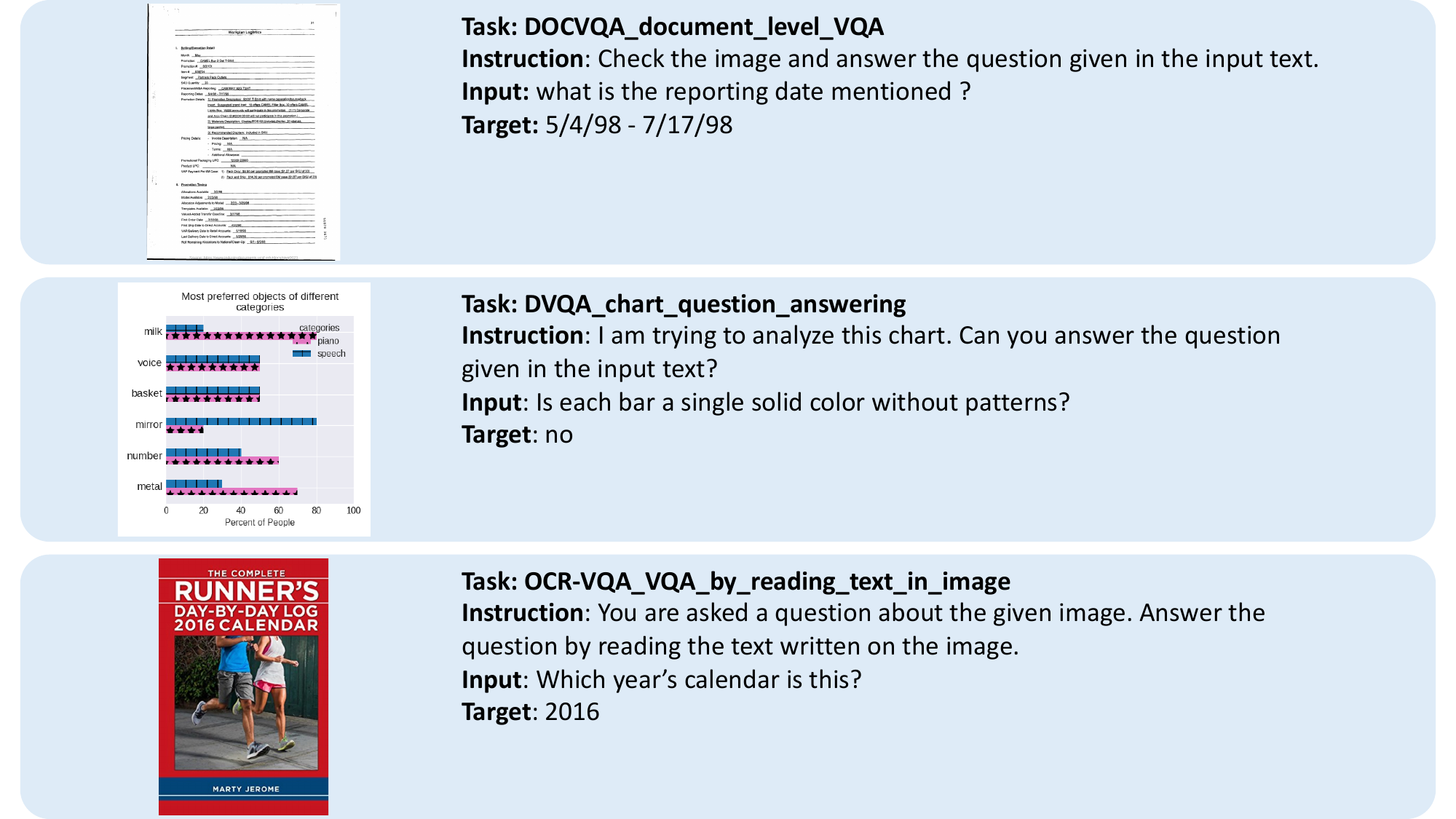}
   \caption{}
   \label{fig:}
\end{figure*}
\end{document}